\documentclass[lettersize,journal]{IEEEtran}
\usepackage{amsmath,amsfonts}
\usepackage{algorithmic}
\usepackage{algorithm}
\usepackage{array}
\usepackage[caption=false,font=normalsize,labelfont=sf,textfont=sf]{subfig}
\usepackage{yfonts}
\usepackage{textcomp}
\usepackage{stfloats}
\usepackage{url}
\usepackage{verbatim}
\usepackage{graphicx}
\usepackage{cite}
\usepackage{arydshln} 

\usepackage{mathtools}
\usepackage{upgreek}
\usepackage{stmaryrd}

\usepackage{tikz}
\usetikzlibrary{3d, arrows.meta}
\usetikzlibrary{calc}
 \usetikzlibrary{shapes.geometric, arrows} \tikzstyle{startstop} = [rectangle, rounded corners, minimum width=2cm, minimum height=1cm,text centered, draw=black, fill=gray!30] \tikzstyle{process} = [rectangle, minimum width=2cm, minimum height=1cm, text centered, draw=black, fill=gray!30] \tikzstyle{decision} = [diamond, minimum width=1cm, minimum height=1cm, text centered, draw=black, fill=gray!30] \tikzstyle{arrow} = [thick,->,>=stealth]

\newcommand{\vect}[1]{\boldsymbol{\mathbf{#1}}}

\newcommand{\Tvect}[1]{\Tilde{\boldsymbol{\mathbf{#1}}}}

\newcommand{\Bvect}[1]{\Bar{\boldsymbol{\mathbf{#1}}}}

\begin{document}

\title{Fault-Tolerant Multi-Modal Localization of Multi-Robots on Matrix Lie Groups 
}

\author{Mahboubeh Zarei and Robin Chhabra,~\IEEEmembership{Senior Member,~IEEE,}
\thanks{M. Zarei is with the Autonomous Space Robotics and Mechatronics Laboratory, Carleton University, Ottawa, ON, Canada.\\
\indent R. Chhabra is with the Mechanical, Industrial, and Mechatronics Engineering Department, Toronto Metropolitan University, Toronto, ON, Canada.}}

\markboth{IEEE Transactions on Robotics,~Vol.~, No.~, April~2025}%
{Multi-Robot Multi-Modal Localization}

\IEEEpubid{0000--0000/00\$00.00~\copyright~2021 IEEE}

\maketitle

\begin{abstract}
  Consistent localization of cooperative multi-robot systems during navigation presents substantial challenges. This paper proposes a fault-tolerant, multi-modal localization framework for multi-robot systems on matrix Lie groups. We introduce novel stochastic operations to perform composition, differencing, inversion, averaging, and fusion of correlated and non-correlated  estimates on Lie groups, enabling pseudo-pose construction for filter updates. The method integrates a combination of proprioceptive and exteroceptive measurements from inertial, velocity, and pose (pseudo-pose) sensors on each robot in an Extended Kalman Filter (EKF) framework. The prediction step is conducted on the Lie group $\mathbb{SE}_2(3) \times \mathbb{R}^3 \times \mathbb{R}^3$, where each robot's pose, velocity, and inertial measurement biases are propagated. The proposed framework uses body velocity, relative pose measurements from fiducial markers, and inter-robot communication to provide scalable EKF update across the network on the Lie group $\mathbb{SE}(3) \times \mathbb{R}^3$. A fault detection module is implemented, allowing the integration of only reliable pseudo-pose measurements from fiducial markers. We demonstrate the effectiveness of the method through experiments with a network of wheeled mobile robots equipped with inertial measurement units, wheel odometry, and ArUco markers. The comparison results highlight the proposed method's real-time performance, superior efficiency, reliability, and scalability in multi-robot localization, making it well-suited for large-scale robotic systems.
\end{abstract}

\section{Introduction}
\IEEEPARstart{M}{obile} robots, whether operating individually or as part of a team, are widely deployed in indoor environments to perform a variety of tasks, such as inspection, source searching, delivery, and quality control~\cite{kwon2012design,wang2024exploration}. Fully autonomous navigation of mobile robots requires accurate, fast, and cost-effective localization systems, optimized by careful selection of motion sensors, information exchange strategies, and computing algorithms. Over the past few decades, numerous localization methods have emerged, integrating proprioceptive and exteroceptive sensors. Inertial odometry, relying on proprioceptive inertial sensors such as accelerometers, gyroscopes, magnetometers, and Inertial Measurement Units (IMUs), provides a simple and self-contained solution~\cite{barshan1995inertial}. However, it faces challenges especially in translational directions of motion due to the well-known noise propagation problem in accelerometers~\cite{borenstein1996measurement,barshan1995inertial}.  To mitigate these errors, velocity measurements---such as those from encoders or optical flow sensors---can be fused with inertial data within a sensor fusion framework~\cite{borenstein1996gyrodometry,cho2011dead,wang2023enhanced,bloeschfusion}. However, these methods still struggle to provide precise estimates due to environmental factors and inherent systematic errors. As a result, periodic exteroceptive position updates using sensors such as camera, Ultra-WideBand (UWB), and LiDAR are essential to maintaining accuracy in a multi-sensor fusion framework~\cite{shalaby2024multi,liu2024edge,perttula2014distributed}. In a multi-robot localization problem, providing pose updates to all agents is often impractical or costly. Alternatively, we can leverage pose updates of neighboring agents to significantly enhance robots' localization via enabling inter-robot communication. Accordingly, leader-follower (collaborative) localization has emerged as a robust and scalable solution~\cite{sharma2008cooperative,mariottini2009vision,van2020board,fang2023integrated}. In this approach, a subset of robots, known as leaders, access accurate global localization information, which is shared with the rest of the robots (followers), improving their pose estimates. Computational algorithms are the last key aspect of the localization system design that are categorized into estimation-based and optimization-based approaches. The estimation-based approaches such as Extended Kalman Filter (EKF) offer significant advantages in handling uncertainty, fusing data from multiple sensors, providing real-time updates, and enabling cost-effective computations~\cite{assa2015kalman}. Moreover, they are  highly scalable, making them well-suited for multi-robot systems~\cite{pedroso2023distributed}. Robots' motion is rigorously captured on the Special Euclidean Lie Group $\mathbb{SE}(3)$, whose elements provide an unambiguous pose representation unlike Euler angles and quaternions ~\cite{kenwright2013quaternions}, removing the need for calculating intricate Jacobians of representation maps. Implementing Lie groups in the estimation process, however, necessitates systematic representation of uncertainty on these nonlinear spaces.

Building upon foundational concepts of pose representation and sensor fusion for localization within the framework of matrix Lie groups, this paper first develops stochastic operations on matrix Lie groups, followed by the presentation of a multi-robot multi-sensor localization system that fuses inertial, velocity, and pose data. To scale the method to include multiple robots, we assume access to global and relative pose information using an exteroceptive sensor such as fiducial markers. In the following section, we explore techniques for estimation and fusion on matrix Lie groups and review relevant works on multi-sensor and multi-robot localization.

\subsection{Related works}

\subsubsection{Estimation on Matrix Lie Groups}
Localization of systems evolving on matrix Lie groups centers around characterizing random pose variables on matrix Lie groups and their associated Lie algebras. The use of differential homogeneous transformations to represent uncertainty on Special Euclidean group 
$\mathbb{SE}(3)$ has been explored in earlier works~\cite{paul1981robot,durrant1987consistent,su1992manipulation}. These studies laid the groundwork for handling uncertainty propagation in the context of rigid body motion, with an emphasis on matrix representations of spatial transformations. A more comprehensive framework for expressing and advancing uncertainty on matrix Lie groups is presented in~\cite{wang2006error,chirikjian2011stochastic,hertzberg2013integrating}.

Recent work by Long \textit{et al.} demonstrated that the observed probability distributions on
$\mathbb{SE}(n)$ are normal in the exponential coordinates~\cite{long2013banana}. This result offers valuable insights into using exponential maps for uncertainty propagation in localization, as the estimation framework can accommodate higher errors without concern for the breakdown of the normality assumption in the presence of high uncertainties. Building on these advancements, Barfoot \textit{et al.}~\cite{barfoot2014associating} proposed a framework for capturing pose uncertainties in the Lie algebra and using the exponential map to propagate these uncertainties for discrete-time variables on $\mathbb{SE}(n)$.  This method was then extended to capture continuous-time random variables on  $\mathbb{SE}(3)$~\cite{tang2019white,wong2020data}. 
To complete the requirements of pose estimation, Mangelson \textit{et al.} enabled operations on two correlated pose estimates, such as composition, inversion, and relative pose, while accounting for uncertainties in the Lie algebra~\cite{mangelson2020characterizing}.  These advancements have facilitated the development of consistent and efficient estimation solutions, such as the Extended Kalman Filter (EKF) on Lie groups~\cite{brossard2018exploiting, heo2018consistent}, the Invariant EKF (IEKF)~\cite{bonnabel2007left, xu2022distributed}, the Unscented Kalman Filter (UKF) on Lie groups~\cite{sjoberg2021lie, bohn2012unscented}, particle filters on Lie groups~\cite{marjanovic2016engineer}, and preintegration methods~\cite{forster2016manifold}.

Fusion of stochastic poses is a critical operation in the localization of robots, especially in scenarios where estimates from different robots are available. Excluding cross-correlations, Drummond \textit{et al.}~\cite{drummond2003computing} and Barfoot \textit{et al.}~\cite{barfoot2014associating} explored iterative naive methods to combine two  and multiple pose estimates, respectively. Non-iterative naive solutions for combining multiple poses on $\mathbb{SE}(3)$ have been introduced in~\cite{wolfe2011bayesian,long2013banana,hwang2022novel,petersen2022tracking2}. To address cross-correlations, Li \textit{et al.}~\cite{li2021joint} presented a conservative iterative method that extends the split Covariance Intersection (CI) approach to matrix Lie groups. Accordingly, conservative distributed IEKF solutions based on CI were developed in~\cite{xu2023distributed} and~\cite{xu2022distributed}. Recently,~\cite{zarei2024consistent} proposed a non-iterative fusion method for multiple pose estimates that ensures consistency by propagating and incorporating cross-correlations.

\subsubsection{Multi-robot Localization on Lie Groups}
  Many localization methods for robots have been developed that rely on sensor fusion techniques such as EKF or UKF, using inertial and visual sensors~\cite{heo2018consistent,brossard2018unscented,tsao2023analytic}, inertial sensor and Global Navigation Satellite System (GNSS)~\cite{du2022lie}, or inertial and velocity measurements~\cite{vial2024lie}. However, their extension to multi-robot systems is not straightforward and requires consideration of communication constraints in the network of robots, combining information from sensors with different modalities, and computational efficiency. Recent studies have explored various approaches to address these challenges.  For instance, Li and Chirikjian~\cite{li2016lie} proposed decentralized localization frameworks for multiple robots based on a second-order sensor fusion technique, focusing on how the geometry of $\mathbb{SE}(3)$ can improve the localization performance.  The method can be computationally expensive as the computation increases with the number of measurements taken and the technique is susceptible to error accumulation when no neighboring robot has an accurate pose estimate.   Dubois \textit{et al.}~\cite{dubois2022sharing,dubois2019data} used visual and inertial sensors to enable decentralized collaborative Simultaneous Localization and Mapping (SLAM) that improves pose estimation by designing of efficient data sharing methods. However, this method has the potential to drift especially when dealing with featureless areas, low lighting conditions, or rapid motion, which can lead to inaccurate map building and localization. Moreover, it is computationally expensive, requiring significant processing power and resources to operate in real-time. Knuth and Barooah~\cite{knuth2009distributed} proposed a model-free distributed framework on $\mathbb{SE}(3)$ for localization of a multi-robot system based on relative poses from the neighboring robots in a GPS-denied environment. The method was evaluated by numerical simulations and there is no analytical guarantee on the performance.   
Shin \textit{et al.}~\cite{shin2024multi} proposed a multi-robot pose estimation framework on $\mathbb{SE}(2)$, with a particular focus on the observability analysis of relative pose estimation. They compared the performance of method utilizing EKF and pose graph optimization, both relying on relative pose data obtained from UWB sensors. To address the inconsistencies of classical distributed EKF on Euclidean spaces, Xu \textit{et al.}~\cite{xu2023distributed} proposed a CI-based distributed IEKF for cooperative target tracking on $\mathbb{SE}_2(3)$, although they overlooked relative noise and IMU biases. Moreover, the IEKF has been applied to distributed cooperative localization~\cite{zhou2024distributed} and multi-robot object-level pose SLAM~\cite{li2024distributed}.

 \subsubsection{Fiducial Marker-based Pose Estimate }
Fiducial markers such as ArUco and AprilTag offer a simple, reliable and cost-effective means to update pose in structured settings~\cite{kalaitzakis2021fiducial}. Marker poses can be used directly or fused with other sensor data in the filter update step~\cite{fourmy2019absolute,zheng2018visual,li2023indoor}. The benefit of fiducial markers is their ability to easily obtain relative pose measurements between robots without relying on external sources. Accordingly, several multi-robot navigation frameworks have been proposed. 
For example, Joon \textit{ et al.} proposed a leader-follower control system with estimation-based localization~\cite{joon2023leader} in which multiple ArUco markers are installed on the leader and the follower robot employs a rotating camera to detect them. Localization is achieved by fusing IMU, encoder, and ArUco pose data through two EKFs for the follower and one for the leader. Zhou \textit{et al.} in~\cite{zhou2021multi} presented a cooperative multi-robot localization, utilizing real-time communication and a two-stage EKF that fuses IMU and encoder data with relative pose measurements obtained from the ArUco marker on the follower robot. In~\cite{li2018cooperative}, a marker-based cooperative localization method is introduced that mimics caterpillar-like motion. It combines absolute pose estimates from static features with relative pose estimates from mobile fiducial markers, framed as an online optimization problem integrating two objectives from different feature sources. Shan \textit{et al.}~\cite{shan2016probabilistic} proposed a Bayesian leader-follower trajectory estimation, where the follower estimates the leader's trajectory using noisy odometry data and relative fiducial marker observations. The leader-follower formation problem based on the relative pose measurements of the leader was introduced in~\cite{oh2023leader}, while ~\cite{jayarathne2019vision} proposed a control law for cooperative object handling in a leader-follower system. All of the aforementioned works were developed within Euclidean spaces and are limited to networks of size two, or leader-follower frameworks.

\subsection{Contributions}
This paper proposes stochastic  operations for multiple correlated and uncorrelated estimates on Lie groups and uses them to advance a fault-tolerant, multi-modal localization method for multi-robot systems evolving on matrix Lie groups. Each robot in the network runs an EKF to fuse inertial measurements on the proposed manifold $\mathbb{SE}_2(3)\times\mathbb{R}^3\times\mathbb{R}^3$ with high-frequency velocity measurements on $\mathbb{R}^3$ or velocity and pseudo-pose data on a measurement manifold of $\mathbb{SE}(3)\times\mathbb{R}^3$. Using relative pose information, inter-robot communications, and developed stochastic operations, the proposed estimation framework enables scalable collaborative localization across all robots in the network.  Thus, the key contributions of this work are as follows:
\begin{itemize}
  \item[(i)] This paper proposes stochastic operations on an arbitrary matrix Lie group. In contrast to previous approaches~\cite{barfoot2014associating,mangelson2020characterizing}, the operations are not limited to poses, batch formulas are developed for multiple correlated estimates, and new operations such as constrained fusion and averaging are considered.  
   
 \item[(ii)] We present a new practical multi-robot multi-modal localization method on Lie groups, which, unlike~\cite{li2016lie,dubois2022sharing,shin2024multi}, fuses inertial measurements on $\mathbb{SE}_2(3)\times\mathbb{R}^3\times\mathbb{R}^3$ with velocity and pose data on $\mathbb{SE}(3)\times\mathbb{R}^3$ . In contrast to previous approaches limited to two robots~\cite{joon2023leader,zhou2021multi,li2018cooperative,shan2016probabilistic}, our method scales cooperative localization across the entire robotic network considering three types of robots: leader, follower type 1 and follower type 2. It also integrates a fault detection module on each robot's local filter to identify the reliable pose measurements.
 
\item[(iv)] We conducted experiments on a network of wheeled mobile robots to evaluate the localization system, combining IMU data with velocity measurements from wheel encoders and pseudo-pose measurements obtained from ArUco markers.

\end{itemize}

\section{Preliminaries}
\subsection{Matrix Lie groups in Robotics}

Here, the extended pose that encompasses vectors such as velocity and landmark positions, is described by a  matrix $\vect{\mathcal{X}}\in\mathbb{R}^{\kappa+3\times\kappa+3}$ that belongs to the 
 $\kappa$-Direct
Isometries group~\cite{barrau2015non}:
\begin{align}
    \mathbb{SE}_\kappa(3)\!\coloneqq\!\begin{Bmatrix}\!\vect{\mathcal{X}}\!=\! \left[\begin{smallmatrix}\vect{\mathcal{R}}&\vect{p}_1&\vect{p}_2&\cdots&\vect{p}_\kappa\\\vect{0}_{3\times 1}^\top
&1&0&\cdots&0\\\vect{0}_{3\times 1}^\top
&0&1&\cdots&0\\\vdots&\vdots&\vdots&\ddots&\vdots\\\vect{0}_{3\times 1}^\top
&0&0&\cdots&1\end{smallmatrix}\!\right]\!\bigg|{\vect{\mathcal{R}}\in \mathbb{SO}(3), \vect{p}_i\in\mathbb{R}^3 }\!\end{Bmatrix},
\end{align}
where $\vect{p}_i,~i\in\{1,\cdots,\kappa\}$, is a translation vector with respect to some reference frame, and $\vect{\mathcal{R}}$ is a rotation matrix that belongs to the Special Orthogonal group $    \mathbb{SO}(3)\coloneqq\begin{Bmatrix}
         \vect{\mathcal{R}}\in\mathbb{R}^{3\times3}|\vect{\mathcal{R}}\vect{\mathcal{R}}^\top=\vect{I}_{3},~\det{(\vect{\mathcal{R}})}=1
     \end{Bmatrix}$. Here, $\vect{I}_n$ and $\vect{0}_n$ denote an $n$-dimensional identity matrix and zero matrix, respectively. Note that for $\kappa=0$ and $\kappa=1$ the group $\mathbb{SE}_\kappa(3)$ reduces to $\mathbb{SO}(3)$ and  $\mathbb{SE}(3)$, respectively.
The tangent space at the identity element $\vect{I}_{\kappa+3}$  along with a bracket operator (commutator) forms the Lie algebra $\mathfrak{se}_\kappa(3)$: 
\begin{align*}
    \mathfrak{se}_\kappa(3)\!\coloneqq\!\begin{Bmatrix}\!
       \left[\vect{\zeta}\right]_\wedge\!=\!\left[\!\begin{smallmatrix}
           \left[\vect{\varphi}\right]_{\times}&\vect{\rho}_1&\vect{\rho}_2&\cdots&\vect{\rho}_\kappa\\\vect{0}_{3\times 1}^\top
&0&0&\cdots&0\\\vect{0}_{3\times 1}^\top
&0&0&\cdots&0\\\vdots&\vdots&\vdots&\ddots&\vdots\\\vect{0}_{3\times 1}^\top
&0&0&\cdots&0 
        \end{smallmatrix}\!\right]\!\in\!\mathbb{R}^{\kappa+3\times\kappa+3}\!\bigg|  \vect{\varphi}, \vect{\rho}_i\in \mathbb{R}^3  
    \!\end{Bmatrix},
\end{align*}
where the operator $\left[.\right]_\wedge$ indicates the isomorphism between  $\mathbb{R}^{3+3\kappa}$ and $\mathfrak{se}_\kappa(3)$, with the inverse operator being denoted by $\left[.\right]_\vee$. That is, $\vect{\zeta}=\left[\begin{smallmatrix}
     \vect{\varphi}& \vect{\rho}_1& \vect{\rho}_2& \cdots&\vect{\rho}_\kappa  
\end{smallmatrix}\right]^\top\in \mathbb{R}^{3+3\kappa}$. 
 Moreover, the operator $\left[.\right]_\times$ generates the skew-symmetric matrix of the vector $\vect{\varphi}$ and $\left[.\right]_\otimes$ does the inverse.

 The exponential map $\exp: \mathfrak{se}_\kappa(3)\rightarrow \mathbb{SE}_\kappa(3)$ maps a Lie algebra element $\left[\vect{\zeta}\right]_\wedge\in\mathfrak{se}_\kappa(3)$ to an element $\vect{\mathcal{X}}\in \mathbb{SE}_\kappa(3)$ whose inverse is called the logarithm mapping $\log: \mathbb{SE}_\kappa(3)\rightarrow \mathfrak{se}_\kappa(3)$, with explicit forms: 
\begin{align}\nonumber
    \vect{\mathcal{X}}&\!=\!\exp(\left[\vect{\zeta}\right]_\wedge)\!=\!\left[\begin{smallmatrix}\exp(\left[\vect{\varphi}\right]_{\times})&\vect{V}(\vect{\varphi})\vect{\rho}_1&\vect{V}(\vect{\varphi})\vect{\rho}_2&\cdots&\vect{V}(\vect{\varphi})\vect{\rho}_\kappa\\\vect{0}_{3\times1}^\top&1&0&\cdots&0\\\vect{0}_{3\times1}^\top&0&1&\cdots&0\\\vdots&\vdots&\vdots&\ddots&\vdots\\\vect{0}_{3\times1}^\top&0&0&\cdots&1\end{smallmatrix}\right],\\\nonumber
    \vect{\zeta}&=\left[\log(\vect{\mathcal{X}})\right]_\vee=\begin{bmatrix}
        \left[\log(\vect{\mathcal{R}})\right]_{\otimes}\\ \vect{V}^{-1}(\left[\log(\vect{\mathcal{R}})\right]_{\otimes})\vect{p}_1\\\vect{V}^{-1}(\left[\log(\vect{\mathcal{R}})\right]_{\otimes})\vect{p}_2\\\vdots\\\vect{V}^{-1}(\left[\log(\vect{\mathcal{R}})\right]_{\otimes})\vect{p}_\kappa
    \end{bmatrix}.
\end{align}

Here, for non-zero $\phi\coloneqq  \lVert \vect{\varphi} \rVert\neq 0$, it is derived~\cite{chirikjian2011stochastic}:
\begin{align}
\vect{V}(\vect{\varphi})&=\vect{I}_{3}+\frac{1-\cos\phi}{\phi^2}\left[\vect{\varphi}\right]_{\times}+\frac{\phi-\sin\phi}{\phi^3}\left[\vect{\varphi}\right]^2_{\times},\\
  \vect{V}^{-1}(\vect{\varphi})&=\vect{I}_{3}-\frac{1}{2}\left[\vect{\varphi}\right]_{\times}+\big(\frac{1}{\phi^2}-\frac{1+\cos\phi}{2\phi\sin\phi}\big)\left[\vect{\varphi}\right]^2_{\times},\\
  \vect{\mathcal{R}}\!=\!\exp&(\left[\vect{\varphi}\right]_{\times})\!=\!\vect{I}_{3}\!+\!\frac{\sin\phi}{\phi}\left[\vect{\varphi}\right]_{\times}\!+\!\frac{1-\cos\phi}{\phi^2}\left[\vect{\varphi}\right]_{\times}^2,
\end{align}
 and when  $\phi=0$, we have $ \vect{\mathcal{R}}=\vect{V}(\vect{\varphi})=\vect{I}_{3}$. Moreover, for  $\psi=\cos^{-1}(\frac{tr(\vect{\mathcal{R}})-1}{2})$, we have $\vect{\varphi}=\left[\log(\vect{\mathcal{R}})\right]_{\otimes}=\left[\frac{\psi}{2\sin\psi}(\vect{\mathcal{R}}-\vect{\mathcal{R}}^\top)\right]_\otimes.$
 

For an element $\vect{\mathcal{X}}\in \mathbb{SE}_\kappa(3)$, the Adjoint $\vect{Ad}_{\vect{\mathcal{X}}}: \mathfrak{se}_k(3)\rightarrow \mathfrak{se}_k(3) $ is defined as a mapping that represents the conjugate action of $\mathbb{SE}_\kappa(3)$ on its Lie algebra, i.e., $\vect{Ad}_{\vect{\mathcal{X}}}(\vect{\zeta})\coloneqq [\vect{\mathcal{X}}\left[\vect{\zeta}\right]_\wedge\vect{\mathcal{X}}^{-1}]_\vee$. Thus, the Adjoint representation of $\mathbb{SE}_\kappa(3)$ takes the matrix form~\cite{murray2017mathematical,chirikjian2011stochastic}:
\begin{align}
   \vect{Ad}_{\vect{\mathcal{X}}}\!=\!\begin{bmatrix}
        \vect{\mathcal{R}}& \vect{0}_{3}&\vect{0}_{3}&\cdots&\vect{0}_{3}\\\left[\vect{p}_1\right]_{\times}\vect{\mathcal{R}}&\vect{\mathcal{R}}&\vect{0}_{3}&\cdots&\vect{0}_{3}\\\left[\vect{p}_2\right]_{\times}\vect{\mathcal{R}}&\vect{0}_{3}&\vect{\mathcal{R}}&\cdots&\vect{0}_{3}\\\vdots&\vdots&\vdots&\ddots&\vdots\\\left[\vect{p}_\kappa\right]_{\wedge}\vect{\mathcal{R}}&\vect{0}_{3}&\vect{0}_{3}&\cdots&\vect{\mathcal{R}}
        \end{bmatrix}\!\in\! \mathbb{R}^{3\kappa+3\times 3\kappa+3}.
\end{align}

For the vector $\vect{\zeta}$, we can also define the adjoint representation of the Lie algebra $\mathfrak{se}_\kappa(3)$ with the mapping $\vect{ad}_{\vect{\zeta}}: \mathfrak{se}_\kappa(3) \rightarrow\mathfrak{se}_\kappa(3)$ defined by
\begin{align}
 \vect{ad}_{\vect{\zeta}}(\vect{\eta})\!\coloneqq\! \big[\!\left[\vect{\zeta}\right]_\wedge\!,\!\left[\vect{\eta}\right]_\wedge\!\big]_\vee\!=\!\big[\left[\vect{\zeta}\right]_\wedge\left[\vect{\eta}\right]_\wedge\!-\!\left[\vect{\eta}\right]_\wedge\left[\vect{\zeta}\right]_\wedge\big]_\vee, 
\end{align}
where $\left[\vect{\zeta}\right]_\wedge,\left[\vect{\eta}\right]_\wedge\in \mathfrak{se}_k(3)$ and $[.,.]$ is the commutator operator. This map in its matrix form is as follows:

\begin{align}
    \vect{ad}_{\vect{\zeta}}\!=\!\!\begin{bmatrix}
     \left[ \vect{\varphi}\right]_{\times}& \vect{0}_{3}&\vect{0}_{3}&\cdots&\vect{0}_{3}\\\left[\vect{\rho}_1\right]_{\times}&\left[\vect{\varphi}\right]_{\times}&\vect{0}_{3}&\cdots&\vect{0}_{3}\\\left[\vect{\rho}_2\right]_{\times}&\vect{0}_{3}&\left[\vect{\varphi}\right]_{\times}   &\cdots&\vect{0}_{3}\\\vdots&\vdots&\vdots&\ddots&\vdots\\\left[\vect{\rho}_\kappa\right]_{\times}&\vect{0}_{3}&\vect{0}_{3}&\cdots&\left[\vect{\varphi}\right]_{\times}\!  
    \end{bmatrix}\!\in\!\mathbb{R}^{3\kappa+3\times 3\kappa+3}.
\end{align}

\subsection{The Baker–Campbell–Hausdorff (BCH) Formula}
Let $\vect{\mathcal{X}}=\exp(\left[\vect{\zeta}\right]_\wedge)\in \mathbb{SE}_\kappa(3)$ and $\vect{\mathcal{Y}}=\exp(\left[\vect{\eta}\right]_\wedge)\in \mathbb{SE}_\kappa(3)$, where $\left[\vect{\zeta}\right]_\wedge,\left[\vect{\eta}\right]_\wedge\in\mathfrak{se}_\kappa(3)$. If 
\begin{align}\label{eq: BCH 1}
\exp(\left[\vect{\lambda}\right]_\wedge)\coloneqq\vect{\mathcal{X}}\vect{\mathcal{Y}}=\exp(\left[\vect{\zeta}\right]_\wedge)\exp(\left[\vect{\eta}\right]_\wedge),
\end{align}
then based on the BCH formula \cite{murray2017mathematical}, 
\begin{align}\label{eq: BCH 4}
\vect{\lambda}\!=\!\vect{\zeta}\!+\!\vect{\eta}\!+\!\frac{1}{2}\vect{ad}_{\vect{\zeta}}\vect{\eta}\!+\!\frac{1}{12} \vect{ad}_{\vect{\zeta}}\vect{ad}_{\vect{\zeta}}\vect{\eta}\!+\!\frac{1}{12}\vect{ad}_{\vect{\eta}}\vect{ad}_{\vect{\eta}}\vect{\zeta} \!+\!\cdots
\end{align}
For small values of $\vect{\eta}$, it is shown in~\cite{barfoot2017state} that
\begin{align}\label{eq: BCH 52}
\vect{\lambda}\approx\vect{\zeta}+\mathcal{J}_r^{-1}(\vect{\zeta})\vect{\eta},
\end{align}
where the right Jacobian of  $\mathbb{SE}_\kappa(3)$ for non-zero rotation angles is~\cite{barfoot2017state,chirikjian2011stochastic}:
\begin{align}\nonumber
\mathcal{J}_r(\vect{\zeta})&=\vect{I}_{3\kappa+3}-\frac{4-\phi \sin(\phi)-4\cos(\phi)}{2\phi^2}\vect{ad}_{\vect{\zeta}}\\\nonumber&+\frac{4\phi-5\sin(\phi)+\phi\cos(\phi)}{2\phi^3}\vect{ad}_{\vect{\zeta}}^2\\\nonumber&-\frac{2-\phi\sin(\phi)-2\cos(\phi)}{2\phi^4}\vect{ad}_{\vect{\zeta}}^3\\\nonumber&+\frac{2\phi-3\sin(\phi)+\phi\cos(\phi)}{2\phi^5}\vect{ad}_{\vect{\zeta}}^4.
\end{align}
When the rotation angle is zero,  $\mathcal{J}_r(\vect{\zeta})=\vect{I}_{3\kappa+3}+\frac{1}{2}\vect{ad}_{\vect{\zeta}}$. 

\subsection{Extended Kalman Filter on Matrix Lie Groups}\label{sec: ekf on lie}
\begin{figure}[htbp]\setlength{\belowcaptionskip}{0pt}
    \begin{center}
\begin{tikzpicture}[node distance=2cm,auto,thick,scale=0.55, every node/.style={scale=0.55},every text node part/.style={align=center}]

\tikzstyle{sensor} = [circle, minimum width=1.4cm,text centered, draw=black, fill=white!30]

\tikzstyle{fusion} = [rectangle, minimum width=2cm, minimum height=1cm, text centered, draw=black, fill=white!30]

\tikzstyle{estimator} = [rectangle, minimum width=1.5cm, minimum height=1cm, text centered, draw=black, fill=white!30]

\node[inner sep=0pt] (LieGroup) at (0,0)
    {\includegraphics[width=.5\textwidth]{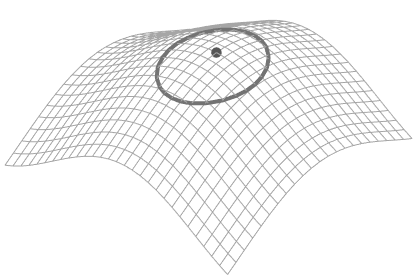}};

\node[inner sep=0pt] (LieAlgebra) at (8,0)
    {\includegraphics[width=.35\textwidth]{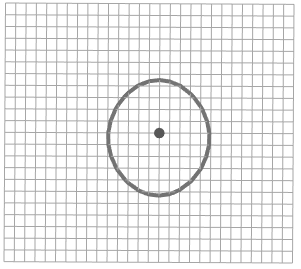}};
    
\draw[-latex,black] ($(LieGroup.north east)+(-2.5,0)$) arc
    [
        start angle=150,
        end angle=25,
        x radius=1.5cm,
        y radius =1cm
    ] ;

 \node[text width=3cm] at (3.5,-3.8) {$\vect{\mathcal{X}}=\Bvect{\mathcal{X}}\exp{([\vect{\zeta}]_\wedge)}$};

\draw[-latex,black] ($(LieAlgebra.south west)+(0,0)$) arc
    [
        start angle=-25,
        end angle=-150,
        x radius=1.5cm,
        y radius =1cm
    ] ;

    \node[text width=3cm] at (3.5,4) {$\vect{\zeta}=[\log(\Bvect{\mathcal{X}}^{-1}\vect{\mathcal{X}})]_\vee$};

    \draw[-latex,blue] ($(LieGroup.north east)+(-4.4,-1.15)$) arc
    [
        start angle=73,
        end angle=18,
        x radius=1.2cm,
        y radius =1.3cm
    ] ;

\draw[-latex,blue] ($(LieAlgebra.north east)+(-3,-2.8)$) arc
    [
        start angle=73,
        end angle=18,
        x radius=1.7cm,
        y radius =0cm
    ] ;

\node[text width=3cm] at (8,-3.25) {Lie algebra $\mathfrak{g}_{\mathcal{X}}$};

\node[text width=3cm] at (0,-3.25) {Lie group $\mathcal{G_X}$};

\node[text width=3cm] at (8.2,-1.75) { $\vect{\zeta}\!\sim\!\mathcal{N}(\vect{0}_{m\times 1},\vect{P})$};

\node[text width=3cm] at (0,0.5) {$\exp{([\vect{\zeta}]_\wedge)}$};

\end{tikzpicture}
 \end{center}
 \caption{Uncertainty on Lie group  $\mathcal{G_X}$ and its Lie algebra $\mathfrak{g}_\mathcal{X}$   }
    \label{fig: lie group and lie algebra}
\end{figure}
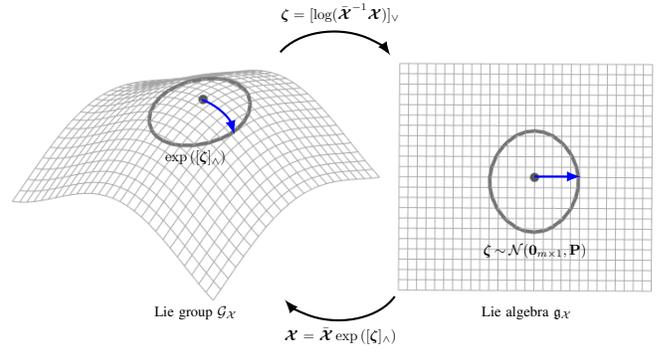

As illustrated in Fig.\ref{fig: lie group and lie algebra}, on an arbitrary Lie group ${\mathcal{G_X}}$, a stochastic member is defined as $\vect{\mathcal{X}}=\Bvect{\mathcal{X}}\exp([\vect{\zeta}]_\wedge)$  
where   $\Bvect{\mathcal{X}}$ denotes the deterministic part and $\vect{\zeta}\sim\mathcal{N}(\vect{0}_{\mathfrak{m}\times 1}, \vect{P})$ is a zero-mean Gaussian vector with the covariance $\vect{P}\in\mathbb{R}^{\mathfrak{m}\times \mathfrak{m}}$. Let a discrete-time system on ${\mathcal{G_X}}$  with noisy measurements from a sensor  be~\cite{bourmaud2015continuous}:
\begin{align}\label{eq: process model}
 & \vect{\mathcal{X}}{\scriptstyle (k+1)}=\!\vect{\mathcal{X}}{\scriptstyle (k)}\exp(\left[\vect{f}(\vect{\mathcal{X}}{\scriptstyle (k)}, \vect{u}{\scriptstyle (k)})\!+\!\vect{w}{\scriptstyle (k)}\right]_\wedge)  ,\\\label{eq: process model2}
  &\vect{\mathcal{Z}}{\scriptstyle (k)}\!=\!\vect{h}(\vect{\mathcal{X}}{\scriptstyle (k)})\exp(\left[{\vect{m}{\scriptstyle (k)}}\right]_{\wedge}), 
\end{align}
where the measurement signal $\vect{\mathcal{Z}}{\scriptstyle (k)}\in{\mathcal{G_Z}}$ is on a $q$-dimensional matrix Lie group according to the model $\vect{h}: {\mathcal{G_X}}\rightarrow{\mathcal{G_Z}}$ and  $\vect{u}{\scriptstyle (k)}\in\mathbb{R}^r$ is the control input to the process whose model is described by the function
$\vect{f}: {\mathcal{G_X}}\times\mathbb{R}^r\rightarrow \mathbb{R}^\mathfrak{m}$. The covariance of the normal white noise sequences $\vect{w}{\scriptstyle (k)}\sim\mathcal{N}(\vect{0}_{\mathfrak{m}\times 1},\vect{Q}{\scriptstyle (k)}) $ and $\vect{m}{\scriptstyle (k)}\sim\mathcal{N}(\vect{0}_{q\times 1},\vect{R}{\scriptstyle (k)})$ are $\vect{Q}{\scriptstyle (k)}\in \mathbb{R}^{\mathfrak{m}\times \mathfrak{m}}$ and $\vect{R}{\scriptstyle (k)}\in\mathbb{R}^{q\times q}$, respectively. We assume that the measurement and process noise signals are independent, i.e., $\mathbb{E}[\vect{m}{\scriptstyle (k)}\vect{w}^\top{\scriptstyle (l)}]=\vect{0}_{q\times \mathfrak{m}}$.


We revisit the method of discrete-time EKF on Lie groups~\cite{bourmaud2013discrete} to set our notation. 
Given the mean and covariance at the time ${\scriptstyle k-1}$, represented as $\Bvect{\mathcal{X}}{\scriptstyle (k-1|k-1)}$ and $\vect{P}{\scriptstyle (k-1|k-1)}$, respectively, the prediction step calculates the prior estimates
\begin{align}\label{eq: mean and cov in propagation}
 \Bvect{\mathcal{X}}{\scriptstyle (k|k-1)}&=\Bvect{\mathcal{X}}{\scriptstyle (k-1|k-1)}\exp(\left[\Bvect{f}{\scriptstyle (k-1)}\right]_\wedge),\\\label{eq: mean and cov in propagation 2}\nonumber 
 \vect{P}{\scriptstyle (k|k-1)}&=\vect{\mathcal{F}}{\scriptstyle (k-1)}\vect{P}{\scriptstyle (k-1|k-1)}\vect{\mathcal{F}}^\top{\scriptstyle (k-1)}\\&+\vect{\mathcal{J}}_r(\Bvect{f}{\scriptstyle (k-1)})\vect{Q}{(\scriptstyle k-1)}\vect{\mathcal{J}}_r^\top(\Bvect{f}{\scriptstyle (k-1)}),
\end{align}
where $\Bvect{f}({\scriptstyle k-1})=\vect{f}(\Bvect{\mathcal{X}}({\scriptstyle k-1|k-1}), \vect{u}({\scriptstyle k-1}))$ and
\begin{align}\label{eq: matrix F}
 & \vect{\mathcal{F}}{\scriptstyle (k)}=\vect{Ad}_{\exp(\left[-\Bvect{f}{\scriptstyle (k-1)}\right]_\wedge)}+ \vect{\mathcal{J}}_r\big(\Bvect{f}{\scriptstyle (k-1)}\big)\vect{\mathcal{D}}{\scriptstyle (k-1)},\\\label{eq: matrix D}
&\vect{\mathcal{D}}{\scriptstyle (k-1)}\!=\!\frac{\partial}{\partial \vect{\zeta}}\vect{f}(\Bvect{\mathcal{X}}{\scriptstyle (k-1|k-1)}\exp([\vect{\zeta}]_\wedge),\vect{u}{\scriptstyle (k-1)})\Big|_{\vect{\zeta}=\vect{0}}. 
 \end{align}
At time step ${\scriptstyle k}$, the update step is executed based on 
 \begin{align}
    \Bvect{\mathcal{X}}{\scriptstyle (k|k)}&=\Bvect{\mathcal{X}}{\scriptstyle (k|k-1)}\exp\big(\big[\vect{K}{\scriptstyle (k)}\vect{\nu}{\scriptstyle (k)}\big]_\wedge\big),\\\nonumber
    \vect{P}{\scriptstyle (k|k)}&=\vect{\mathcal{J}}_r\big(\vect{K}{\scriptstyle (k)}\vect{\nu}{\scriptstyle (k)}\big)\big(\vect{I}_{\mathfrak{m}}-\vect{K}{\scriptstyle (k)}\vect{H}{\scriptstyle (k)}\big)\vect{P}{\scriptstyle (k|k-1)}\\&~~~\vect{\mathcal{J}}_r^\top\big(\vect{K}{\scriptstyle (k)}\vect{\nu}{\scriptstyle (k)}\big),
\end{align}
where the innovation $\vect{\nu}{\scriptstyle (k)}$, the EKF gain, and matrix $\vect{\mathcal{H}}{\scriptstyle (k)}$ are
\begin{align}
    \vect{\nu}{\scriptstyle (k)}&=\big[\log\big(\vect{h}^{-1}(\Bvect{\mathcal{X}}{\scriptstyle (k|k-1)})\vect{\mathcal{Z}}{\scriptstyle (k)}\big)\big]_\vee,\\\label{eq: kalman gain}
    \vect{K}{\scriptstyle (k)}&\!=\!\vect{P}{\scriptstyle (k|k-1)}\vect{\mathcal{H}}^\top\!{\scriptstyle (k)}\big(\vect{\mathcal{H}}{\scriptstyle (k)}\vect{P}{\scriptstyle (k|k-1)}\vect{\mathcal{H}}^\top\!{\scriptstyle (k)}\!+\!\vect{R}{\scriptstyle (k)}\big)^{-1}\!,\!\\\label{eq: matrix H}
    \vect{\mathcal{H}}{\scriptstyle (k)}&\!=\!\frac{\partial}{\partial\vect{\zeta}}\!\big[\log\big(\scriptstyle{\vect{h}^{-1}(\Bvect{\mathcal{X}}(k|k-1))\vect{h}(\Bvect{\mathcal{X}}(k|k-1)\exp([\vect{\zeta}]_\wedge))\big)\big]_\vee\!\Big|_{\vect{\zeta}=\vect{0}}}.
\end{align}

\section{Stochastic Operations on Lie Groups}\label{sec: stochastic configuration}
\begin{figure}[htbp]
  \unitlength=0.1in
   \centering 
   
   \subfloat[]{\scalebox{0.8}{\begin{tikzpicture}
 \draw[thick, gray!50,  dotted] (-0.5,-1) grid[step=0.5] (4.5,1);
\draw[->, thick, >=stealth, line width=0.6mm,gray] (0,0) .. controls (1,0.75) and (2.2,1.) .. (4,0) node[midway, below,yshift=-0.1cm,xshift=0.3cm,font=\fontsize{8}{12}] {$\vect{\mathcal{X}}_{\text{inv}}\!\sim\!(\Bvect{\mathcal{X}}_{\text{inv}},\vect{P}_{\text{inv}})$};
\draw[->, thick, >=stealth, line width=0.6mm] (4,0) .. controls (2.2,-1) and (1.2,-.5) .. (0,0) node[midway, below,yshift=0.6cm,xshift=0.3cm,font=\fontsize{8}{12}] {$\vect{\mathcal{X}}\!\sim\!(\Bvect{\mathcal{X}},\vect{P})$};
\end{tikzpicture}}} \subfloat[]{\scalebox{0.8}{ \begin{tikzpicture}
             \draw[thick, gray!50,  dotted] (-0.5,-1) grid[step=0.5] (5,1);
            \draw[->, thick, >=stealth, line width=0.6mm] (0,0) .. controls (0.5,-0.25) and (1,-0.25) .. (1.5,0) node[midway, below,yshift=0cm,xshift=0.25cm,font=\fontsize{8}{12}] {$\vect{\mathcal{X}}_1\!\sim\!(\Bvect{\mathcal{X}}_1,\vect{P}_{11})$};
            \draw[thick, dotted, line width=0.6mm] (1.5,0) .. controls (2,-0.25) and (2.5,-0.25) .. (3,0)  node[midway, below,yshift=-0.1cm,xshift=0.1cm] {$\cdots$};
            \draw[->, thick, >=stealth, line width=0.6mm] (3,0) .. controls (3.5,-0.25) and (4,-0.25) .. (4.5,0) node[midway, below,font=\fontsize{8}{12}] {$\vect{\mathcal{X}}_{n}\!\sim\!(\Bvect{\mathcal{X}}_{n},\vect{P}_{nn})$};
            \draw[->, thick, >=stealth, line width=0.6mm,gray] (0,0) .. controls (1,1.) and (2.5,1.25) .. (4.5,0) node[midway, below,yshift=-0.2cm,xshift=0.3cm,font=\fontsize{8}{12}] {$\vect{\mathcal{X}}_{\text{com}}\!\sim\!(\Bvect{\mathcal{X}}_{\text{com}},\vect{P}_{\text{com}})$};
        \end{tikzpicture}}}\\ \subfloat[]{\scalebox{0.8}{\begin{tikzpicture}
 \draw[thick, gray!50,  dotted] (-0.5,-0.5) grid[step=0.5] (4.5,2);
\draw[->, thick, >=stealth, line width=0.6mm,gray] (0,0.5) .. controls (0.5,1.15) and (1,1.15) .. (1.5,1.25) node[midway, below,yshift=0cm,xshift=0.75cm,font=\fontsize{8}{12}] {$\vect{\mathcal{X}}_{\text{dif}}\!\sim\!(\Bvect{\mathcal{X}}_{\text{dif}},\vect{P}_{\text{dif}})$};
\draw[->, thick, >=stealth, line width=0.6mm] (4,0.5) .. controls (2,-0.45) and (1.,0) .. (0.0,0.5) node[midway, below,yshift=0.65cm,xshift=0.3cm,font=\fontsize{8}{12}] {$\vect{\mathcal{X}}_1\!\sim\!(\Bvect{\mathcal{X}}_1,\vect{P}_{11})$};
\draw[->, thick, >=stealth, line width=0.6mm] (4,0.5) .. controls (3,1.15) and (2.5,1.25) .. (1.5,1.25) node[midway, below,yshift=0.6cm,xshift=0.6cm,font=\fontsize{8}{12}] {$\vect{\mathcal{X}}_2\!\sim\!(\Bvect{\mathcal{X}}_2,\vect{P}_{22})$};
\end{tikzpicture}}}\subfloat[]{\scalebox{0.8}{\begin{tikzpicture}
 \draw[thick, gray!50,  dotted] (-0.5,-1.5) grid[step=0.5] (5,1.);
\draw[->, thick, >=stealth, line width=0.6mm] (0,0) .. controls (1.5,1.) and (3,0.75) .. (4.5,.1) node[midway, above,yshift=-0.15cm,xshift=1.5cm,font=\fontsize{8}{12}] {$\vect{\mathcal{X}}_1\!\sim\!(\Bvect{\mathcal{X}}_1,\vect{P}_{11})$};
\draw[->, thick,dotted, >=stealth, line width=0.6mm] (0,0) .. controls (1.5,-.5) and (3,-1) .. (4.5,0.1) node[midway, above,yshift=-0.0cm,xshift=0.3cm,font=\fontsize{8}{12}] {};
\draw[->, thick, >=stealth, line width=0.6mm,gray] (0,0) .. controls (2,-.5) and (3.5,-0.5) .. (4.5,0.1) node[midway, above,yshift=0.0cm,xshift=0.1cm,font=\fontsize{8}{12}] {$\vect{\mathcal{X}}^\star_\text{fus}\!\sim\!(\Bvect{\mathcal{X}}^\star_\text{fus},\vect{P}_{\text{fus}})$};
\draw[->, thick, >=stealth, line width=0.6mm] (0,0) .. controls (1.75,-1) and (3,-1.25) .. (4.5,.1) node[midway, below,yshift=-0.0cm,xshift=0.3cm,font=\fontsize{8}{12}] {$\vect{\mathcal{X}}_n\!\sim\!(\Bvect{\mathcal{X}}_n,\vect{P}_{nn})$};
\draw[->,white] (0,0) -- (2,0) node[midway, above,yshift=0.1cm,xshift=1cm,font=\fontsize{8}{12},black] {$\vdots$};
\end{tikzpicture}}}\\
            \caption{Illustration of stochastic   operations: (a) inverse, (b) composition,  (c) difference, and (d) fusion.}
    \label{fig:config_operations}
\end{figure}
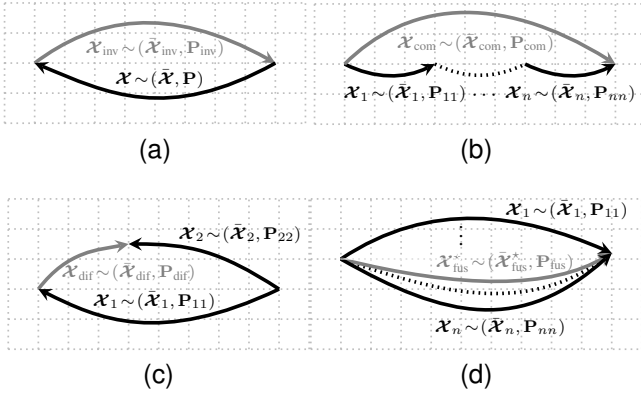

Here, we develop necessary mechanisms to perform fundamental operations on stochastic members of a matrix Lie group. The operations include composition, inverse, difference, averaging, and fusion, which are crucial for the development of our proposed localization algorithm. As shown in Fig.~\ref{fig:config_operations}, each operation derives an equivalent stochastic member based on the mean and covariance of the input members.

\subsection{Inverse Operation}
Let $\mathcal{G_X}$ be an arbitrary Lie group. The inverse operation is illustrated in~Fig.\ref{fig:config_operations}-(a), where for the stochastic member $\vect{\mathcal{X}}=\Bvect{\mathcal{X}}\exp([\vect{\zeta}]_\wedge)\in{\mathcal{G_X}}$ with $\vect{\zeta}\sim\mathcal{N}(\vect{0}_{\mathfrak{m}\times 1},\vect{P})$, its inverse  is defined as $\vect{\mathcal{X}}_{\text{inv}}=\vect{\mathcal{X}}^{-1}=\Bvect{\mathcal{X}}_{\text{inv}}\exp([\vect{\zeta}_{\text{inv}}]_\wedge)\in{\mathcal{G_X}}$ with $\vect{\zeta}_{\text{inv}}\sim\mathcal{N}(\vect{0}_{\mathfrak{m}\times1},\vect{P}_{\text{inv}})$. The mean $\Bvect{\mathcal{X}}_{\text{inv}}\in{\mathcal{G_X}}$ and covariance matrix $\vect{P}_{\text{inv}}\in\mathbb{R}^{\mathfrak{m}\times\mathfrak{m}}$ are calculated as (see Appendix~\ref{app: inverse}): 
\begin{align}
 \Bvect{\mathcal{X}}_{\text{inv}}&=\Bvect{\mathcal{X}}^{-1},\\ \vect{P}_{\text{inv}}&=\vect{Ad}_{\Bvect{\mathcal{X}}}\vect{P}\vect{Ad}_{\Bvect{\mathcal{X}}}^\top.
\end{align}

\subsection{Composition Operation}
As shown in Fig.\ref{fig:config_operations}-(b), let the stochastic  member $\vect{\mathcal{X}}_{\text{com}}\in\mathcal{G_X}$ be composed of
\begin{align}
\label{eq::randomPose}
\vect{\mathcal{X}}_\text{com}=\vect{\mathcal{X}}_1\vect{\mathcal{X}}_2\cdots\vect{\mathcal{X}}_n,
\end{align}
where the $i^{th}$ stochastic member is $\vect{\mathcal{X}}_i=\Bvect{\mathcal{X}}_i\exp([\vect{\zeta}_i]_\wedge)\in\mathcal{G_X}$  with $\vect{\zeta}_i\sim\mathcal{N}(\vect{0}_{\mathfrak{m}\times 1},\vect{P}_{ii})$ and   non-zero cross-covariance matrices are $\vect{P}_{ij}=\mathbb{E}[\vect{\zeta}_i\vect{\zeta}_j^\top]$ for $i,j\in\{1,\cdots,n\}, i\neq j$. The composition $\vect{\mathcal{X}}_{\text{com}}$ is calculated as
\begin{align}\nonumber
\vect{\mathcal{X}}_{\text{com}}&=\Bvect{\mathcal{X}}_1\Bvect{\mathcal{X}}_2\cdots\Bvect{\mathcal{X}}_{n-1}\Bvect{\mathcal{X}}_n\exp([\vect{Ad}_{\Bvect{\mathcal{X}}_n^{-1}\cdots\Bvect{\mathcal{X}}_2^{-1}}\vect{\zeta}_1]_\wedge)\\\nonumber&\exp([\vect{Ad}_{\Bvect{\mathcal{X}}_n^{-1}\cdots\Bvect{\mathcal{X}}_3^{-1}}\vect{\zeta}_2]_\wedge)\cdots\exp([\vect{Ad}_{\Bvect{\mathcal{X}}_n^{-1}}\vect{\zeta}_{n-1}]_\wedge)\\&\exp([\vect{\zeta}_n]_\wedge).\end{align}
We let $\vect{\mathcal{X}}_{\text{com}}=\Bvect{\mathcal{X}}_{\text{com}}\exp([\vect{\zeta}_{\text{com}}]_\wedge)$ with  $\vect{\zeta}_{\text{com}}\sim\mathcal{N}(\vect{0}_{\mathfrak{m}\times 1},\vect{P}_{\text{com}})$, then the mean $\Bvect{\mathcal{X}}_\text{com}\in\mathcal{G_X}$ and its associated  covariance matrix $\vect{P}_{\text{com}}\in\mathbb{R}^{\mathfrak{m}\times\mathfrak{m}}$  are  
\begin{align}
\label{eq::randomPose205}
&\Bvect{\mathcal{X}}_{\text{com}}=\Bvect{\mathcal{X}}_1\Bvect{\mathcal{X}}_2\cdots\Bvect{\mathcal{X}}_{n-1}\Bvect{\mathcal{X}}_n,\\&
\vect{P}_{\text{com}}\approx\vect{\mathcal{M}}\begin{bmatrix}
  \vect{P}_{nn}&\vect{P}_{nn-1}&\cdots&\vect{P}_{n1}\\ 
  \vect{P}_{n-1n}&\vect{P}_{n-1n-1}&\cdots&\vect{P}_{n-11}\\
  \vdots&\vdots&\ddots&\vdots\\
  \vect{P}_{1n}&\vect{P}_{1n-1}&\cdots&\vect{P}_{11}
 \end{bmatrix}\vect{\mathcal{M}}^\top,
 \end{align}
where $\vect{\mathcal{M}}=\text{diag}\begin{bmatrix}
    \vect{I}_{\mathfrak{m}}&\vect{Ad}_{\Bvect{\mathcal{X}}_n^{-1}}&\cdots&\vect{Ad}_{\Bvect{\mathcal{\mathcal{X}}}_n^{-1}\cdots\Bvect{\mathcal{X}}_2^{-1}}
\end{bmatrix}\in\mathbb{R}^{\mathfrak{m}n\times \mathfrak{m}n}$. The proof is detailed in Appendix~\ref{app: comp}.


\subsection{Difference Operation}
According to Fig.\ref{fig:config_operations}-(c), for two stochastic members of $\mathcal{G_X}$, $\vect{\mathcal{X}}_1=\Bvect{\mathcal{X}}_1\exp([\vect{\zeta}_1]_\wedge)$ and $\vect{\mathcal{X}}_2=\Bvect{\mathcal{X}}_2\exp([\vect{\zeta}_2]_\wedge)$, where $\vect{\zeta}_i\sim\mathcal{N}(\vect{0}_{\mathfrak{m}\times1},\vect{P}_{ii}),~i\in\{1,2\}$, the difference is defined as 
\begin{align}
    \vect{\mathcal{X}}_{\text{dif}}=\vect{\mathcal{X}}_1^{-1}\vect{\mathcal{X}}_2.
\end{align}
This operation is a special case of composition when the first input is inverted. We let $\vect{\mathcal{X}}_{\text{dif}}=\Bvect{\mathcal{X}}_{\text{dif}}\exp([\vect{\zeta}_{\text{dif}}]_\wedge)$ with   $\vect{\zeta}_{\text{dif}}\sim\mathcal{N}(\vect{0}_{\mathfrak{m}\times 1},\vect{P}_{\text{dif}})$, then its mean  $\Bvect{\mathcal{X}}_{\text{dif}}$ and covariance matrix $\vect{P}_{\text{dif}}$  are (see Appendix~\ref{app: diff}):
\begin{align}
 \Bvect{\mathcal{X}}_{\text{dif}}&=\Bvect{\mathcal{X}}_1^{-1}\Bvect{\mathcal{X}}_2,\\\nonumber
 \vect{P}_{\text{dif}}&\approx\vect{P}_{22}-\vect{P}_{12}^\top\vect{Ad}_{\Bvect{\mathcal{X}}_2^{-1}\Bvect{\mathcal{X}}_1}^\top-\vect{Ad}_{\Bvect{\mathcal{X}}_2^{-1}\Bvect{\mathcal{X}}_1}\vect{P}_{12}\\&+\vect{Ad}_{\Bvect{\mathcal{X}}_2^{-1}\Bvect{\mathcal{X}}_1}\vect{P}_{11}\vect{Ad}_{\Bvect{\mathcal{X}}_2^{-1}\Bvect{\mathcal{X}}_1}^\top.
\end{align}

\subsection{Averaging Operation}
Here, the averaging operation of stochastic members $\vect{\mathcal{X}}_i=\Bvect{\mathcal{X}}_i\exp([\vect{\zeta}_i]_\wedge)\in\mathcal{G_X}$  with $\vect{\zeta}_i\sim\mathcal{N}(\vect{0}_{\mathfrak{m}\times 1},\vect{P}_{ii})$ for $i\in\{1,\cdots,n\}$ is defined as
\begin{align}
\label{eq::randomPose200}
\vect{\mathcal{X}}_\text{avg}=\vect{\mathcal{X}}_1^{\alpha_1}\vect{\mathcal{X}}_2^{\alpha_2}\cdots\vect{\mathcal{X}}_n^{\alpha_n},
\end{align}
where $\sum_{i=1}^n\alpha_i=1$ and $\alpha_i\in[0,1]$. We let $\vect{\mathcal{X}}_{\text{avg}}=\Bvect{\mathcal{X}}_{\text{avg}}\exp([\vect{\zeta}_{\text{avg}}]_\wedge)$ with  $\vect{\zeta}_{\text{avg}}\sim\mathcal{N}(\vect{0}_{\mathfrak{m}\times 1},\vect{P}_{\text{avg}})$, then the mean $\Bvect{\mathcal{X}}_\text{avg}\in\mathcal{G_X}$ and covariance matrix $\vect{P}_{\text{avg}}\in\mathbb{R}^{\mathfrak{m}\times\mathfrak{m}}$  are  
\begin{align}
\label{eq::randomPose2}
&\Bvect{\mathcal{X}}_{\text{avg}}=\Bvect{\mathcal{X}}_1^{\alpha_1}\Bvect{\mathcal{X}}_2^{\alpha_2}\cdots\Bvect{\mathcal{X}}_{n-1}^{\alpha_{n-1}}\Bvect{\mathcal{X}}_n^{\alpha_n},\\
&\vect{P}_{\text{avg}}\!\approx\!\vect{\mathcal{M}}_\alpha\!\begin{bmatrix}
  \vect{P}_{nn}&\vect{P}_{nn-1}&\cdots&\vect{P}_{n1}\\ 
  \vect{P}_{n-1n}&\vect{P}_{n-1n-1}&\cdots&\vect{P}_{n-11}\\
  \vdots&\vdots&\ddots&\vdots\\
  \vect{P}_{1n}&\vect{P}_{1n-1}&\cdots&\vect{P}_{11}
 \end{bmatrix}\vect{\mathcal{M}}_\alpha^\top,
 \end{align}
where $\vect{\mathcal{M}}_\alpha=\text{diag}\left[\begin{smallmatrix}
    \alpha_n\vect{I}_{\mathfrak{m}}&\alpha_{n-1}\vect{Ad}_{\Bvect{\mathcal{X}}_n^{-\alpha_n}}&\cdots&\alpha_1\vect{Ad}_{\Bvect{\mathcal{\mathcal{X}}}_n^{-\alpha_n}\cdots\Bvect{\mathcal{X}}_2^{-\alpha_2}}
\end{smallmatrix}\right]$. The proof follows a procedure similar to that of the composition in Appendix~\ref{app: comp}, using the fact that $\vect{\mathcal{X}}_i^{\alpha_i}=\Bvect{\mathcal{X}}_i^{\alpha_i}\exp(\alpha_i[\vect{\zeta}_i]_\wedge)=\Bvect{\mathcal{X}}_i^{\prime}\exp([\vect{\zeta}_i^\prime]_\wedge)$ for the $i^{th}$ member.


\subsection{ Fusion Operation}\label{sec: configuration fusion}
As illustrated in Fig.\ref{fig:config_operations}-(d), the  fusion problem aims to find an optimum consistent fused member $ \vect{\mathcal{X}}^\star_\text{fus}=\Bvect{\mathcal{X}}^\star_\text{fus}\exp([\vect{\zeta}_\text{fus}]_\wedge)\in\mathcal{G_X}$ with $\vect{\zeta}_\text{fus}\sim\mathcal{N}(\vect{0}_{\mathfrak{m}\times1},\vect{P}_\text{fus})$  from a given set of $n$ consistent and correlated stochastic members  $\vect{\mathcal{X}}_i=\Bvect{\mathcal{X}}_i\exp([\vect{\zeta}_i]_\wedge)\in\mathcal{G_X}$  with $\vect{\zeta}_i\sim\mathcal{N}(\vect{0}_{\mathfrak{m}\times 1},\vect{P}_{ii})$.  
The optimal fused member can be obtained from the following minimization problem~\cite{zarei2024consistent}:
\begin{align}\label{eq: opt problem 1}
 \vect{\mathcal{X}}^\star_\text{fus}=\arg \min_{\vect{\mathcal{X}}_\text{fus}}\begin{bmatrix}
 \vect{\varepsilon}_1\\
 \vdots\\
  \vect{\varepsilon}_n
\end{bmatrix}^\top\begin{bmatrix}
        \vect{P}_{11}&\cdots&\vect{P}_{1n}\\\vdots&\ddots&\vdots\\\vect{P}_{1n}^\top&\cdots&\vect{P}_{nn}\end{bmatrix}^{-1}\begin{bmatrix}
 \vect{\varepsilon}_1\\
 \vdots\\
  \vect{\varepsilon}_n
\end{bmatrix}, \end{align}
where the error between the $i^{th}$ known member and the fused member is 
$\vect{\varepsilon}_i\!=\!\left[\log(\Bvect{\mathcal{X}}_i^{-1}\vect{\mathcal{X}}_\text{fus})\right]_\vee\in\mathbb{R}^{\mathfrak{m}}$.
The closed-form solution to this minimization problem is presented in~\cite{zarei2024consistent}. For brevity, we only present the final derivation, as follows:
\begin{align}\label{eq: fused track final}
\Bvect{\mathcal{X}}_\text{fus}&=\Bvect{\mathcal{X}}_s\exp(\left[\Bvect{\zeta}_\text{fus}^\star\right]_\wedge),
\\ \label{eq: fused cov final}\vect{P}_\text{fus}&=\mathcal{J}_r(\Bvect{\zeta}_\text{fus}^\star)\vect{P}_{\vect{\zeta}\vect{\zeta}}^\star\mathcal{J}_r^\top(\Bvect{\zeta}_\text{fus}^\star).
\end{align}
Here, $\Bvect{\mathcal{X}}_s$ is an arbitrary reference member and
\begin{align}\nonumber
  \Bvect{\zeta}^\star_\text{fus}&=-\Big(\sum_{i=1}^N \sum_{j=1}^N  \vect{\mathcal{J}}_r^{-\top}(\vect{\zeta}_j)\vect{G}_{ij}^\top\vect{\mathcal{J}}_r^{-1}(\vect{\zeta}_i)\\\nonumber&~~~~~~~+  \vect{\mathcal{J}}_r^{-\top}(\vect{\zeta}_i)\vect{G}_{ij}\vect{\mathcal{J}}_r^{-1}(\vect{\zeta}_j)\Big)^{-1}\\&\sum_{i=1}^N \sum_{j=1}^N\vect{\mathcal{J}}_r^{-\top}(\vect{\zeta}_j)\vect{G}_{ij}^\top\vect{\zeta}_i+\vect{\mathcal{J}}_r^{-\top}(\vect{\zeta}_i)\vect{G}_{ij}\vect{\zeta}_j,\\\nonumber
\vect{P}_{\vect{\zeta}\vect{\zeta}}^\star&= \Big(\sum_{i=1}^N \sum_{j=1}^N  \vect{\mathcal{J}}_r^{-\top}(\vect{\zeta}_j)\vect{G}_{ij}^\top\vect{\mathcal{J}}_r^{-1}(\vect{\zeta}_i)\\&~~~~~+  \vect{\mathcal{J}}_r^{-\top}(\vect{\zeta}_i)\vect{G}_{ij}\vect{\mathcal{J}}_r^{-1}(\vect{\zeta}_j)\Big)^{-1},  
\end{align}
 where $\vect{\zeta}_i=\left[\log(\Bvect{\mathcal{X}}_i^{-1}\Bvect{\mathcal{X}}_s)\right]_\vee$ and
\begin{align}\nonumber
    \vect{G}\!=\!\left[\begin{smallmatrix}
        \vect{P}_{11}&\cdots&\vect{P}_{1n}\\\vdots&\ddots&\vdots\\\vect{P}_{1n}^\top&\cdots&\vect{P}_{nn}\end{smallmatrix}\right]^{-1}=\left[\begin{smallmatrix}
        \vect{G}_{11}&\cdots&\vect{G}_{1N}\\
        \vdots&\ddots&\vdots\\\vect{G}_{N1}&\cdots&\vect{G}_{NN}
    \end{smallmatrix}\right]\in\mathbb{R}^{\mathfrak{m}n\times \mathfrak{m}n}.
\end{align}

Furthermore, in an EKF-based estimation framework, the cross-covariance matrix between correlated estimates is calculated from the following recursion~\cite{zarei2024consistent}:  
\begin{align*}
&\nonumber\vect{P}_{ij}{\scriptstyle (k|k)}\!=\!
\big(\vect{I}_{\mathfrak{m}}\!-\!\vect{K}_i{\scriptstyle (k)}\vect{\mathcal{H}}_i{\scriptstyle (k)}\big)\Big(\vect{\mathcal{F}}_i{\scriptstyle(k-1)}\vect{P}_{ij}{\scriptstyle( k-1|k-1)}\vect{\mathcal{F}}_j^\top{\scriptstyle( k-1)}\\&+\vect{\mathcal{J}}_r(\Bvect{f}_i{\scriptstyle( k-1)})\vect{Q}{\scriptstyle( k-1)}\vect{\mathcal{J}}_r^\top(\Bvect{f}_j{\scriptstyle( k-1)})\!\Big)\big(\vect{I}_{\mathfrak{m}}\!-\!\vect{K}_j{\scriptstyle (k)}\vect{\mathcal{H}}_j{\scriptstyle (k)}\big)\!^\top.
\end{align*}

\subsection{Constrained Fusion}
In some  frameworks, the estimated member is required to lie on a constrained surface. In this section, we propose a method to project the unconstrained fused estimate $\Bvect{\mathcal{X}}_u=\Bvect{\mathcal{X}}_\text{fus}^\star$ onto
a constraint space defined by $\left\{\Bvect{\mathcal{X}}_c\in\mathcal{G_X}|\vect{g}(\Bvect{\mathcal{X}}_c)=\vect{D}\right\}$. Here, $\Bvect{\mathcal{X}}_c$ is the constrained estimate, $\vect{g}:\mathcal{G_X}\rightarrow\mathcal{G_X}_c$ is a map to the constraint matrix Lie group $\mathcal{G_X}_c$ with the dimension $r\leq \mathfrak{m}$, and $\vect{D}\in\mathcal{G_X}_c$ is a constant member. To this end, we solve the following minimization problem:  
 \begin{align}\label{eq: const track fusion}\nonumber
     \Bvect{\mathcal{X}}_c^\star&=\arg \min_{\Bvect{\mathcal{X}}_c}([\log(\Bvect{\mathcal{X}}_u^{-1}\Bvect{\mathcal{X}}_c)]_\vee)^\top\vect{W}_c([\log(\Bvect{\mathcal{X}}_u^{-1}\Bvect{\mathcal{X}}_c)]_\vee),\\ \vect{D}&=\vect{g}(\Bvect{\mathcal{X}}_c),   
    \end{align}
where,  $\vect{W}_c\in\mathbb{R}^{\mathfrak{m}\times \mathfrak{m}}$ is any symmetric positive definite  weighting matrix.
To solve this problem, we define a reference member $\Bvect{\mathcal{X}}_s$ and write $\Bvect{\mathcal{X}}_c=\Bvect{\mathcal{X}}_s\exp([\vect{\zeta}_c]_\wedge)$. We then find the optimum value for the unknown error $\vect{\zeta}_c$ on Lie algebra $\mathfrak{g}_{\mathcal{X}}$ using the known vector $\vect{\zeta}_u=[\log(\Bvect{\mathcal{X}}_u^{-1}\Bvect{\mathcal{X}}_s)]_\vee$. Assuming $\vect{\zeta}_c$ is small and using~\eqref{eq: BCH 52}, the minimization problem  reduces to
\begin{align}\label{eq: tau constraint}\nonumber
 \vect{\zeta}_c^\star&=\arg\min_{\vect{\zeta}_c}  (\vect{\zeta}_u+\mathcal{J}^{-1}_r(\vect{\zeta}_u)\vect{\zeta}_c)^\top\vect{W}_c(\vect{\zeta}_u+\mathcal{J}^{-1}_r(\vect{\zeta}_u)\vect{\zeta}_c),\\
 \Tvect{d}&=\Tvect{g}(\vect{\zeta}_c),
\end{align}
where,  $\Tvect{d}=[\log(\vect{D})]_\vee$ and $\Tvect{g}(\vect{\zeta}_c)=[\log(\vect{g}(\Bvect{\mathcal{X}}_s\exp([\vect{\zeta}_c]_\wedge)))]_\vee$ are vectors in $\mathbb{R}^r$, $r\leq  \mathfrak{m}$. The
constrained minimization problem~\eqref{eq: tau constraint} can be solved using the Lagrange multiplier method~\cite{moon2000mathematical}. The Lagrangian is defined as:
\begin{align}\label{eq: Lagrangian}
    \mathcal{L}\!=\!(\vect{\zeta}_u\!+\!\mathcal{J}^{-1}_r\!(\vect{\zeta}_u\!)\vect{\zeta}_c)^{\!\top}\vect{W}_c(\vect{\zeta}_u\!+\!\mathcal{J}^{-1}_r\!(\vect{\zeta}_u)\vect{\zeta}_c)\!+\!2\vect{\lambda}^{\!\top}(\Tvect{g}(\vect{\zeta}_c\!)\!-\!\Tvect{d}),
\end{align}
where, $\vect{\lambda}\in\mathbb{R}^r$ is the vector of Lagrange multipliers. We then find the necessary conditions for the minimum, which are obtained by solving the following set of algebraic equations:
\begin{align*}
  &\frac{\partial \mathcal{L}}{\partial\vect{\zeta}_c}\!=\!\mathcal{J}^{-\top}_r({\vect{\zeta}_u})\vect{W}_c(\vect{\zeta}_u\!+\!\mathcal{J}^{-1}_r(\vect{\zeta}_u)\vect{\zeta}_c)\!+\!\frac{\partial\Tvect{g}^\top}{\partial\vect{\zeta}_c}(\vect{\zeta}_c)\vect{\lambda}\!=\!\vect{0}_{m\times 1},\\
  &\frac{\partial \mathcal{L}}{\partial\vect{\lambda}}=\Tvect{g}(\vect{\zeta}_c)-\Tvect{d}=\vect{0}_{r\times 1}.
\end{align*}
Here, we solve this system for the specific case where the constraint function is linear in $\vect{\zeta}_c$, i.e., $\Tvect{g}(\vect{\zeta}_c)=\vect{\Gamma}\vect{\zeta}_c$ where $\vect{\Gamma}\in\mathbb{R}^{r\times \mathfrak{m}},~r<\mathfrak{m}$. We also assume the constraints are linearly independent, i.e.,  $\vect{\Gamma}$ has rank $r$. In this case the closed form solution is 
\begin{align}\label{eq: const tc333}\nonumber
    \vect{\zeta}_c^\star&=\vect{\mathfrak{J}}\Big[\Big(\vect{\Gamma}^\top(\vect{\Gamma}\vect{\mathfrak{J}}\vect{\Gamma}^\top)^{-1}\vect{\Gamma}\vect{\mathfrak{J}}\mathcal{J}_r^{-\top}(\vect{\zeta}_u)\vect{W}_c\\&-\mathcal{J}_r^{-\top}(\vect{\zeta}_u)\vect{W}_c\Big)\vect{\zeta}_u+\vect{\Gamma}^\top(\vect{\Gamma}\vect{\mathfrak{J}}\vect{\Gamma}^\top)^{-1}\Tvect{d}\Big],
\end{align}
where, $\vect{\mathfrak{J}}=\big(\mathcal{J}_r^{-\top}({\vect{\zeta}_u})\vect{W}_c\mathcal{J}_r^{-1}({\vect{\zeta}_u})\big)^{-1}$. Note that if $r=\mathfrak{m}$, then~\eqref{eq: const tc333} reduces to $\vect{\zeta}_c^\star=\vect{\Gamma}^{-1}\Tvect{d}$. The derivation process of~\eqref{eq: const tc333} is detailed in Appendix~\ref{app: const}.




%

\section{Localization Framework}
In this section, we use the stochastic operations defined in Section \ref{sec: stochastic configuration} to propose a collaborative localization system for a group of $n$ robots operating under the following conditions:
\begin{itemize}
\item The robots operate in a GPS-denied indoor environment and they are equipped with inertial sensors (e.g., IMU), velocity sensors (e.g., wheel encoders), and pose estimation modules (e.g., marker-based). 

\item They are equipped with communication devices such as WiFi that enable inter-robot communication.

\item The robots are classified into three types: leader, which  obtains global pose information from stationary markers; follower type 1, which communicates with the leader and obtains relative pose information from the leader's mobile markers; and follower type 2, which communicates with a follower (either type 1 or 2) and obtains relative pose information from its mobile markers. The follower type is dynamic, i.e., a robot may be follower type 1 or 2, depending on whether a leader is in its field of view.

\item Each robot runs a local EKF, in which inertial sensor measurements are used for prediction, while velocity and pose measurements are used for update.

\end{itemize}

\subsection{Inertial Prediction}
Given an inertial sensor such as IMU within a body coordinate frame $\{B\}$ moving relative to the robot's global coordinate frame 
$\{O\}$, and assuming its readings are solely affected by noise and bias errors, the measured angular velocity 
$\Tvect{\omega}^{OB}$  (from gyros of IMU) and acceleration 
$\Tvect{a}_B^O$
  (from accelerometers of IMU) are
\begin{align}
    \Tvect{\omega}^{OB}&=\vect{\omega}^{OB}+\vect{b}_{g}+\vect{w}_g,\\
    \Tvect{a}_B^O&=\vect{a}_B^O+\vect{b}_a+\vect{w}_a.
\end{align}
Here, $\vect{\omega}^{OB}\in\mathbb{R}^3$ denotes the true angular velocity of frame $\{B\}$ relative to frame $\{O\}$, while $\vect{a}_B^O\in\mathbb{R}^3$ is the true acceleration of point $B$ relative to frame $\{O\}$ and expressed in the frame $\{B\}$. The variables $\vect{b}_g\in\mathbb{R}^3$ and $\vect{b}_a\in\mathbb{R}^3$ denote the biases of the gyros and accelerometers, while $\vect{w}_g\sim\mathcal{N}(\vect{0}_{3\times1},\vect{Q}_g)\in\mathbb{R}^3$ and $\vect{w}_a\sim\mathcal{N}(\vect{0}_{3\times1},\vect{Q}_a)\in\mathbb{R}^3$ are zero-mean Gaussian white noise processes with covariances $\vect{Q}_g\in\mathbb{R}^{3\times3}$ and $\vect{Q}_a\in\mathbb{R}^{3\times3}$, respectively. 
Therefore, the kinematic motion of frame $\{B\}$ relative to frame $\{O\}$ yields 
\begin{align}
\dot{\vect{\mathcal{R}}}^{OB}\!&=\vect{\mathcal{R}}^{OB}\!\big[\Tvect{\omega}^{OB}-\vect{b}_{g}-\vect{w}_g\big]_{\times},\\
\dot{\vect{v}}_B^O{\scriptstyle(t)}&=\vect{\mathcal{R}}^{OB}(\Tvect{a}_B^O-\vect{b}_a-\vect{w}_a)+\vect{g},\\
\dot{\vect{p}}_B^O&=\vect{v}_B^O,\\
\dot{\vect{b}}_g&=\vect{w}_{b_g},\\
\dot{\vect{b}}_a&=\vect{w}_{b_a}.
\end{align}
Here, $\vect{\mathcal{R}}^{OB}\in\mathbb{SO}(3)$ is the rotation matrix of frame $\{B\}$ relative to frame $\{O\}$, while $\vect{v}_B^O\in\mathbb{R}^3$ and $\vect{p}_B^O\in\mathbb{R}^3$ respectively denote the velocity and position of point $B$ relative to frame $\{O\}$ and expressed in frame $\{O\}$. The gravity vector in the global frame is denoted as $\vect{g}\in\mathbb{R}^3$ while the biases are modeled as random walk processes with the Gaussian zero-mean noises $\vect{w}_{b_g}\sim\mathcal{N}(\vect{0}_{3\times1},\vect{Q}_{b_g})$ and $\vect{w}_{b_a}\sim\mathcal{N}(\vect{0}_{3\times1},\vect{Q}_{b_a})$. The matrix Lie group $\mathcal{G}_{\mathcal{X}}\coloneq\mathbb{SE}_2(3)\times\mathbb{R}^3\times\mathbb{R}^3$ is proposed here (Appendix~\ref{app: lie grop GX}) to capture the kinematics of motion by
\begin{align}
    \dot{\vect{\mathcal{X}}}=\vect{\mathcal{X}}\big[\vect{f}\big(\vect{\mathcal{X}},\vect{u}\big)+\vect{w}\big]_\wedge,
\end{align}
whose members are represented as
\begin{align}\nonumber
\vect{\mathcal{X}}\!=\!\begin{bmatrix}
   \begin{bmatrix}
       \vect{\mathcal{R}}^{OB}&\vect{p}_B^O&\vect{v}_B^O\\\vect{0}_{1\times3}&1&0\\\vect{0}_{1\times3}&0&1
   \end{bmatrix}&\vect{0}_{5}\\\vect{0}_{5}&\begin{bmatrix} \vect{I}_{3}&\vect{b}_g&\vect{b}_a\\\vect{0}_{1\times3}&1&0\\\vect{0}_{1\times3}&0&1\end{bmatrix}
\end{bmatrix}\! \in\mathbb{R}^{10\times10},
\end{align}
where $\vect{f}\big(\vect{\mathcal{X}},\vect{u}\big)$ and $\vect{w}$ denote the generalized velocity and  the noise vectors, respectively. The matrix $\big[\vect{f}\big(\vect{\mathcal{X}},\vect{u}\big)+\vect{w}\big]_\wedge=\vect{\mathcal{X}}^{-1}\dot{\vect{\mathcal{X}}}\in\mathfrak{g}_{\mathcal{X}}$, where $\mathfrak{g}_{\mathcal{X}}$ denotes the Lie algebra of $\mathcal{G_X}$ and its members are in the following form
\begin{align}\nonumber
{ \setlength{\arraycolsep}{1.75pt} 
\begin{bmatrix}
   \left[\begin{smallmatrix}
       [\Tvect{\omega}^{O\!B}\!-\!\vect{b}_{g}\!-\!\vect{w}_g]_\times&\vect{\mathcal{R}}^{{O\!B}\!^\top}\!\!\vect{v}_B^O&\Tvect{a}_B^O\!-\!\vect{b}_a\!-\!\vect{w}_a\!+\!\vect{\mathcal{R}}^{{O\!B}\!^\top}\!\!\vect{g}\\\vect{0}_{1\times3}&0&0\\\vect{0}_{1\times3}&0&0\end{smallmatrix}\right]&\vect{0}_5\\\vect{0}_5&\left[\begin{smallmatrix} \vect{0}_{3}&\vect{w}_{b_g}&\vect{w}_{b_a}\\\vect{0}_{1\times3}&0&0\\\vect{0}_{1\times3}&0&0\\\vect{0}_{1\times3}&0&0\end{smallmatrix}\right]
\end{bmatrix}\!. 
}
\end{align}

Let the input vector be $\vect{u}=\begin{bmatrix}
    \Tvect{\omega}^{{OB}^\top}&\Tvect{a}_B^{O^\top}
\end{bmatrix}^\top$, then 
\begin{align}
&\vect{f}\big(\vect{\mathcal{X}},\vect{u}\big)\!=\!\begin{bmatrix}
    \Tvect{\omega}^{OB}\!-\!\vect{b}_g\\\vect{\mathcal{R}}^{{OB}^\top}\vect{v}_B^O\\\Tvect{a}_B^O\!-\!\vect{b}_a\!+\!\vect{\mathcal{R}}^{{OB}^\top}\vect{g}\\\vect{0}_{3\times1}\\\vect{0}_{3\times1}\\\vect{0}_{3\times1}
\end{bmatrix}  \in\mathbb{R}^{18},
\end{align}
\begin{align}
\vect{w}=\begin{bmatrix}
    -\vect{w}_g\\\vect{0}_{3\times1}\\-\vect{w}_a\\\vect{0}_{3\times1}\\\vect{w}_{b_g}\\\vect{w}_{b_a}
\end{bmatrix}\equiv\begin{bmatrix}
    \vect{w}_g\\\vect{0}_{3\times1}\\\vect{w}_a\\\vect{0}_{3\times1}\\\vect{w}_{b_g}\\\vect{w}_{b_a}
\end{bmatrix}\sim\mathcal{N}(\vect{0}_{18\times 1},\vect{Q}),
\end{align}
where, the process noise covariance matrix is $\vect{Q}=\operatorname{diag}([\vect{Q}_g,\vect{0}_{3\times3},\vect{Q}_a,\vect{0}_{3\times3},\vect{Q}_{b_g},\vect{Q}_{b_a}])\in\mathbb{R}^{18\times18}$.

The Euler method is then employed to perform discretization
of this continuous-time stochastic system:
\begin{align}
\vect{\mathcal{X}}\!{\scriptstyle (k+1)}\!=\!\vect{\mathcal{X}}\!{\scriptstyle (k)}\exp\big(\!\Delta t [\vect{f}({ \vect{\mathcal{X}}\!\scriptstyle(k)},\!\vect{u}{\scriptstyle(k)})]_\wedge\!+\!\sqrt{\Delta t}[\vect{w}{\scriptstyle (k)}]_\wedge\big),
\end{align}  
where, $\Delta t=t_{k+1}-t_k$ represents the discretization interval, and $\vect{w}{\scriptstyle (k)}\sim\mathcal{N}(\vect{0}_{18\times1},\vect{Q})$ captures inertial sensor's (IMU) uncertainties, including scale factor variations, misalignment, cross-axis coupling, and temperature-induced errors. 

\subsection{Velocity Update }
  Integrating  velocity measurements with inertial measurements compensates for the limitations of these sensors and provides more robust and accurate estimations. The measurement model based on the body velocity is formulated on the manifold $\mathbb{R}^3$ as follows:
 \begin{align}
     \vect{\mathcal{Z}}_v{\scriptstyle(k)}=\vect{h}(\vect{\mathcal{X}}{\scriptstyle(k)})=\vect{\mathcal{R}}^{{OB}^\top}\!{\scriptstyle(k)}\vect{v}_B^O{\scriptstyle(k)}+\vect{m}_v{\scriptstyle(k)}.
 \end{align}
The noise signal  $\vect{m}_v{\scriptstyle(k)}\sim\mathcal{N}(\vect{0}_{3\times1},\vect{R}_v)$ represents the uncertainties associated with the velocity sensors with $\vect{R}_v\in\mathbb{R}^{3\times3}$ denoting the covariance matrix of this noise. If the velocity update is performed using a Wheel Odometry (WO), the uncertainties will correspond to factors such as wheel slippage, uneven terrain, wheel deformation, inaccuracies in the  wheel radius and wheelbase, and limited encoder resolution.

\subsection{Pose Update }\label{sec: pose update fid mark}
In this section, we perform pose updates within each EKF using appropriate exteroceptive sensors, such as cameras. While our focus is on the use of fiducial markers for pose estimation, the proposed method is general and not limited to a specific type of pose measurement. Unlike the previous sections, which utilize proprioceptive sensors for inertial predictions and velocity updates, the approach here employs exteroceptive sensing to construct pseudo-pose measurements tailored to the specific role of each robot. We consider three categories of robots within the network: a leader, follower type 1, and follower type 2. To distinguish between the measurements and their models associated with each type, we introduce indexing notations $l$, $f_1$, and $f_2$, respectively. The subsequent sections provide a detailed methodology for generating pseudo-pose measurements for each robot type and for updating the EKFs accordingly on the state space $\mathcal{G_Z} \coloneq \mathbb{SE}(3) \times \mathbb{R}^3$.

\subsubsection{Pose Update for Leaders}
\begin{figure}[htbp]
  \unitlength=0.5in
  \centering 
\subfloat[]{%
    \scalebox{0.85}{\includegraphics[width=9cm]{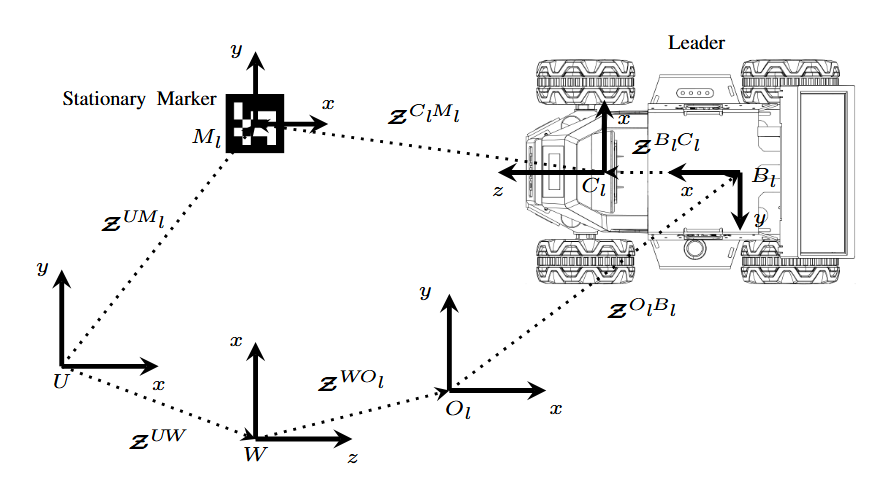}}
  }
    \\
  \subfloat[]{\scalebox{0.85}{\includegraphics[width=9cm]{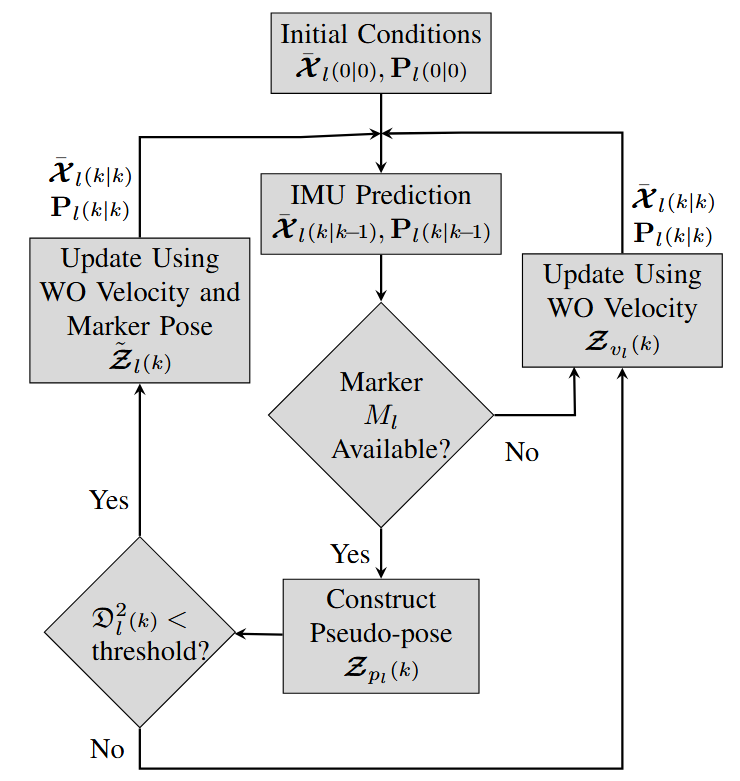}}}
  \caption{A leader robot's (a) coordinate frames, and (b) estimation framework. }
  \label{fig: leader}
\end{figure}

To construct the pseudo-pose of a leader using stationary landmarks, we first establish the relevant coordinate frames as shown in Fig.~\ref{fig: leader}-(a). We consider a global, user-defined coordinate frame, denoted as $\{U\}$, within which the stationary landmarks' poses are known. The leader robot's global coordinate frame, $\{O_l\}$, serves as the reference for the robot's inertial and velocity sensors (IMU and WO), as conventionally set on the onboard computer. The robot's body frame is denoted as $\{B_l\}$ while the camera coordinate frame on the leader is $\{C_l\}$. Finally, $\{M_l\}$ serves as the landmark's coordinate frame. To follow the ground truth motion of the robots in Section~\ref{sec:exp and results}, we also define the coordinate frame of a motion tracker system, denoted as $\{W\}$, which has a fixed offset relative to $\{U\}$. 

To determine a leader's pseudo-pose in the form of
${\vect{\mathcal{Z}}_{p_l}}\coloneq\vect{\mathcal{Z}}^{O_l\!B_l}=\Big[\begin{smallmatrix}
    \vect{\mathcal{R}}^{O_l\!B_l}_p&\vect{p}_{{B_l}_p}^{O_l}\\\vect{0}_{1\times3}&1 
\end{smallmatrix}\Big]\in\mathbb{SE}(3)$, we use the relation:
\begin{align}
 \vect{\mathcal{Z}}^{O_l\!B_l}\!=\!(\vect{\mathcal{Z}}^{U\!W}\!\vect{\mathcal{Z}}^{W\!O_l})\!^{-1}\vect{\mathcal{Z}}^{U\!M_l}(\vect{\mathcal{Z}}^{B_l\!C_l}\vect{\mathcal{Z}}^{C_l\!M_l})^{-1}\!, 
\end{align}
where throughout this paper, we denote the relative pose between two generic frames $\{X\}$ and $\{Y\}$ by $\vect{\mathcal{Z}}^{XY}=\Big[\begin{smallmatrix}
    \vect{\mathcal{R}}^{XY}&\vect{p}_{Y}^{X}\\\vect{0}_{1\times3}&1 
\end{smallmatrix}\Big]\in\mathbb{SE}(3)$. 
 The members $\vect{\mathcal{Z}}^{U\!W},\vect{\mathcal{Z}}^{W\!O_l},\vect{\mathcal{Z}}^{U\!M_l}$ and $\vect{\mathcal{Z}}^{B_l\!C_l}$ are known and deterministic, while $\vect{\mathcal{Z}}^{C_l\!M_l}$ represents the stochastic member provided by camera measurements. These measurements are corrupted by noise and we assume their associated uncertainty follows a zero-mean Gaussian process. Thus, the leader's camera measurements  are  $\vect{\mathcal{Z}}^{C_l\!M_l}=\Bvect{\mathcal{Z}}^{C_l\!M_l}\exp(\left[\vect{\zeta}_{\!M_l}^{C_l}\right]_\wedge)$ with $\vect{\zeta}_{\!M_l}^{C_l}\sim\mathcal{N}(\vect{0}_{1\times6},\vect{R}_{m_l})$ and noise covariance $\vect{R}_{m_l}\in\mathbb{R}^{6\times6}$. We then  apply the operations developed in Section~\ref{sec: stochastic configuration} to derive the mean and covariance matrix of $\vect{\mathcal{Z}}_{p_l}$ as follows:
\begin{align}
 {\Bvect{\mathcal{Z}}_{p_l}}&=(\vect{\mathcal{Z}}^{U\!W}\!\vect{\mathcal{Z}}^{W\!O_l})^{-1}\vect{\mathcal{Z}}^{U\!M_l}(\vect{\mathcal{Z}}^{B_l\!C_l}\Bvect{\mathcal{Z}}^{C_l\!M_l})^{-1},\\
 {\vect{R}_{p_l}}&\approx \vect{Ad}_{\vect{\mathcal{Z}}^{B_l\!C_l}}\vect{Ad}_{\Bvect{\mathcal{Z}}^{C_l\!M_l}}\vect{R}_{m_l}\vect{Ad}^\top_{\Bvect{\mathcal{Z}}^{C_l\!M_l}}\vect{Ad}^\top_{\vect{\mathcal{Z}}^{B_l\!C_l}}.
\end{align}
When both velocity measurements ${\vect{\mathcal{Z}}_{v_l}}{\scriptstyle(k)}=\vect{v}_{{B_l}_v}^{O_l}\!{\scriptstyle(k)}$ and  pseudo-pose measurements $\vect{\mathcal{Z}}_{p_l}{\scriptstyle(k)}$ are available and reliable (refer to Section~\ref{sec: FDM} for sensor fault detection), the update is done on $\mathcal{G}_{\mathcal{Z}}\coloneq\mathbb{SE}(3)\times\mathbb{R}^3$ whose members are in the following form:
\begin{align*}
    \Tvect{\mathcal{Z}}_l{\scriptstyle(k)}\!=\!\begin{bmatrix}\!
   \begin{bmatrix}
       \vect{\mathcal{R}}^{O_l\!B_l}_p\!{\scriptstyle(k)}&\vect{p}_{{B_l}_p}^{O_l}\!{\scriptstyle(k)}\\\vect{0}_{1\times3}&1
   \end{bmatrix}&\vect{0}_4\\\vect{0}_4&\begin{bmatrix} \vect{I}_{3}&\vect{v}_{{B_l}_v}^{O_l}\!{\scriptstyle(k)}\\\vect{0}_{1\times3}&1\!\end{bmatrix}
\end{bmatrix}\!\in\!\mathbb{R}^{8\!\times\!8}.
\end{align*}
 Details of the matrix Lie group $\mathcal{G}_{\mathcal{Z}}$ are provided in Appendix~\ref{app: lie grop GX}. Thus, the measurement model is as follows:
\begin{align*}
    \vect{\mathcal{Z}}_l{\scriptstyle(k)}\!=\!\begin{bmatrix}\!
   \left[\begin{smallmatrix}
       \vect{\mathcal{R}}^{O_l\!B_l}\!{\scriptstyle(k)}&\vect{p}_{{B_l}}^{O_l}\!{\scriptstyle(k)}\\\vect{0}_{1\times3}&1
   \end{smallmatrix}\right]&\vect{0}_4\\\vect{0}_4&\left[\begin{smallmatrix} \vect{I}_{3}& \vect{\mathcal{R}}^{{O_l\!B_l}^\top}\!{\scriptstyle(k)}\vect{v}_{{B_l}}^{O_l}\!{\scriptstyle(k)}\\\vect{0}_{1\times3}&1\!\end{smallmatrix}\right]
\!\end{bmatrix}\!\exp(\left[{\vect{m}_l{\scriptstyle (k)}}\right]_{\wedge}\!),
\end{align*}
where, ${\vect{m}_l{\scriptstyle (k)}}$ is a zero-mean noise vector with the noise covariance matrix $\vect{R}_l=\operatorname{diag}([{\vect{R}_{p_l}},\epsilon\vect{I}_3,{\vect{R}_{v_l}}])\in\mathbb{R}^{12\times12}$, and $\epsilon\ll 1$ to avoid inversion of zero.

\subsubsection{Pose Update for Follower Type 1}\label{sec: leader-follower update}
\begin{figure}[htbp]
  \centering 
\subfloat[]{%
    \scalebox{0.85}{\includegraphics[width=9cm]{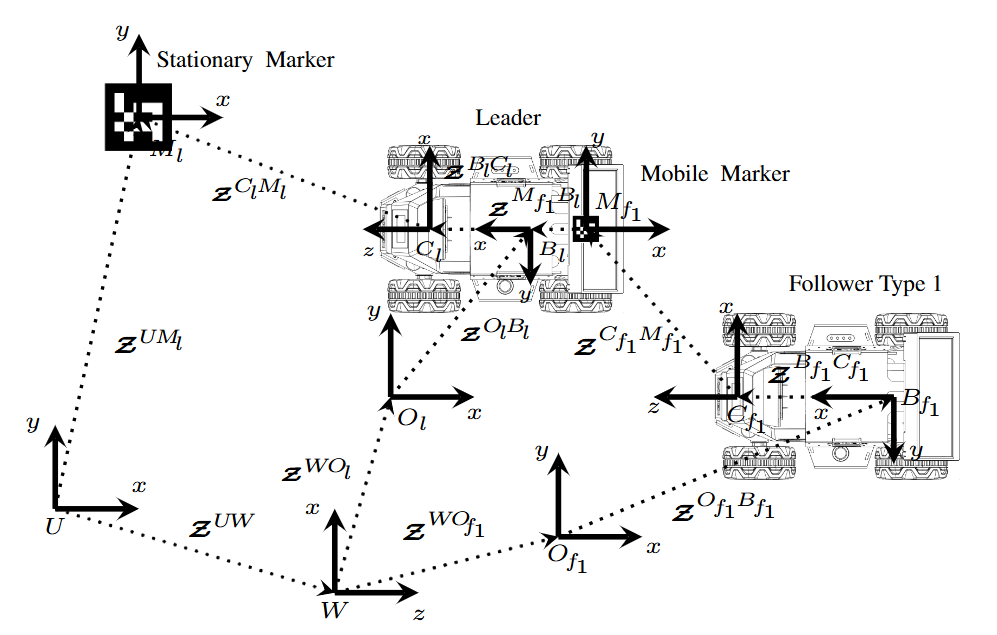}}
  }
    \\
    \subfloat[]{%
    \scalebox{0.85}{\includegraphics[width=9cm]{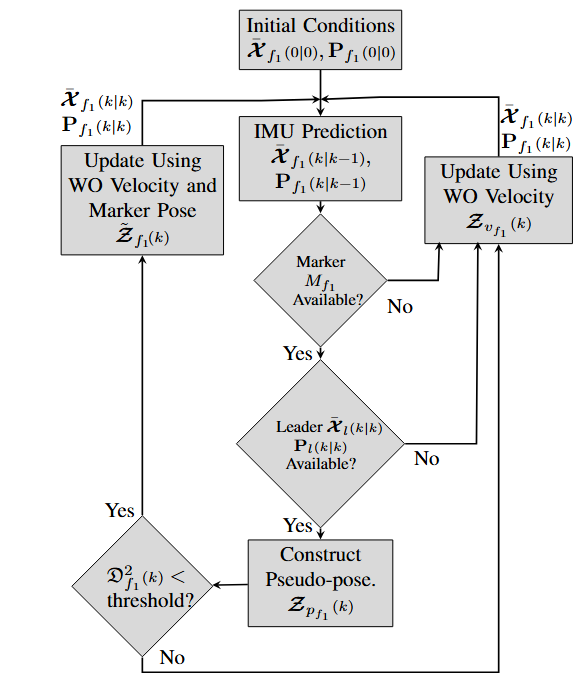}}
  }
  \caption{Follower type 1 robot, (a) coordinate frames, (b) estimation framework.}
  \label{fig: follower 1}
\end{figure}

Here, we introduce additional coordinate frames for the follower type 1 robot including the global frame 
$\{O_{f_1}\}$, body frame 
$\{B_{f_1}\}$, camera frame 
$\{C_{f_1}\}$, and mobile landmark (fiducial marker) frame on the leader and observed by the follower type 1
 $\{M_{f_1}\}$. As shown in Fig.\ref{fig: follower 1}-(a), there are two paths to construct $\vect{\mathcal{Z}}^{O_{f_1}\!B_{f_1}}$. The first approach relies on stochastic members $\vect{\mathcal{Z}}^{C_{f_1}\!M_{f_1}}$ and $\vect{\mathcal{Z}}^{C_l\!M_l}$ —the raw observations of the mobile and stationary markers $M_{f_1}$ and $M_l$, respectively.
The second path relies on the estimated pose of the leader robot and the follower's relative pose observation of the mobile marker $M_{f_1}$, i.e.,
 \begin{align*}
 \vect{\mathcal{Z}}^{O\!_{f_1}\!B\!_{f_1}}\!=\!(\!\vect{\mathcal{Z}}^{W\!O\!_{f_1}}\!)^{-1}\vect{\mathcal{Z}}^{W\!O_l}\!\vect{\mathcal{Z}}^{O_l\!B_l}(\!\vect{\mathcal{Z}}^{B\!_{f\!_1}\!C\!_{f\!_1}}\vect{\mathcal{Z}}^{C\!_{f\!_1}\!M\!_{f\!_1}}\vect{\mathcal{Z}}^{M\!_{f\!_1}\!B\!_l}\!)^{-1}\!,  
\end{align*}
where we use
$\vect{\mathcal{Z}}^{O_l\!B_l}=\Bvect{\mathcal{X}}_l^{\mathbb{SE}(3)}\exp(\left[\vect{\zeta}_l^{\mathbb{SE}(3)}\right]_\wedge)$ with $\vect{\zeta}_{l}^{\mathbb{SE}(3)}\sim\mathcal{N}(\vect{0}_{1\times6},\vect{P}_l^{\mathbb{SE}(3)})$ and $\vect{P}_l^{\mathbb{SE}(3)}\in\mathbb{R}^{6\times6}$. Note that $\Bvect{\mathcal{X}}_l^{\mathbb{SE}(3)}$ is the pose part of the current estimated state of the leader in its local EKF and $\vect{\zeta}_l^{\mathbb{SE}(3)}$ is its associated uncertainty. This method's advantage lies in using the leader robot's estimated pose, which is available at a higher frequency and can be transmitted more quickly than raw marker measurements. We then let $\vect{\mathcal{Z}}^{C_{f_1}\!M_{f_1}}=\Bvect{\mathcal{Z}}^{C_{f_1}\!M_{f_1}}\exp(\left[\vect{\zeta}_{\!M_{f_1}}^{C_{f_1}}\right]_\wedge)$ with $\vect{\zeta}_{\!M_{f_1}}^{C_{f_1}}\sim\mathcal{N}(\vect{0}_{1\times6},\vect{R}_{m_{f_1}})$ with $\vect{R}_{m_{f_1}}\in\mathbb{R}^{6\times6}$ to obtain the mean and covariance of the stochastic member $\vect{\mathcal{Z}}_{{p_{f\!_1}}}\!\coloneqq \!\vect{\mathcal{Z}}^{O\!_{f\!_1}\!B\!_{f\!_1}}=\Big[\begin{smallmatrix}
    \vect{\mathcal{R}}^{O\!_{f\!_1}\!B\!_{f\!_1}}_p&\vect{p}_{{B\!_{f\!_1}}_p}^{O\!_{f\!_1}}\\\vect{0}_{1\times3}&1 
\end{smallmatrix}\Big]\in\mathbb{SE}(3)$  as follows:
\begin{align*}
 \Bvect{\mathcal{Z}}_{p_{f_1}}\!&=\!(\!\vect{\mathcal{Z}}^{W\!O_{f_1}}\!)^{-1}\!\vect{\mathcal{Z}}^{W\!O_l}\!\Bvect{\mathcal{X}}_{l}^{\mathbb{SE}(3)}(\vect{\mathcal{Z}}^{B_{f_1}\!C_{f_1}}\!\Bvect{\mathcal{Z}}^{C_{f_1}\!M_{f_1}}\!\vect{\mathcal{Z}}^{M_{f_1}\!B_l})^{-1}\!,\\
    \vect{R}_{p_{f_1}}&\!\approx\!\vect{Ad}_{\vect{\mathcal{Z}}^{B\!_{f\!_1}\!C\!_{f\!_1}}}\vect{Ad}_{\Bvect{\mathcal{Z}}^{C\!_{f\!_1}\!M\!_{f\!_1}}}\vect{R}_{m_{f_1}}\vect{Ad}^\top_{\Bvect{\mathcal{Z}}^{C\!_{f\!_1}\!M\!_{f\!_1}}}\vect{Ad}^\top_{\vect{\mathcal{Z}}^{B\!_{f\!_1}\!C\!_{f\!_1}}}+\\&\vect{Ad}_{\vect{\mathcal{Z}}^{B\!_{f\!_1}\!C\!_{f\!_1}}\Bvect{\mathcal{Z}}^{C\!_{f\!_1}\!M\!_{f\!_1}}\Bvect{\mathcal{Z}}^{M\!_{f\!_1}\!B_l}}\vect{P}_{l}^{\mathbb{SE}(3)}\vect{Ad}^\top_{\vect{\mathcal{Z}}^{B\!_{f\!_1}\!C\!_{f\!_1}}\Bvect{\mathcal{Z}}^{C\!_{f\!_1}\!M\!_{f\!_1}}\Bvect{\mathcal{Z}}^{M\!_{f\!_1}\!B_l}}.
    \end{align*}
The next step updates the follower robot's predictions using its  velocity measurements ${\vect{\mathcal{Z}}_{v_{f_1}}}{\scriptstyle(k)}=\vect{v}_{{B_{f_1}}_v}^{O_{f_1}}\!{\scriptstyle(k)}$ and the constructed pseudo-pose measurements $\vect{\mathcal{Z}}_{p_{f_1}}$ on the manifold $\mathcal{G_Z}\coloneq\mathbb{SE}(3)\times\mathbb{R}^3$ provided that the constructed pseudo-poses are reliable (refer to Section~\ref{sec: FDM}). Thus, the measurement member is in the following form:
\begin{align*}
    \Tvect{\mathcal{Z}}\!_{f\!_1}\!{\scriptstyle(k)}\!=\!\begin{bmatrix}\!
   \begin{bmatrix}\!
       \vect{\mathcal{R}}^{O\!_{f\!_1}\!B\!_{f\!_1}}_p\!{\scriptstyle(k)}&\vect{p}_{{B\!_{f\!_1}}_p}^{O\!_{f\!_1}}\!{\scriptstyle(k)}\\\vect{0}_{1\times3}&1
   \!\end{bmatrix}&\vect{0}_4\\\vect{0}_4&\!\begin{bmatrix} \!\vect{I}_{3}&\vect{v}_{{B\!_{f\!_1}}_v}^{O\!_{f\!_1}}\!{\scriptstyle(k)}\\\vect{0}_{1\times3}&1\!\end{bmatrix}\!
\end{bmatrix}\!\in\!\mathbb{R}^{8\!\times\!8}\!.
\end{align*}
The covariance matrix of the measurement $\Tvect{\mathcal{Z}}\!_{f\!_1}$ is obtained from $\vect{R}_{f_1}=\operatorname{diag}([{\vect{R}_{p_{f_1}}},\epsilon\vect{I}_3,{\vect{R}_{v_{f_1}}}])\in\mathbb{R}^{12\times12}$.

\subsubsection{Pose Update for Follower Type 2}\label{sec: follower-follower}

\begin{figure}[htbp]
  \unitlength=0.5in
  \centering 
\subfloat[]{%
    \scalebox{0.85}{\includegraphics[width=9cm]{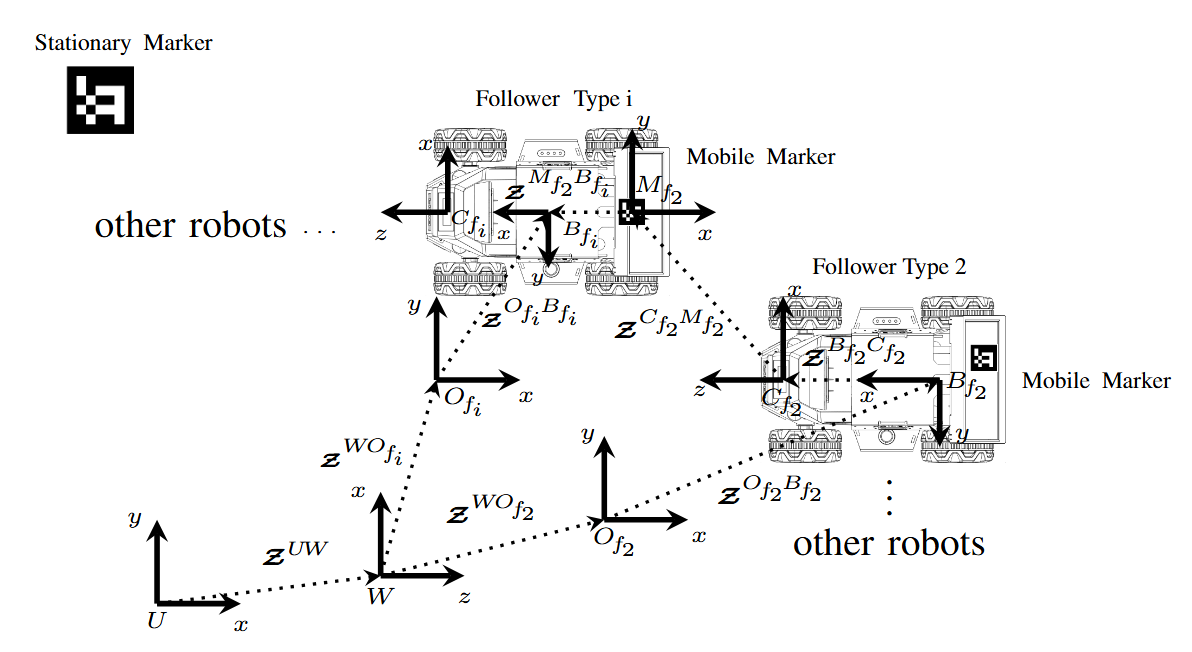}}
  }
  \\
  \subfloat[]{\scalebox{0.85}{\includegraphics[width=9cm]{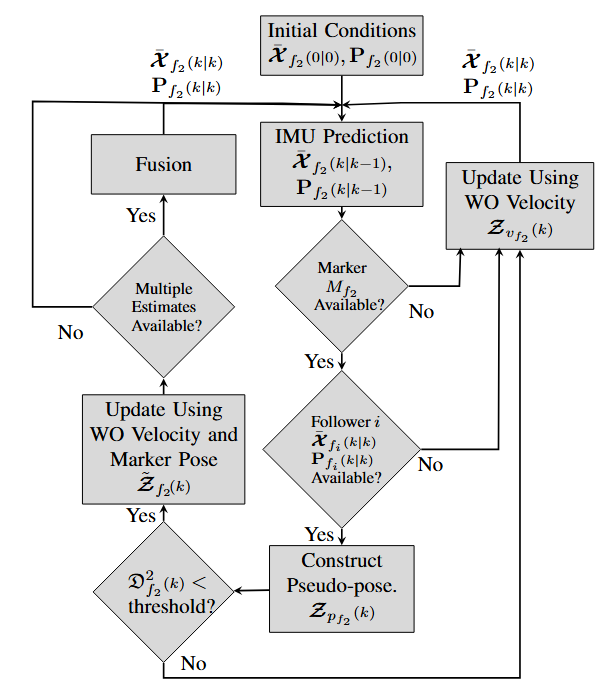}}}

  \caption{Follower type 2 robot, (a) coordinate frames, (b) estimation framework.}
  \label{fig: follower 2}
\end{figure}

Here, we develop the pseudo-pose of a follower robot type 2 with robot global frame $\{O_{f_2}\}$, camera frame $\{C_{f_2}\}$ and body frame $\{B_{f_2}\}$ based on its observations of a mobile landmark (fiducial marker) with coordinate frame $\{M_{f_2}\}$ installed on a neighboring follower robot, as well as the estimates from a type $i$ (can be $1$ or $2$) neighboring robot with robot and body frames $\{O_{f_i}\}$ and $\{B_{f_i}\}$. As illustrated in Fig.\ref{fig: follower 2}-(a) the following relation holds:
\begin{align*}\nonumber
 &\vect{\mathcal{Z}}_{{p_{f_2}}}\!\coloneqq \vect{\mathcal{Z}}_i^{O\!_{f\!_2}\!B\!_{f\!_2}}\!=\!\\&(\vect{\mathcal{Z}}^{W\!O\!_{f\!_2}})^{-1}\vect{\mathcal{Z}}^{W\!O\!_{f\!_i}}\vect{\mathcal{Z}}^{O\!_{f\!_i}\!B\!_{f\!_i}}(\vect{\mathcal{Z}}^{B\!_{f\!_2}\!C\!_{f\!_2}}\vect{\mathcal{Z}}^{C\!_{f\!_2}\!M\!_{f\!_2}}\vect{\mathcal{Z}}^{M\!_{f\!_2}\!B\!_{f\!_i}})^{-1}.  
\end{align*}
Applying the inverse and composition methods, the mean and covariance of the pseudo-pose $\vect{\mathcal{Z}}_{{p_{f_2}}}$ are
\begin{align*}\nonumber
 \Bvect{\mathcal{Z}}_{{p_{f_2}}}&\!=\!(\!\vect{\mathcal{Z}}^{W\!O\!_{f\!_2}}\!)^{-1}\!\vect{\mathcal{Z}}^{W\!O\!_{f\!_i}}\!\Bvect{\mathcal{X}}^{\mathbb{SE}(3)}_{f\!_i}(\vect{\mathcal{Z}}^{B\!_{f\!_2}\!C\!_{f\!_2}}\Bvect{\mathcal{Z}}^{C\!_{f\!_2}\!M\!_{f\!_2}}\vect{\mathcal{Z}}^{M\!_{f\!_2}\!B\!_{f\!_i}})^{-1}, \\   \vect{R}_{{p_{f_2}}}&\!\approx\!\vect{Ad}_{\vect{\mathcal{Z}}^{B\!_{f\!_2}\!C\!_{f\!_2}}}\vect{Ad}_{\Bvect{\mathcal{Z}}^{C\!_{f\!_2}\!M\!_{f\!_2}}}\vect{R}\!_{m\!_{f\!_2}}\vect{Ad}^\top_{\Bvect{\mathcal{Z}}^{C\!_{f\!_2}\!M\!_{f\!_2}}}\vect{Ad}^\top_{\vect{\mathcal{Z}}^{B\!_{f\!_2}\!C\!_{f\!_2}}}+\\&\vect{Ad}_{\vect{\mathcal{Z}}^{B\!_{f\!_2}\!C\!_{f\!_2}}\Bvect{\mathcal{Z}}^{C\!_{f\!_2}\!M\!_{f\!_2}}\vect{\mathcal{Z}}^{M\!_{f\!_2}\!B\!_{f\!_i}}}\vect{P}\!_{f\!_i}^{\mathbb{SE}(3)}\vect{Ad}^\top_{\vect{\mathcal{Z}}^{B\!_{f\!_2}\!C\!_{f\!_2}}\Bvect{\mathcal{Z}}^{C\!_{f\!_2}\!M\!_{f\!_2}}\vect{\mathcal{Z}}^{M\!_{f\!_2}\!B\!_{f\!_i}}},
    \end{align*}
where $\vect{\mathcal{Z}}^{O\!_{f\!_i}\!B\!_{f\!_i}}=\Bvect{\mathcal{X}}\!_{f\!_i}^{\mathbb{SE}(3)}\exp(\left[\vect{\zeta}_{f_i}^{\mathbb{SE}(3)}\right]_\wedge)$ with $\vect{\zeta}_{f_i}^{\mathbb{SE}(3)}\sim\mathcal{N}(\vect{0}_{1\times6},\vect{P}_{f_i}^{\mathbb{SE}(3)})$ and $\vect{\mathcal{Z}}^{C\!_{f\!_2}\!M\!_{f\!_2}}=\Bvect{\mathcal{Z}}^{C\!_{f\!_2}\!M\!_{f\!_2}}\exp(\left[\vect{\zeta}^{C\!_{f\!_2}}_{\!M\!_{f\!_2}}\right]_\wedge)$ with $\vect{\zeta}^{C\!_{f\!_2}}_{\!M\!_{f\!_2}}\sim\mathcal{N}(\vect{0}_{1\times6},\vect{R}\!_{m\!_{f\!_2}})$.

We let $\vect{\mathcal{Z}}_{{p_{f_2}}}=\Big[\begin{smallmatrix}
    \vect{\mathcal{R}}^{O\!_{f\!_2}\!B\!_{f\!_2}}_{p}&\vect{p}_{{B\!_{f\!_2}}_{p}}^{O\!_{f\!_2}}\\\vect{0}_{1\times3}&1 
\end{smallmatrix}\Big]$, then the measurement matrix for a follower type 2 from observing the follower type $i$ is as follows:
\begin{align*}
    \Tvect{\mathcal{Z}}\!_{{f\!_2}}{\scriptstyle(k)}\!=\!\begin{bmatrix}\!
   \begin{bmatrix}\!
       \vect{\mathcal{R}}^{O\!_{f\!_2}\!B\!_{f\!_2}}_{p}\!{\scriptstyle(k)}&\vect{p}\!_{{B\!_{f\!_2}}_{p}}^{O\!_{f\!_2}}\!{\scriptstyle(k)}\\\vect{0}_{1\times3}&1
   \!\end{bmatrix}&\vect{0}_4\\\vect{0}_4&\!\begin{bmatrix} \!\vect{I}_3&\vect{v}\!_{{B\!_{f_2}}_v}^{O_{f_2}}\!{\scriptstyle(k)}\\\vect{0}_{1\times3}&1\!\end{bmatrix}
\end{bmatrix}\!\in\!\mathbb{R}^{8\!\times\!8},
\end{align*}

\noindent with the covariance matrix of  $\vect{R}_{{f_2}}=\operatorname{diag}([\vect{R}_{{p_{f_2}}},\epsilon\vect{I}_3,\vect{R}_{v_{f_2}}])\in\mathbb{R}^{12\times12}$. The update is done if the constructed pseudo-poses are reliable (Section~\ref{sec: FDM}).

\subsection{Fault Detection (FD)}\label{sec: FDM}
To diagnose faults in measurements, a statistical index known as the Mahalanobis distance is used~\cite{mahalanobis1930test}. The Mahalanobis distance quantifies the goodness-of-fit between observed measurements $\Tvect{\mathcal{Z}}{\scriptstyle (k)}$ and theoretical expected measurements $\Bvect{\mathcal{Z}}{\scriptstyle (k)}=\vect{h}(\Bvect{\mathcal{X}}{\scriptstyle (k|k-1)})$. This index and its squared have been widely used in the literature for sensor fault/anomaly diagnosis~\cite{lin2010detecting}. Here, we use the squared Mahalanobis distance and define it for the system~\eqref{eq: process model}-\eqref{eq: process model2} as 
\begin{align}
    \mathfrak{D}^2{\scriptstyle (k)}=\vect{\nu}^\top{\scriptstyle (k)}\big(\vect{\mathcal{H}}{\scriptstyle (k)}\vect{P}{\scriptstyle (k|k-1)}\vect{\mathcal{H}}^\top\!{\scriptstyle (k)}\!+\!\vect{R}{\scriptstyle (k)}\big)^{-1}\vect{\nu}{\scriptstyle (k)},
\end{align}
where, the innovation vector on matrix Lie groups is 
\begin{align}
    \vect{\nu}{\scriptstyle (k)}&=\big[\log\big(\vect{h}^{-1}(\Bvect{\mathcal{X}}{\scriptstyle (k|k-1)})\vect{\mathcal{Z}}{\scriptstyle (k)}\big)\big]_\vee.
\end{align}
To filter the faulty measurements, the update step of the EKF is performed when a new measurement 
$\vect{\mathcal{Z}}{\scriptstyle (k)}$ is received and the calculated value of 
$\mathfrak{D}^2{\scriptstyle (k)}$ is found to be below a pre-defined threshold. The Mahalanobis index $\mathfrak{D}_l^2{\scriptstyle (k)}$ for a leader robot
 is calculated when stationary marker measurements are available.  For the follower type 1, when both mobile marker measurements and estimates of the leader robot are available the index $\mathfrak{D}_{f_1}^2{\scriptstyle (k)}$ is calculated. Finally, in the follower type 2 framework, when mobile marker measurements and the pose estimate of the $i^{th}$ follower robot are available the index 
$\mathfrak{D}_{{f_2}}^2{\scriptstyle (k)}$ is calculated. The variables used to calculate the index for these three types of robots are summarized in Table~\ref{table: mahalanobis}. Moreover, the matrix $\vect{\mathcal{H}}$ for all cases is calculated using \eqref{eq: matrix H} and the following measurement function:
\begin{align*}
    \vect{h}(\vect{\mathcal{X}}{\scriptstyle(k)})= \!\begin{bmatrix}\!
   \begin{bmatrix}
       \vect{\mathcal{R}}^{O\!B}\!{\scriptstyle(k)}&\vect{p}_{{B}}^{O}{\scriptstyle(k)}\\\vect{0}_{1\times3}&1
   \end{bmatrix}&\vect{0}_4\\\vect{0}_4&\begin{bmatrix} \vect{I}_{3}&\vect{\mathcal{R}}^{{O\!B}^\top}\!\!\!\!{\scriptstyle(k)}\vect{v}_{{B}}^{O}{\scriptstyle(k)}\\\vect{0}_{1\times3}&1\!\end{bmatrix}
\end{bmatrix}.
\end{align*}
The complete estimation frameworks for a leader robot along with that for follower type 1 and 2 are illustrated in Fig.\ref{fig: leader}-(b), Fig.\ref{fig: follower 1}-(b), and Fig.\ref{fig: follower 2}-(b), respectively. Note that when multiple estimates of the robot's states are available in Fig.\ref{fig: follower 2}-(b), they can be fused using the fusion described in Section~\ref{sec: configuration fusion}.

\begin{table}[h]
\caption{The values used for Mahalanobis index calculations.} 
\centering 
\scalebox{1}{
\begin{tabular}{c c c c c } 
\hline\hline
 Index& $\fontsize{6.5}{12}{\Bvect{\mathcal{X}}{\scriptstyle (k|k-1)}}$&$\Tvect{\mathcal{Z}}{\scriptstyle (k)}$ & $\vect{P}{\scriptstyle (k|k-1)}$ &$\vect{R}{\scriptstyle (k)}$\\
 \hline 
  $\mathfrak{D}_l^2{\scriptstyle (k)}$&$\fontsize{6.5}{12}\Bvect{\mathcal{X}}_l{\scriptstyle (k|k-1)}$& $\vect{\Tilde{\mathcal{Z}}}_{l}{\scriptstyle(k)}$&$\vect{P}_l{\scriptstyle (k|k-1)}$&$\vect{R}_{l}{\scriptstyle (k)}$\\[1mm]
 $\mathfrak{D}_{f_1}^2{\scriptstyle (k)}$&$\fontsize{6.5}{12}\Bvect{\mathcal{X}}_{f_1}{\scriptstyle (k|k-1)}$& $\vect{\Tilde{\mathcal{Z}}}_{f_1}{\scriptstyle(k)}$&$\vect{P}_{f_1}{\scriptstyle (k|k-1)}$&$\vect{R}_{f_1}{\scriptstyle (k)}$\\[1mm]
 $\mathfrak{D}_{{f_2}}^2{\scriptstyle (k)}$&$\fontsize{6.5}{12}\Bvect{\mathcal{X}}_{{f_2}}{\scriptstyle (k|k-1)}$& $\vect{\Tilde{\mathcal{Z}}}_{{f_2}}{\scriptstyle(k)}$&$\vect{P}_{{f_2}}{\scriptstyle (k|k-1)}$&$\vect{R}_{{f_2}}{\scriptstyle (k)}$\\ 
\hline 
\end{tabular}
}\label{table: mahalanobis} 
\end{table}

\section{Experiments and Results}\label{sec:exp and results}
In our experimental study, we established a controlled and representative testing environment, accounting for key factors including lighting conditions, vibrations, marker size, and terrain variations. The experimental setup is depicted in Fig.~\ref{fig: exp setup}, with detailed descriptions of the systems and sensors involved provided in the subsequent sections.



\subsection{Motion Tracker System}
To enable validation of robot motions, we executed a ground truth  using the PhaseSpace active optical motion tracker system~\cite{phasespace2025}. As shown in Fig.\ref{fig: exp setup}-(a), this system utilizes data from four ceiling-mounted cameras, oriented around a common workspace while each robot is equipped with eight fixed LEDs configured as a rigid body. The system reports the  poses of rigid bodies with an update rate of up to $960~Hz$.

\subsection{Robot Specification and Onboard Sensors}
\begin{figure}[htbp]
  \unitlength=0.5in
   \centering 
    \subfloat[]{
   \includegraphics[width=1.55in]{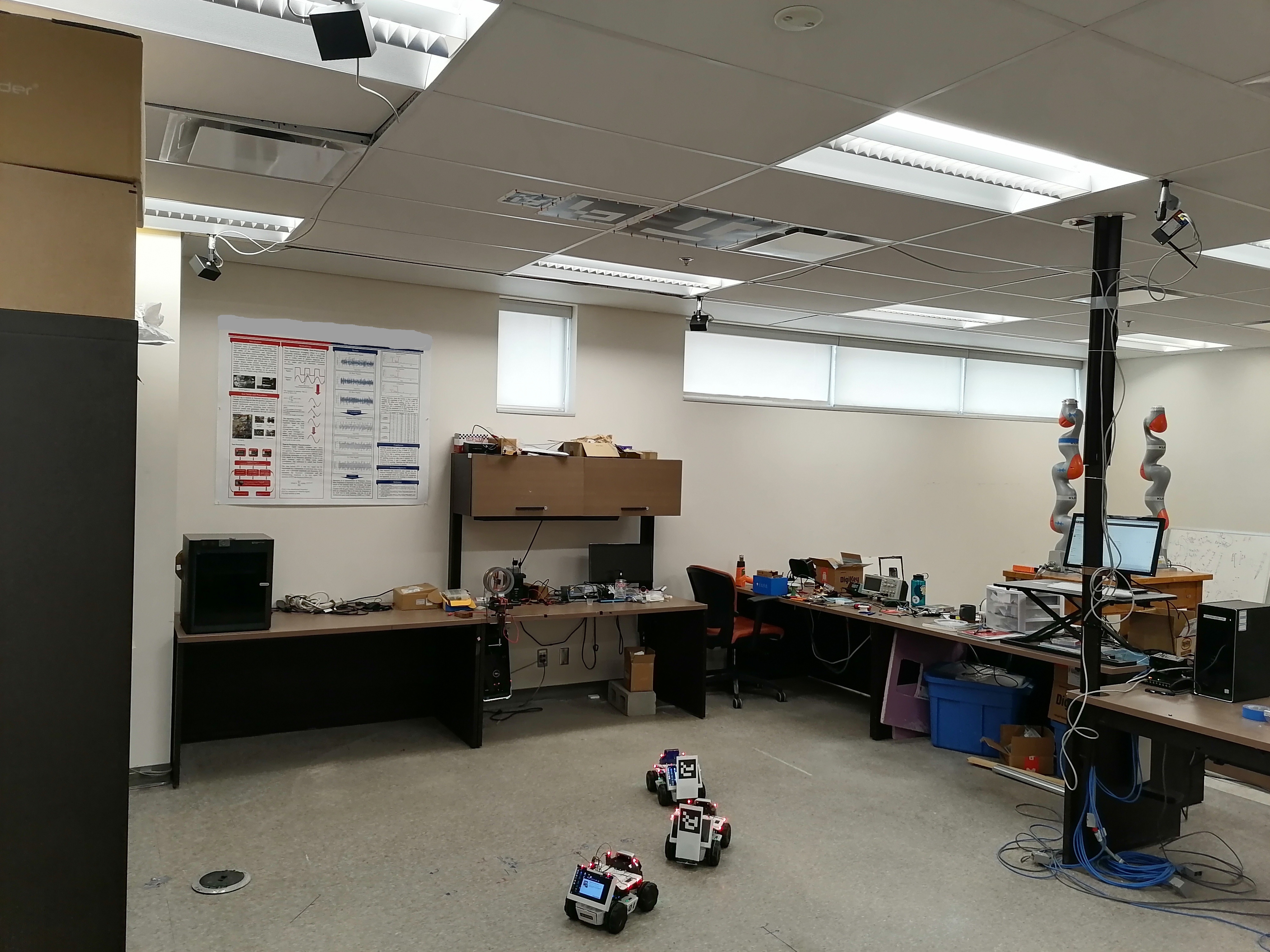}
  }~
      \subfloat[]{\scalebox{0.5}{ 
        \begin{tikzpicture}
            \node[anchor=south west,inner sep=0] (image) at (0,0) {\includegraphics[width=0.3\textwidth]{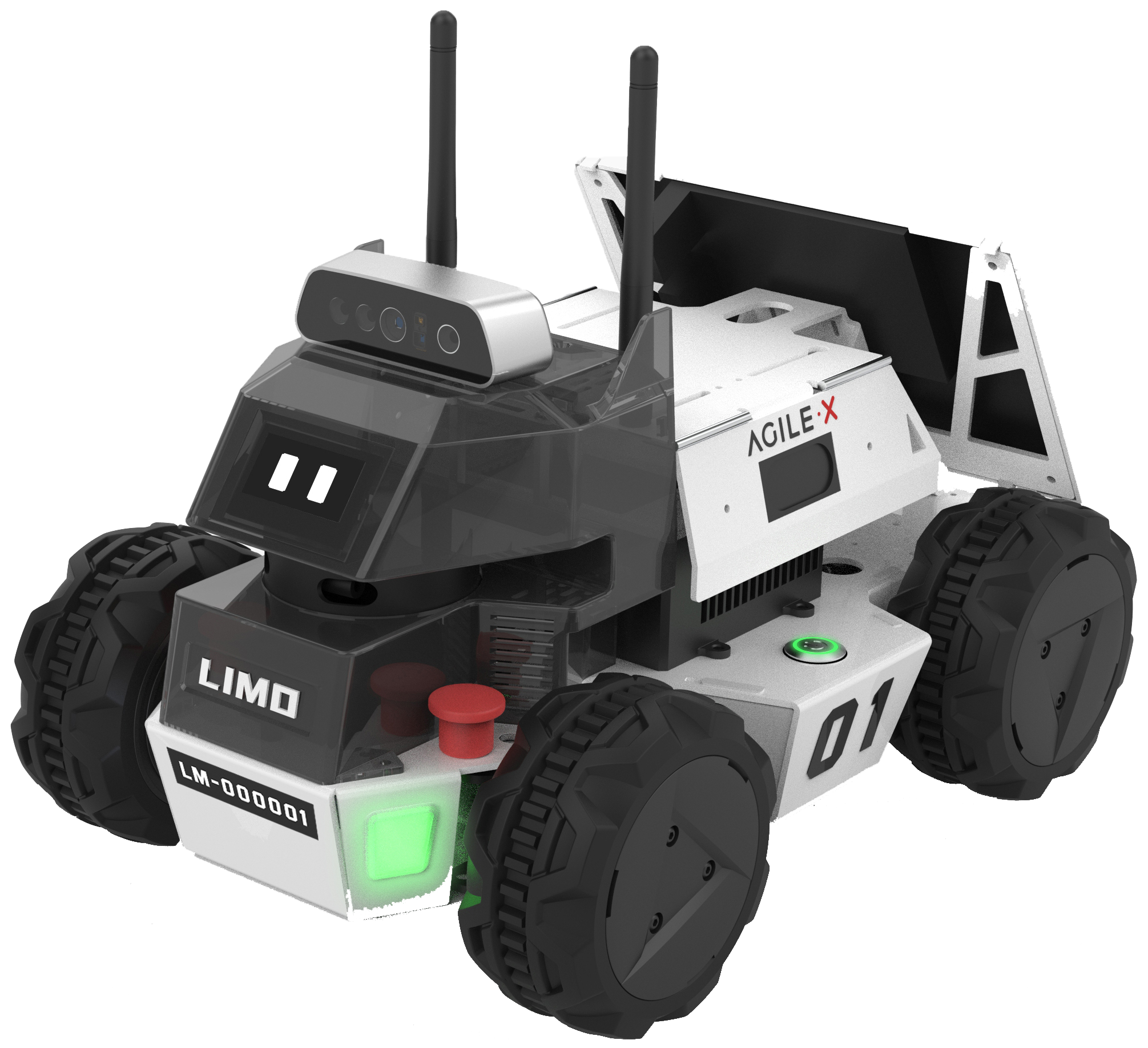}};
            
            \begin{scope}[x={(image.south east)},y={(image.north west)}]
                \node[black,font=\scriptsize] at (-0.1, 0.83) {ORBBEC® Dabai Stereo Depth Camera};
                \draw[->,  >=stealth, line width=0.1mm]  (0.3,0.7) -- (0,.8);
                \node[black,font=\scriptsize] at (0.1, 0.95) {Wi-Fi/Bluetooth Antennas};
                \draw[->,  >=stealth, line width=0.1mm]  (0.38,0.85) -- (0.2,.92);
                \node[black,font=\scriptsize] at (-0.2, 0.65) {EAI X2L LiDAR};
                \draw[->,  >=stealth, line width=0.1mm]  (0.3,0.45) -- (-0.2,.62);  
                \node[black,font=\scriptsize] at (0.6, 1) {Display};
                \draw[->,  >=stealth, line width=0.1mm]  (0.8,0.75) -- (0.65,0.95);
                \node[black,font=\scriptsize] at (0.8, 0.1) {IMU MPU6050};
                \draw[->,  >=stealth, line width=0.1mm]  (0.75,0.3) -- (0.75,0.15);
                \node[black,font=\scriptsize] at (0.0, 0.1) {NVIDIA Jetson Nano};
                \draw[->,  >=stealth, line width=0.1mm]  (0.65,0.43) -- (0.0,0.15);
            \end{scope}
        \end{tikzpicture}
    }
   
  }  \\
  \subfloat[]{
   \includegraphics[width=1.55in]{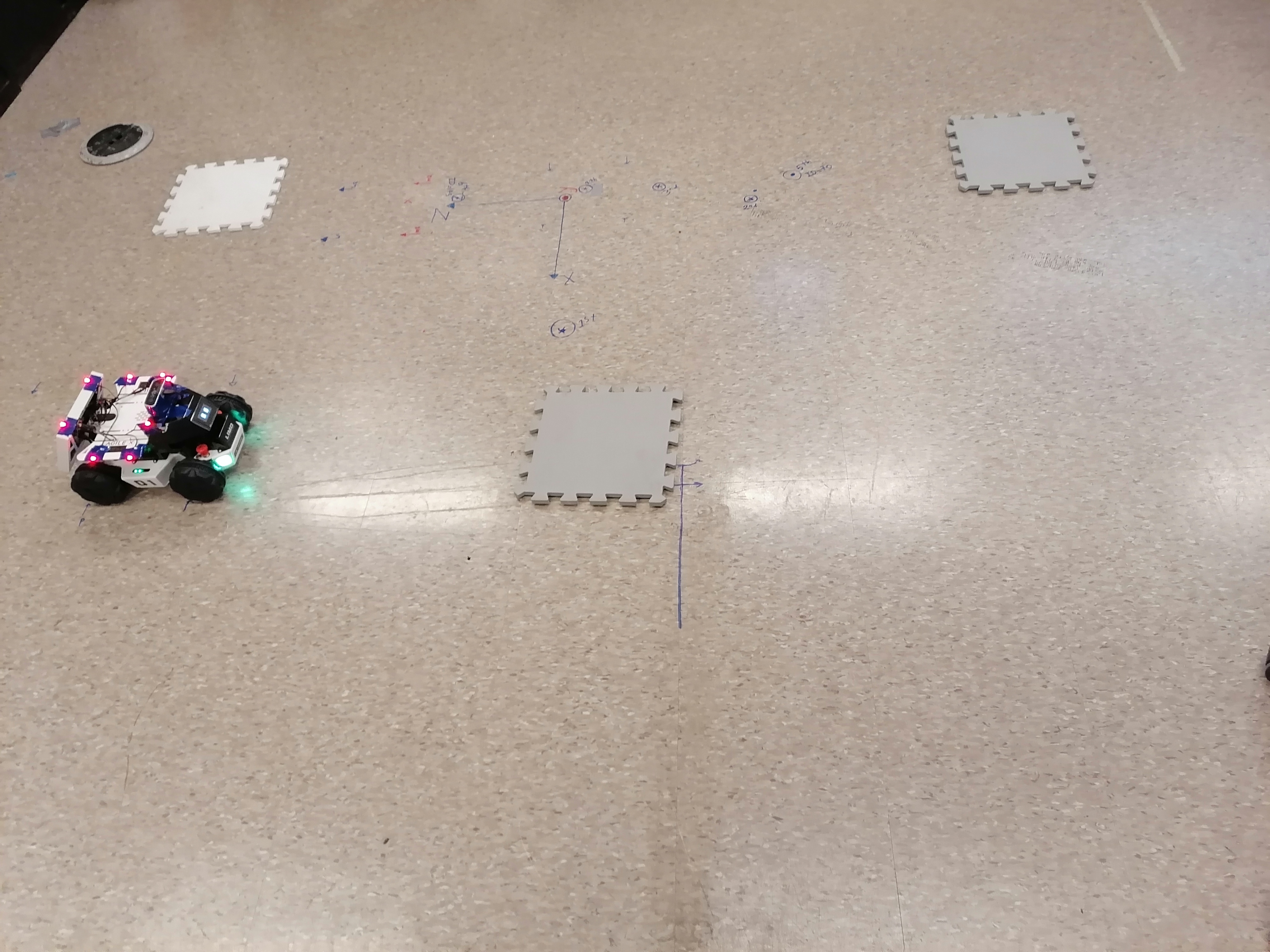}
  }~
  \subfloat[]{
   \includegraphics[width=1.55in]{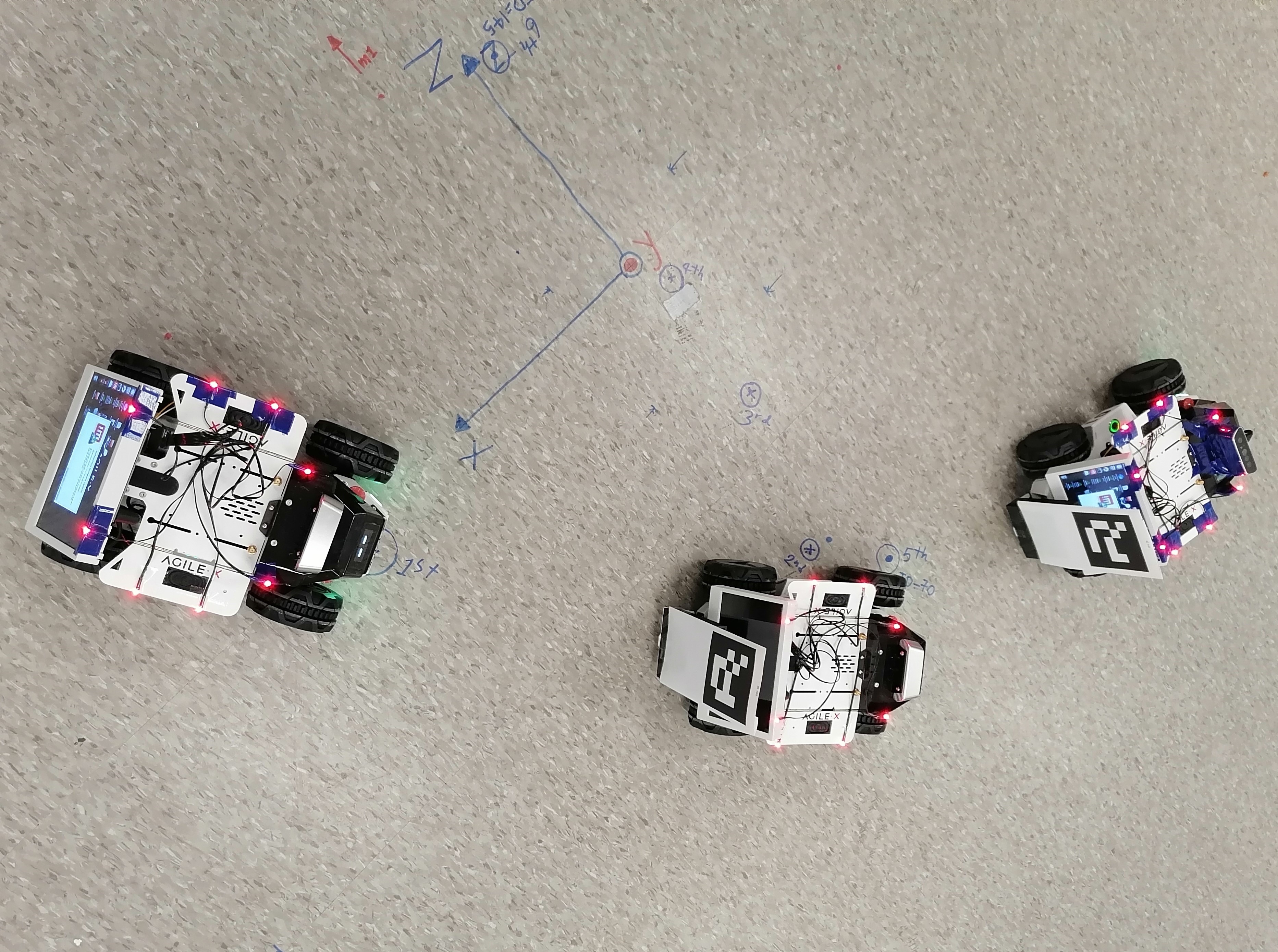}
  }\\
  \caption{The robot platform and  experimental setup. }\label{fig: exp setup}
\end{figure}

AgileX LIMO is a battery-powered, ROS-based robotic platform~\cite{limo2025}. The robot's components and sensors are depicted in Fig.\ref{fig: exp setup}-(b). It is equipped with an NVIDIA Jetson Nano, an EAI XL2 LiDAR, an ORBBEC® Dabai stereo depth camera, an MPU6050 IMU. The encoders are integrated  into the motors and the communication is enabled via WiFi and Bluetooth.  In the experiments, the robot was operated in the Ackermann  mode with a minimum turning radius of $0.4~m$. 

\subsubsection{Robot Camera Calibration}
 The calibration process provides essential data for obtaining the camera's intrinsic parameters, such as focal length and principal point, as well as extrinsic parameters, including rotation and translation vectors. In this study, calibration was performed using the ROS package in~\cite{cameraCalibration2025} and involved capturing 45 images of an $8\!\times\!6$ checkerboard with a square size of $10~cm$.



\subsubsection{IMU Calibration}
MPU-6050 is an integrated 6-axis IMU combining a 3-axis gyroscope, a 3-axis accelerometer, and a digital motion processor. Since the IMU measures normal forces, it is necessary to account for local gravity to correct the acceleration along the axis perpendicular to the ground. At the test site, with an elevation of $70~m$ and a latitude of $45.424721^{\circ}N$, the local gravity is $9.80637~\frac{m}{s^2}$. The IMU calibration involves placing the robot in a stable position on a flat surface and recording raw data for 100 seconds. The offsets for each axis are then computed as the average sensor readings. 

\subsubsection{Aruco Marker Pose Estimation}
 Using the ROS package~\cite{ArUcoROS2025}, the algorithm extracts ArUco marker corners, identifies their unique IDs, and calculates the marker's 3-D pose relative to the camera frame. In our experiments, we utilized the $5\!\times\! 5$ original ArUco dictionary. An ArUco with size $36~cm$ and ID=25 was used for the stationary marker, while mobile marker were $8.5~cm$ with ID=102 and ID=104.

\subsection{Setup Refinement}
To investigate the effects of various factors and refine the setup, we conducted a series of experiments. A robot, equipped with an upward-facing camera, follows a circular path around a stationary marker mounted on the ceiling and the pseudo-pose measurements are constructed based on the model described in Section~\ref{sec: pose update fid mark}. We examined the effects of lighting conditions by comparing results under full and reduced lighting, the impact of marker size by testing three marker dimensions ($17~cm$, $36~cm$, and $56~cm$), and the influence of the robot's forward speed using three speeds ($5~cm/s$, $10~cm/s$, and $50~cm/s$). 
\begin{figure}[htbp]
  \unitlength=0.5in
   \centering 
    \subfloat[]{
   \includegraphics[width=1.65in]{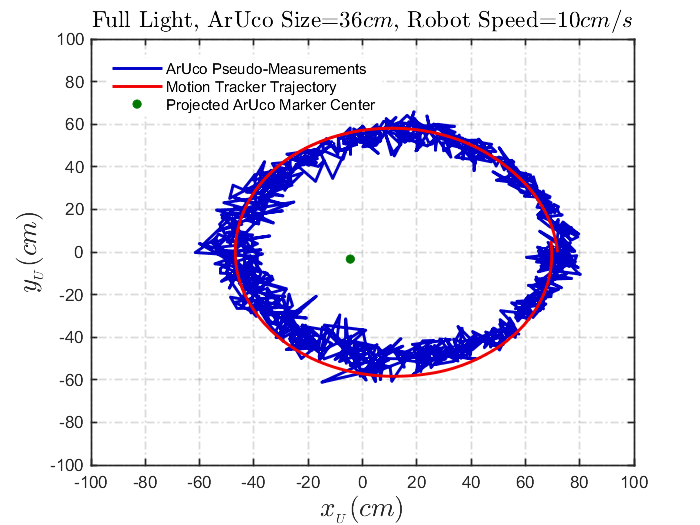}
  }~
      \subfloat[]{
   \includegraphics[width=1.65in]{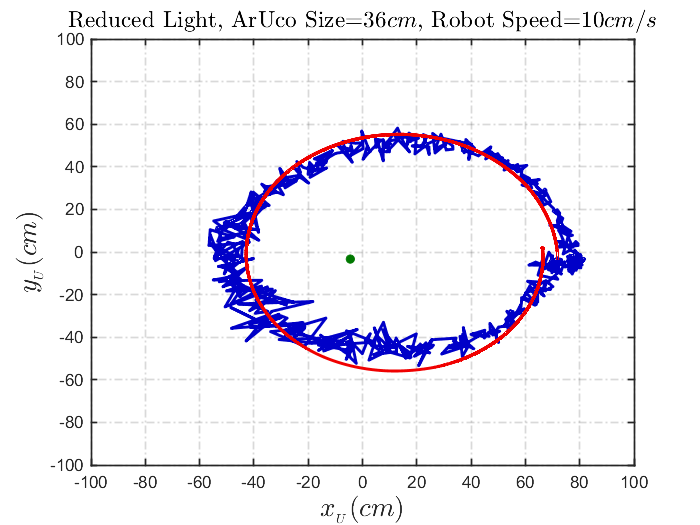}
  }\\\subfloat[]{
   \includegraphics[width=1.65in]{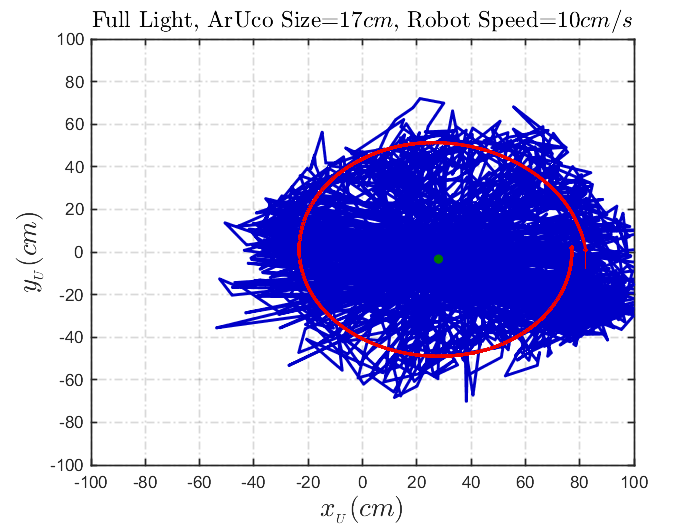}
  }~
      \subfloat[]{
   \includegraphics[width=1.65in]{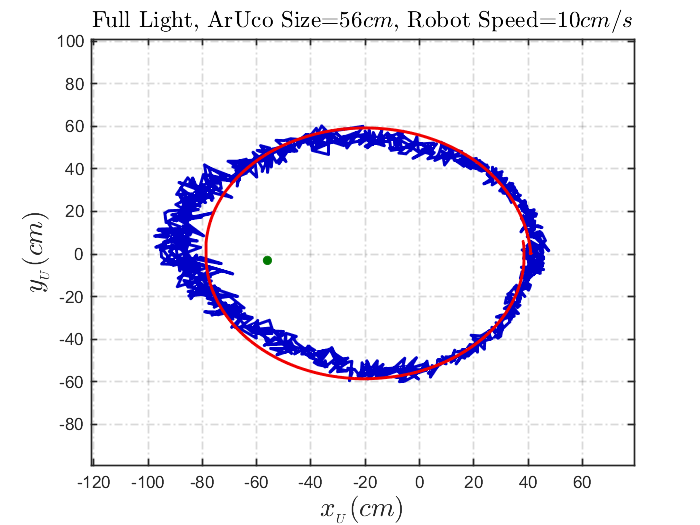}
  }
  \\\subfloat[]{
   \includegraphics[width=1.65in]{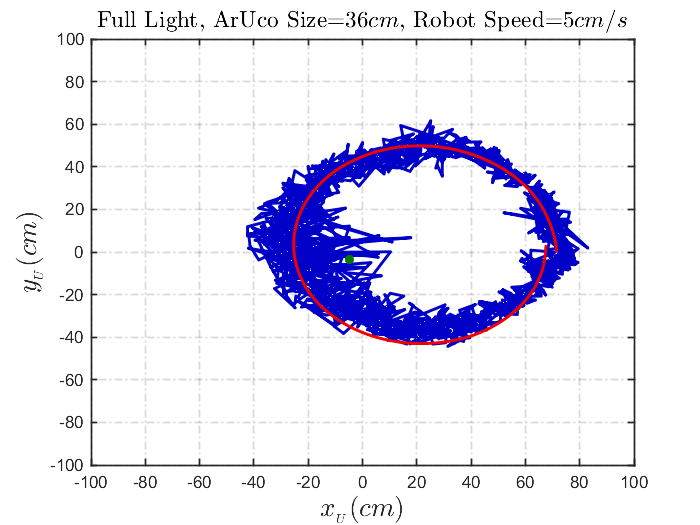}
  }~
      \subfloat[]{
   \includegraphics[width=1.65in]{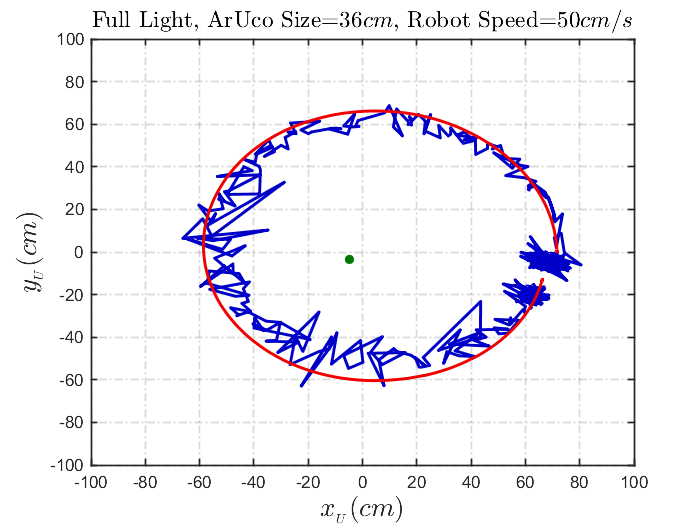}
  }\\
  \caption{Effects of lighting conditions, marker size, and robot speed on the accuracy of ArUco pseudo-pose measurements. }\label{fig: results_setup}
\end{figure}

The results illustrated in Fig.\ref{fig: results_setup} highlight three key points. First, by comparing Fig.\ref{fig: results_setup}-(a) and Fig.\ref{fig: results_setup}-(b), we observe that when the lights are fully on, the increased reflections and glare on the camera lens leads to noisier measurements. This experiment highlights the importance of maintaining a uniform lighting environment. Second, comparing the results  in Fig.\ref{fig: results_setup}-(a), (c), and (d) show that the size of the ArUco marker significantly affects the performance. When the marker size is $17~cm$, the system performs poorly, but as the marker size increases, less noisy measurements are obtained. Finally, a comparison of Fig.\ref{fig: results_setup}-(a), (e), and (f) shows that at higher speeds, the robot's smoother motion helps reduce the effects of small distortions and environmental variations, thereby improving pose estimation performance. However, it is important to note that excessively high speeds may eventually exceed the system’s capacity to detect the marker. In our setup, due to space limitations, as well as considerations regarding the motion tracker, safety, and placement of the ArUco marker, we used a $36~cm$ ArUco, set the robot speed to $10~cm/s$, and keep the lights on for the remaining experiments.

\subsection{Localization Results}

\subsubsection{Leader Robot Localization Results}
\begin{figure}[htbp]
  \unitlength=0.5in
   \centering 
    \subfloat[]{
   \includegraphics[width=1.7in]{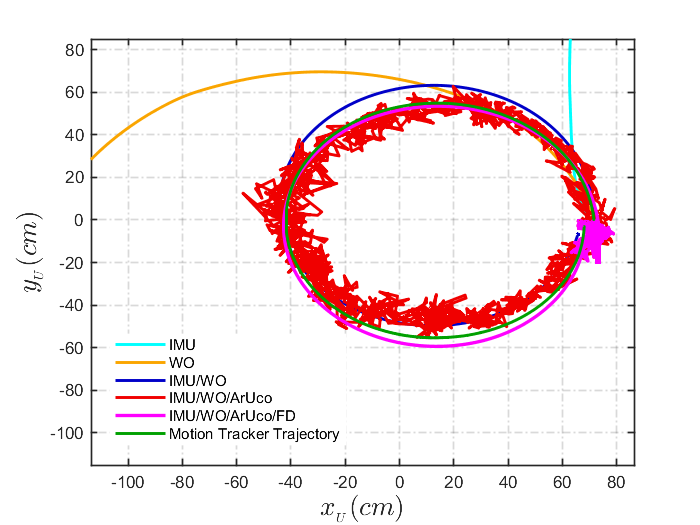}
  }~
      \subfloat[]{
   \includegraphics[width=1.7in]{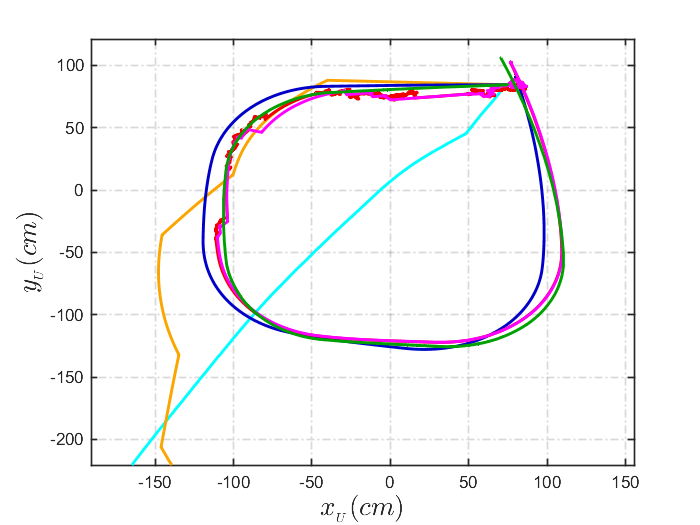}
  }\\\subfloat[]{
   \includegraphics[width=1.7in]{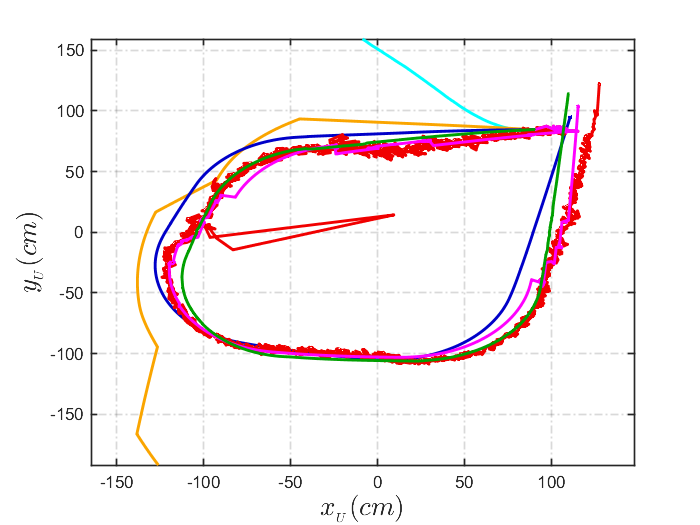}
  }~
      \subfloat[]{
   \includegraphics[width=1.7in]{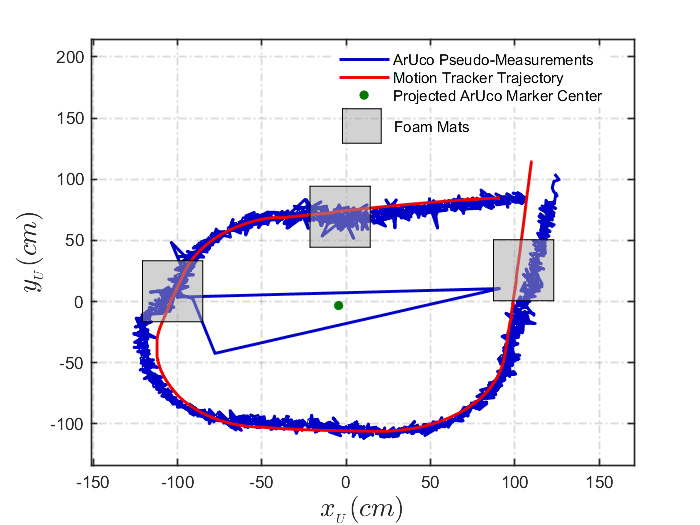}
  }
  \\
  \caption{Leader localization performance; (a) circular motion, (b) cornering motion, (c) uneven motion, and (d) ArUco measurements of uneven motion }\label{fig: results_single}
\end{figure}

Here, we investigate the localization of a leader robot subjected to various trajectories, including circular motion, cornering motions, and traverse over uneven terrain created by square foam mats placed along the path (See Fig.\ref{fig: exp setup}-(c)). In the experiments, sensor fusion is performed with  
$\vect{Q}_g=\vect{Q}_{b_g}=10^{-6}\vect{I}_3~{rad}^2/s^2$, $\vect{Q}_a=\vect{Q}_{b_a}=\vect{I}_3~m^2/s^4$,  
$\vect{R}_{v_l}=0.00015\vect{I}_3~m^2/s^2$, $\vect{R}_{m_l}=\operatorname{diag}([0.0075\vect{I}_3~{rad}^2,0.005\vect{I}_3~{m^2}])$, and $\vect{P}(0|0)=10^{-2}\vect{I}_{18}$. Moreover, the threshold in all cases is set to 40. The results are presented in Fig.\ref{fig: results_single}, from which we draw the following conclusions. The IMU trajectory fails quickly in all scenarios due to the robot being equipped with a low-cost IMU. A comparison between Fig.\ref{fig: results_single}-(a) and Fig.\ref{fig: results_single}-(b) reveals that WO fails when the robot attempts to turn, with the failure being particularly pronounced during the circular motion. Although the fusion of IMU and WO yields reasonable results, notable errors persist, underscoring the importance of incorporating more accurate measurements from the ArUco markers. Furthermore, when the robot turns around the ceiling-mounted lights during the circular motion (Fig.\ref{fig: results_single}-(a)) or traverses the square foam mats (Fig.~\ref{fig: results_single}-(d)),  ArUco pseudo-pose measurements become subject to significant noise. In these cases, the inclusion of the FD module effectively removes outliers. Thus, the fusion of IMU, WO, ArUco pseudo-poses, and FD more accurately captures the ground truth trajectory across all three types of motion.

\begin{figure}[htbp]
  \unitlength=0.5in
   \centering 
    \subfloat[]{
   \includegraphics[width=1.7in]{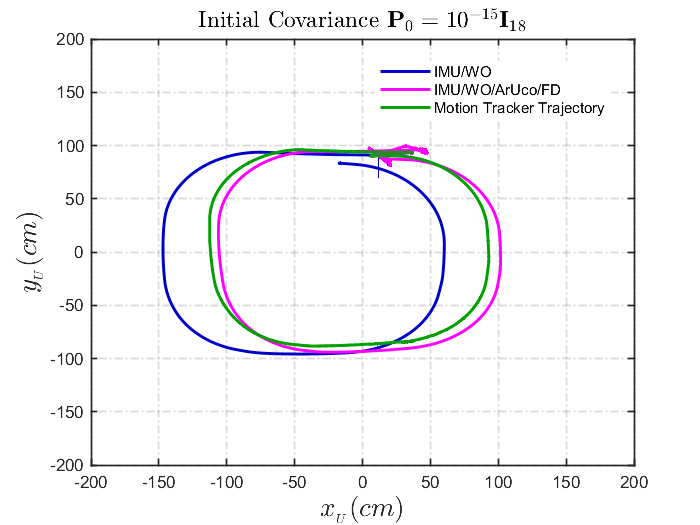}
  }~
      \subfloat[]{
   \includegraphics[width=1.7in]{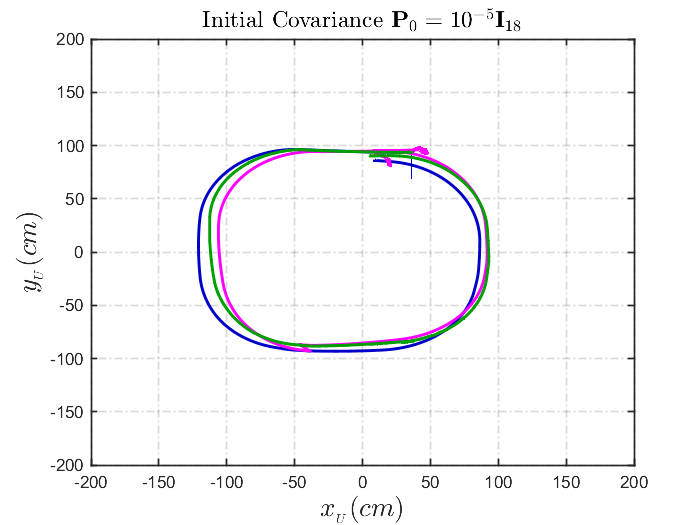}
  }\\
  \subfloat[]{
   
   \includegraphics[width=1.7in]{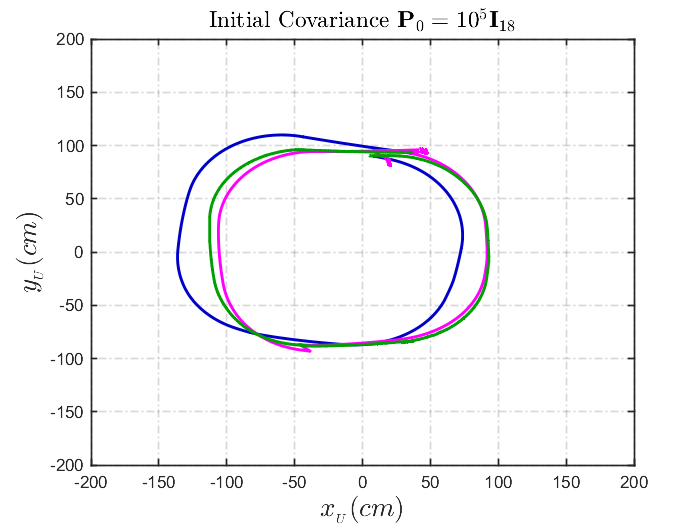}
  }~
      \subfloat[]{
   \includegraphics[width=1.7in]{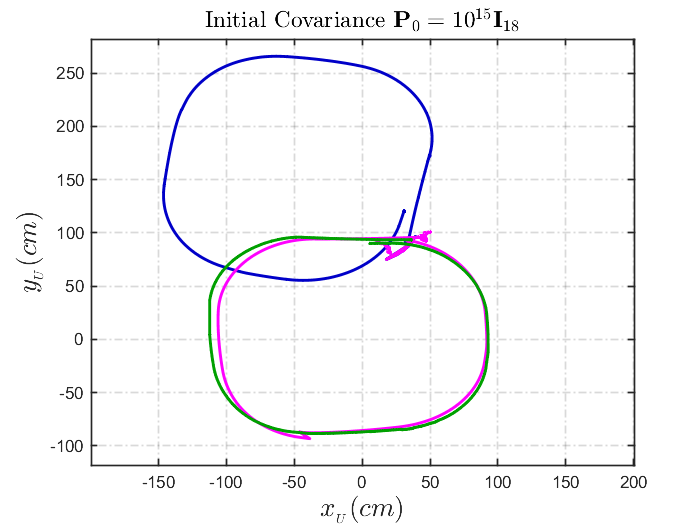}
  }   \\
  \caption{Effect of Initial Covariance $\vect{P}_0$ on Localization Performance}\label{fig: results_cov}
\end{figure}

Initial value of the covariance matrix is another factor impacting the EKF performance. Fig~\ref{fig: results_cov} shows the captured trajectory for the IMU/WO and IMU/WO/ArUco/FD filters with different initial covariance values. The results  verify that when the initial covariance is too low or too high, the IMU/WO trajectory exhibits significant deviations, while the IMU/WO/ArUco/FD filter shows little to no sensitivity to changes in $\vect{P}(0|0)=\vect{P}_0$.

\subsubsection{Follower Type 1 Localization Results}
\begin{figure}[htbp]
  \unitlength=0.5in
   \centering 
    \subfloat[]{
   \includegraphics[width=1.7in]{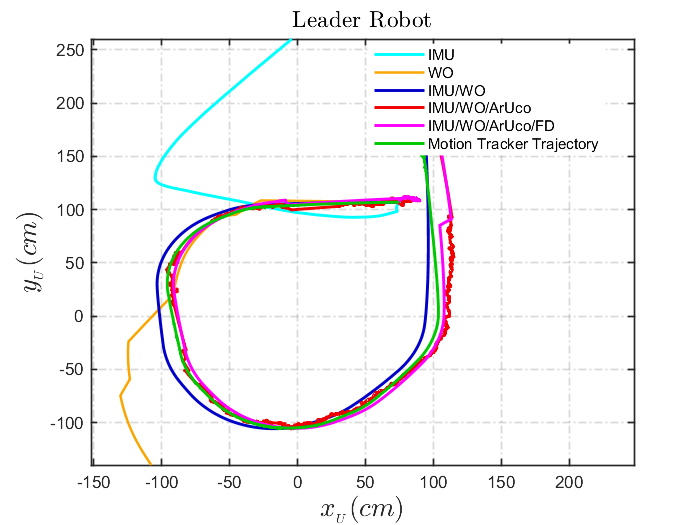}
  }~
      \subfloat[]{
   \includegraphics[width=1.7in]{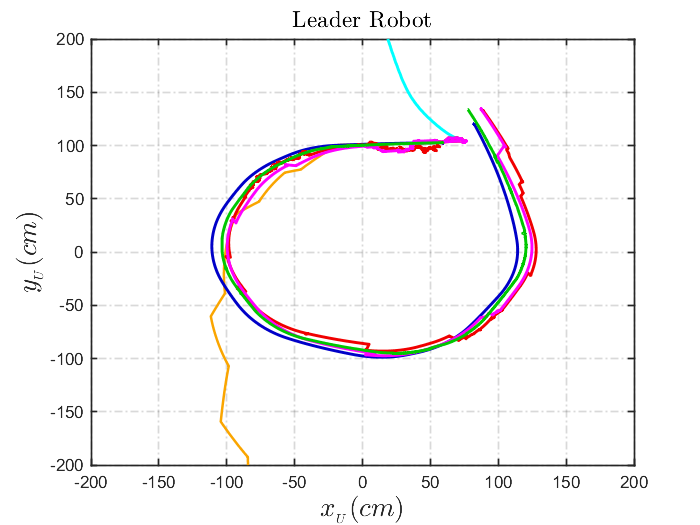}
  }  \\
  \subfloat[]{
   \includegraphics[width=1.7in]{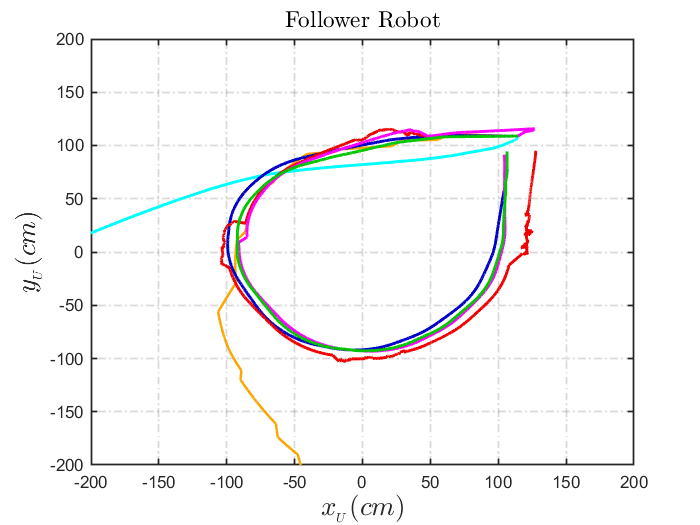}
  }~
      \subfloat[]{
   \includegraphics[width=1.7in]{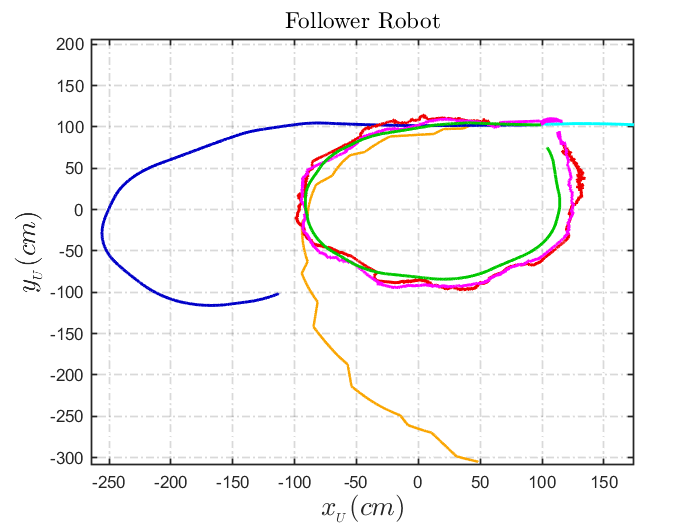}
  }  \\
  \caption{Performance of the Leader-Follower localization framework. }\label{fig: results_Two}
\end{figure}

In this experiment, the leader robot is equipped with an upward-facing camera that captures pose measurements from a stationary $36~cm$ ArUco marker mounted on the ceiling. A mobile $8.5~cm$ ArUco marker is placed on the back of the leader, and the follower constructs the pseudo-pose measurements using the model described in Section~\ref{sec: leader-follower update}.  The leader and follower robots share the same IMU and WO noise specifications and initial covariance conditions. The stationary marker noise specifications are consistent with those in the previous section while the follower's marker measurement noise covariance is  $\vect{R}\!_{m\!_{f_1}}=\operatorname{diag}([0.0050\vect{I}_3~{rad}^2,0.0016\vect{I}_3~{m^2}])$. The results of two tests are illustrated in Fig.\ref{fig: results_Two}. The trajectories of robots under cornering motion as shown in Fig\ref{fig: results_Two}-(a) and (c) demonstrate the overall improvement provided by the proposed method for both  robots. The effectiveness of the method becomes particularly evident during abrupt  decelerations and accelerations triggered by the control system (See Fig.\ref{fig: results_Two}-(b)-(d)). In our experiments, sharp maneuvers were necessary to navigate corners due to the limited coverage area of the motion tracker and the cameras' restricted field of view. This led to the failure of the IMU/WO filter for the follower robot. However, the trajectory is successfully recovered by integrating FD with ArUco marker pseudo-measurements.  

\subsubsection{Multi-Robot Localization Results}
\begin{figure}[htbp]
  \unitlength=0.5in
   \centering 
    \subfloat[]{
   \includegraphics[width=1.7in]{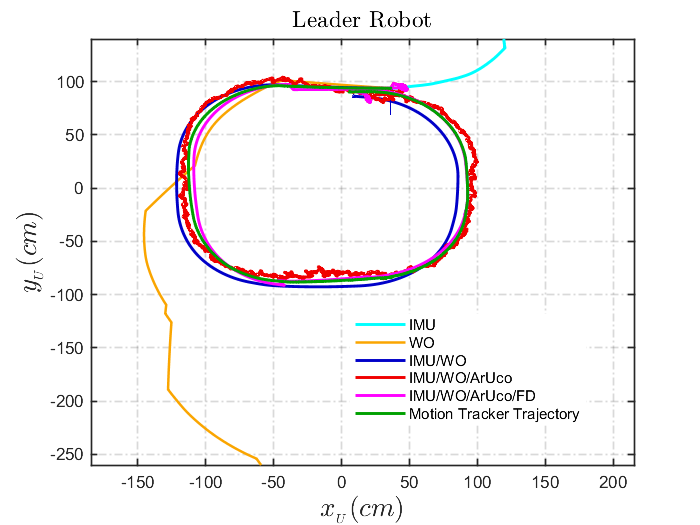}
  }~
      \subfloat[]{
   \includegraphics[width=1.7in]{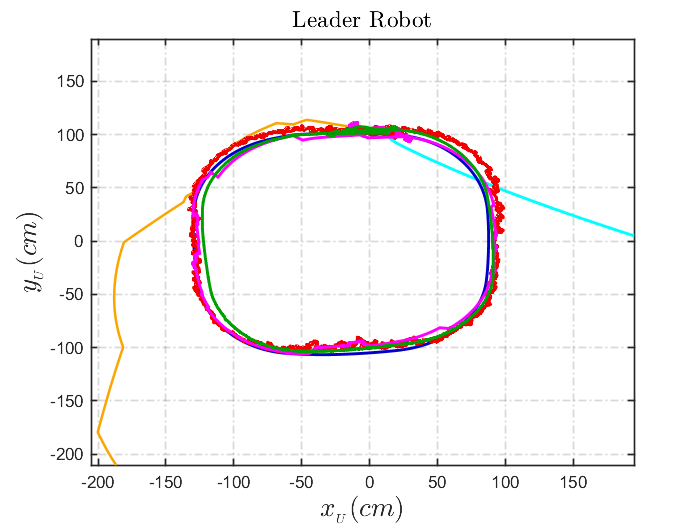}
  }  \\
  \subfloat[]{
   \includegraphics[width=1.7in]{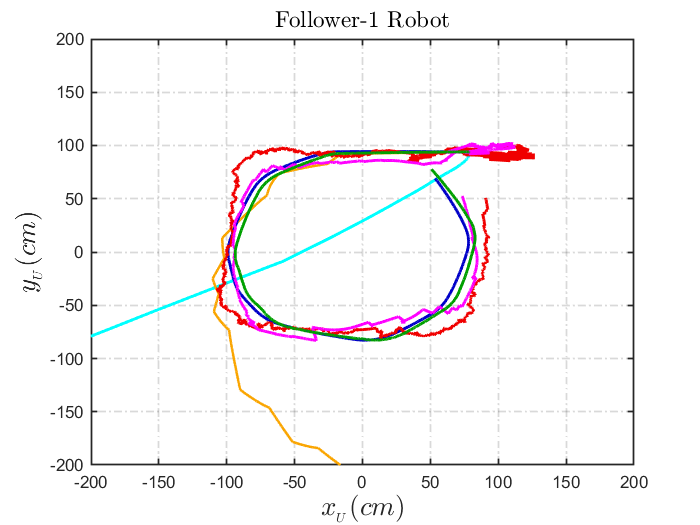}
  }~
      \subfloat[]{
   \includegraphics[width=1.7in]{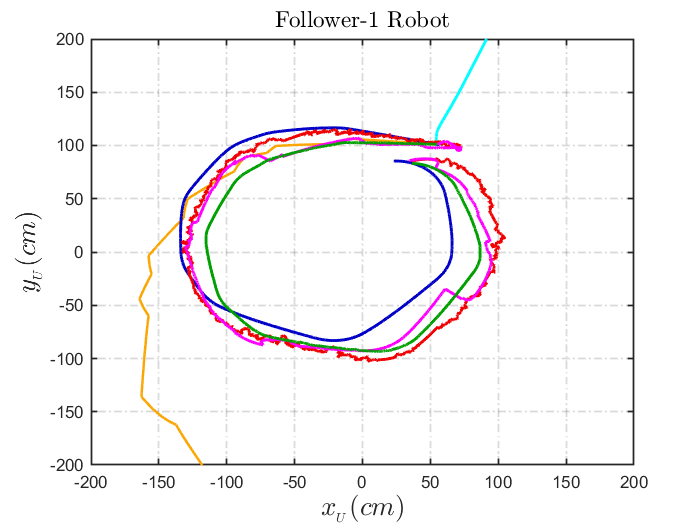}
  } \\
  \subfloat[]{
   \includegraphics[width=1.7in]{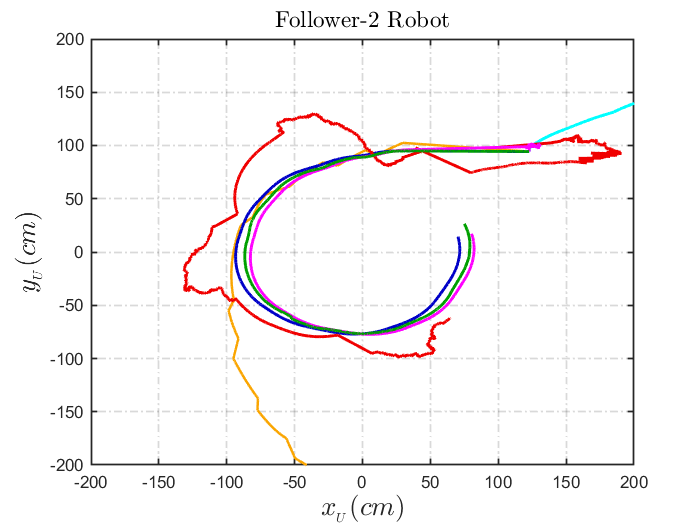}
  }~
      \subfloat[]{
   \includegraphics[width=1.7in]{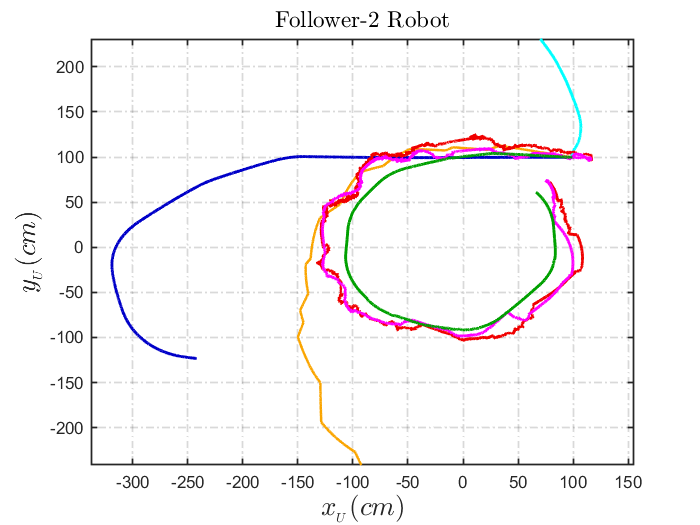}
  } \\
  \caption{Performance of the multi-robot localization framework.}\label{fig: results_Three}
\end{figure}

In these experiments, we use three robots to evaluate the scalability of the proposed method, as outlined in Section~\ref{sec: follower-follower}. The leader observes a stationary $36~cm$ ArUco marker on the ceiling, while a follower type 1 (follower-1) tracks an $8.5~cm$ mobile ArUco marker on the leader. A follower type 2 (follower-2) tracks an $8.5~cm$ mobile ArUco marker on follower-1. The initial covariance and noise properties of the follower-2 are the same as follower-1. Two sets of experiments are presented in Fig.\ref{fig: results_Three}. In both cases, we observe that uncertainty propagates and increases from the leader to follower-1 and follower-2 robots. In the first experiment (Fig.\ref{fig: results_Three}-(a), (c), and (e)), when the IMU/WO filter provides good results for follower-1 and follower-2, but the ArUco marker measurements are less reliable, the FD modules filters out these measurements, and the final localization results closely match the ground truth trajectory for all robots. In the second set of experiments (Fig.\ref{fig: results_Three}-(b), (d), and (f)), when the IMU/WO filter performs poorly for follower-1 and fails for follower-2 due to sharp maneuvers, we observe that the proposed method can recover the ground truth trajectory with good precision.

\subsubsection{Fusion in Localization}
\begin{figure}[htbp]
  \centering
  \subfloat[]{
    \includegraphics[width=1.25in]{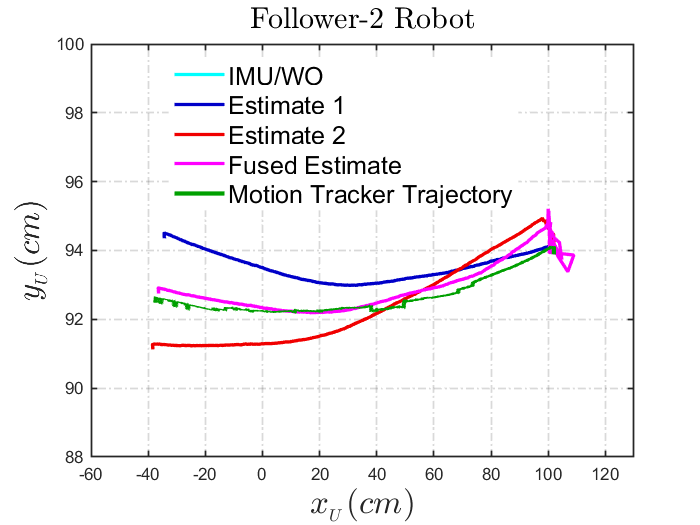}
  }~
  \hspace{-7.5mm} 
  \subfloat[]{
    \includegraphics[width=1.25in]{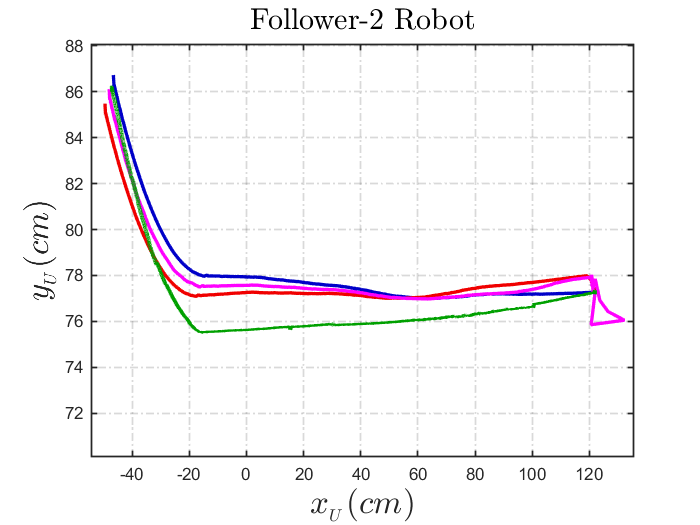}
  }~
  \hspace{-7.5mm}
  \subfloat[]{
    \includegraphics[width=1.25in]{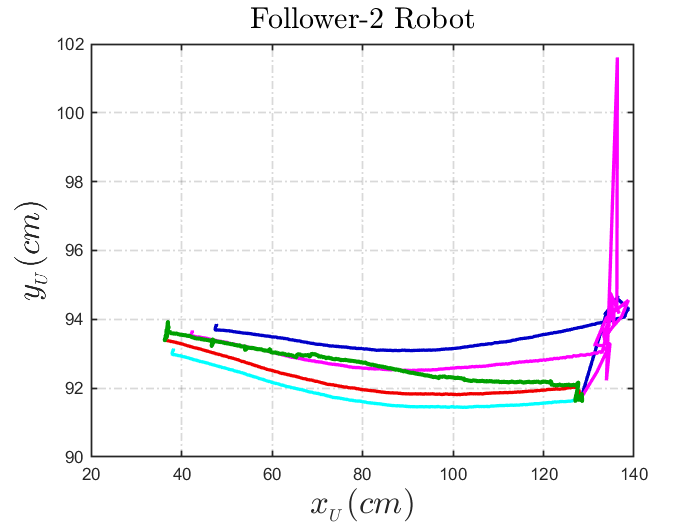}
  }

  \caption{Performance of fusion in the multi-robot localization framework.}
  \label{fig:results_fusion}
\end{figure}

To demonstrate the effectiveness of  fusion, we conducted tests in which follower-2 constructs two sets of pseudo-pose measurement  based on the relative poses from two robots. Using these measurements, follower-2 generates two individual  estimates of its own pose and combines them using the  fusion operation described in Section~\ref{sec: configuration fusion}. The results for three experiments are shown in Fig.~\ref{fig:results_fusion}. Due to hardware limitations, these experiments were conducted only during near-straight motion. In all cases, the reference member required for fusion was calculated using averaging operation with $\alpha_1=\alpha_2=0.5$. The results show that the overall performance of the fused estimate for follower-2 is better than the individual estimates and the IMU/WO filter.

  

\section{Conclusion}
In this paper, we introduced stochastic  operations on matrix Lie groups to perform operations such as composition, differencing, inverse, averaging, fusion, and constrained fusion on correlated estimates.  We then applied them to develop a scalable multi-robot localization framework that integrated multi-modal sensor fusion (inertial, velocity, and pose measurements) and fault detection techniques. Experiments with a network of wheeled mobile robots equipped with IMUs, wheel encoders, and ArUco markers, demonstrated real-time, efficient, and reliable localization across the entire network. The experiments also revealed that the performance of the localization system was influenced by factors such as lighting conditions, robot speed and vibration, uneven terrain, marker size, abrupt maneuvers, and initial conditions. While the proposed system demonstrated significant advances, there are several directions for future work. Future work could focus on integrating diverse sensors such as LiDAR to improve accuracy. Enhancing the fault detection module with machine learning for better detection of faulty sensor readings and expanding the system to handle more sensor faults would further strengthen its long-term robustness. Additionally, scaling the method for larger robot networks in complex environments requires optimizing communication strategies to manage bandwidth limitations.  Finally, adapting the framework for other robotic platforms, such as aerial or underwater robots, could address different localization challenges.

\bibliographystyle{IEEEtran}
\bibliography{refs}

\appendices
\section{Properties of the Developed Lie Group $\mathcal{G_X}$}\label{app: lie grop GX}
We begin by showing  
$\mathcal{G}_{\mathcal{X}}$
is a matrix Lie group, i.e., verifying  its closure, associativity, and the existence of identity and inverse elements. To this end, we first observe that the Lie group $\mathcal{G}_{\mathcal{X}}\coloneq\mathbb{SE}_2(3)\times\mathbb{R}^3\times\mathbb{R}^3$  is a special case of  Lie group:
\begin{align}
\mathcal{G}\!\coloneqq\!\begin{Bmatrix}\vect{\mathcal{X}}=\begin{bmatrix}
    \vect{\mathfrak{X}}_1&\vect{0}_5\\
    \vect{0}_5&\vect{\mathfrak{X}}_2
\end{bmatrix}\bigg|\vect{\mathfrak{X}}_1,\vect{\mathfrak{X}}_2\in\! \mathbb{SE}_2(3)\end{Bmatrix},
\end{align}
when the rotation angle in $\vect{\mathfrak{X}}_2$ is zero. We therefore establish the proofs for the $\mathcal{G}$.
For closure, we show the set of matrices in the Lie group $\mathcal{G}$ is closed under matrix multiplication, i.e., if $\vect{\mathcal{X}}=\left[\begin{smallmatrix}
    \vect{\mathfrak{X}}_1&\vect{0}_5\\
    \vect{0}_5&\vect{\mathfrak{X}}_2
\end{smallmatrix}\right]\in\mathcal{G},~\vect{\mathcal{Y}}=\left[\begin{smallmatrix}
    \vect{\mathfrak{Y}}_1&\vect{0}_5\\
    \vect{0}_5&\vect{\mathfrak{Y}}_2
\end{smallmatrix}\right]\in\mathcal{G}$, then $\vect{\mathcal{Z}}=\vect{\mathcal{X}}\vect{\mathcal{Y}}\in\mathcal{G}$. Specifically, we have
$\vect{\mathcal{Z}}=\vect{\mathcal{X}}\vect{\mathcal{Y}}=\left[\begin{smallmatrix}
    \vect{\mathfrak{X}}_1\vect{\mathfrak{Y}}_1&\vect{0}_5\\
    \vect{0}_5&\vect{\mathfrak{X}}_2\vect{\mathfrak{Y}}_2
\end{smallmatrix}\right].  
$
Since $\vect{\mathfrak{X}}_1\vect{\mathfrak{Y}}_1\in\mathbb{SE}_2(3)$ and $\vect{\mathfrak{X}}_2\vect{\mathfrak{Y}}_2\in\mathbb{SE}_2(3)$, therefore $\vect{\mathcal{Z}}\in\mathcal{G}$. The inverse elements of $\mathcal{G}$ is calculated as $\vect{\mathcal{X}}^{-1}=\left[\begin{smallmatrix}
    \vect{\mathfrak{X}}_1^{-1}&\vect{0}_5\\
    \vect{0}_5&\vect{\mathfrak{X}}_2^{-1}
\end{smallmatrix}\right].$
Since $\vect{\mathfrak{X}}_1^{-1}, \vect{\mathfrak{X}}_2^{-1}\in\mathbb{SE}_2(3)$, we conclude $\vect{\mathcal{X}}^{-1}\in\mathcal{G}$. The identity element of $\mathcal{G}$ is $\vect{I}_{10}$ and 
the associativity of the matrix Lie group 
$\mathcal{G}$ follows trivially from the associativity of matrix multiplication.  For a block real squared matrix $\vect{M}=\left[\begin{smallmatrix}
    \vect{M}_1&\vect{0}_n\\\vect{0}_n&\vect{M}_2
\end{smallmatrix}\right]$ we have $\exp(\vect{M})=\left[\begin{smallmatrix}
   \exp(\vect{M}_1)&\vect{0}_n\\\vect{0}_n&\exp(\vect{M}_2)
\end{smallmatrix}\right]$, which verifies that $\vect{\mathcal{X}}=\left[\begin{smallmatrix}
    \vect{\mathfrak{X}}_1&\vect{0}_5\\
    \vect{0}_5&\vect{\mathfrak{X}}_2
\end{smallmatrix}\right]=\left[\begin{smallmatrix}\exp([\vect{\xi}_1]_\wedge)&\vect{0}_5\\\vect{0}_5&\exp([\vect{\xi}_2]_\wedge)\end{smallmatrix}\right]=\exp([\vect{\zeta}]_\wedge)$ with  $[\vect{\zeta}]_\wedge$ being an element of Lie algebra of $\mathcal{G}$ given by 
  \begin{align}
\mathfrak{g}\!\coloneqq\!\begin{Bmatrix}[\vect{\zeta}]_\wedge\!=\!\begin{bmatrix}[\vect{\xi}_1]_\wedge&\vect{0}_5\\\vect{0}_5&[\vect{\xi}_2]_\wedge\end{bmatrix}\bigg|[\vect{\xi}_1]_\wedge,[\vect{\xi}_2]_\wedge\!\in\! \mathfrak{so}_2(3)\!\end{Bmatrix},
\end{align}  
where, $\vect{\zeta}=[\vect{\xi}_1^\top~~\vect{\xi}_2^\top]^\top\in\mathbb{R}^{18}$. The Lie algebra of $\mathcal{G}_{\mathcal{X}}$, denoted as $\mathfrak{g}_{\mathcal{X}}$, is a special case of $\mathfrak{g}$ when rotation angle in $\vect{\xi}_2$ is zero. From the definitions $\vect{Ad}_{\vect{\mathcal{X}}}(\vect{\zeta})\coloneqq [\vect{\mathcal{X}}\left[\vect{\zeta}\right]_\wedge\vect{\mathcal{X}}^{-1}]_\vee$ and $\vect{ad}_{\vect{\zeta}}(\vect{\eta})\!\coloneqq\! \big[\!\left[\vect{\zeta}\right]_\wedge\!,\!\left[\vect{\eta}\right]_\wedge\!\big]_\vee$ we obtain 
\begin{align}
 \vect{Ad}_{\vect{\mathcal{X}}}&=\begin{bmatrix}
  \vect{Ad}_{\vect{\mathfrak{X}}_1}&\vect{0}_9\\\vect{0}_9& \vect{Ad}_{\vect{\mathfrak{X}}_2}
 \end{bmatrix}\in\mathbb{R}^{18\times18},\\   
 \vect{ad}_{\vect{\zeta}}&=\begin{bmatrix}
  \vect{ad}_{\vect{\xi}_1}&\vect{0}_9\\\vect{0}_9& \vect{ad}_{\vect{\xi}_2}
 \end{bmatrix}\in\mathbb{R}^{18\times18}.  
\end{align}
Moreover, the inverse of the right Jacobian of the group $\mathcal{G}$ is
\begin{align}\nonumber
\mathcal{J}_r^{-1}(\vect{\zeta})&=\sum_{n=0}^\infty\frac{B_n}{n!}(-\vect{ad}_{\vect{\zeta}})^n\\\nonumber&=\begin{bmatrix}
  \sum_{n=0}^\infty\frac{B_n}{n!}(-\vect{ad}_{\vect{\xi}_1})^n&\vect{0}_{18} \\\vect{0}_{18}&\sum_{n=0}^\infty\frac{B_n}{n!}(-\vect{ad}_{\vect{\xi}_2})^n 
\end{bmatrix}\\&=\begin{bmatrix}
    \mathcal{J}_r^{-1}(\vect{\xi}_1)&\vect{0}_{18}\\\vect{0}_{18}&\mathcal{J}_r^{-1}(\vect{\xi}_2)
\end{bmatrix},
\end{align}
where, $\mathcal{J}_r^{-1}(\vect{\xi}_i)$ is the inverse of the right Jacobian of the $\mathbb{SE}_2(3)$ for $i\in\{1,2\}$. The proofs for the Lie group $\mathcal{G_Z}\coloneqq\mathbb{SE}(3)\times\mathbb{R}^3$ follows a similar procedure and is left to the reader. 

\section{Proof of Inverse Operation}\label{app: inverse}
We let the inverse of the stochastic member $\vect{\mathcal{X}}=\Bvect{\mathcal{X}}\exp([\vect{\zeta}]_\wedge)$ with $\vect{\zeta}\sim\mathcal{N}(\vect{0},\vect{P})$ to be  $\vect{\mathcal{X}_{\text{inv}}}=\vect{\mathcal{X}}^{-1}$. Thus,
\begin{align}\label{eq: X inv}
 \vect{\mathcal{X}}^{-1}_{\text{inv}}=\Bvect{\mathcal{X}}\exp([\vect{\zeta}]_\wedge)=\exp([\vect{Ad}_{\Bvect{\mathcal{X}}}\vect{\zeta}]_\wedge)\Bvect{\mathcal{X}}.  
\end{align}
Taking the inverse $\vect{\mathcal{X}_{\text{inv}}}=\Bvect{\mathcal{X}}^{-1}\exp(-[\vect{Ad}_{\Bvect{\mathcal{X}}}\vect{\zeta}]_\wedge).$
Thus, mean and covariance of $\vect{\mathcal{X}}_{\text{inv}}=\Bvect{\mathcal{X}}_{\text{inv}}\exp([\vect{\zeta}_{\text{inv}}]_\wedge)$ are 
\begin{align}
 &\Bvect{\mathcal{X}}_{\text{inv}}=\Bvect{\mathcal{X}}^{-1},\\  &\vect{\zeta}_{\text{inv}}=-\vect{Ad}_{\Bvect{\mathcal{X}}}\vect{\zeta},\\ 
  &\vect{P}_{\text{inv}}\!=\!\mathbb{E}[\vect{\zeta}_{\text{inv}}\vect{\zeta}_{\text{inv}}^\top]\!=\!\vect{Ad}_{\Bvect{\mathcal{X}}}\mathbb{E}[\vect{\zeta}\vect{\zeta}^\top]\vect{Ad}_{\Bvect{\mathcal{X}}}^\top \!=\!\vect{Ad}_{\Bvect{\mathcal{X}}}\vect{P}\vect{Ad}_{\Bvect{\mathcal{X}}}^\top.
\end{align}

\section{Proof of Composition Operation}\label{app: comp}
The proof is realized by induction. For the base step, i.e., $n=2$, the two stochastic members are defined as $\vect{\mathcal{X}}_1=\Bvect{\mathcal{X}}_1\exp([\vect{\zeta}_1]_\wedge)$ and $\vect{\mathcal{X}}_2=\Bvect{\mathcal{X}}_2\exp([\vect{\zeta}_2]_\wedge)$ with $\vect{\zeta}_1\sim\mathcal{N}(\vect{0},\vect{P}_{11})$ and $\vect{\zeta}_2\sim\mathcal{N}(\vect{0},\vect{P}_{22})$, respectively. Then we have
\begin{align}
\label{eq::randomPose100}
\vect{\mathcal{X}}_\text{com}^{(1)}=\vect{\mathcal{X}}_1\vect{\mathcal{X}}_2=\Bvect{\mathcal{X}}_1\exp([\vect{\zeta}_1]_\wedge)\Bvect{\mathcal{X}}_2\exp([\vect{\zeta}_2]_\wedge).
\end{align}
Next, we compute the inverse of $\vect{\mathcal{X}}_\text{com}^{(1)}$ as 
\begin{align}
\nonumber
(\vect{\mathcal{X}}_\text{com}^{(1)})^{-1}=\exp([\vect{-\zeta}_2]_\wedge)\Bvect{\mathcal{X}}_2^{-1}\exp([\vect{-\zeta}_1]_\wedge)\Bvect{\mathcal{X}}_1^{-1}.
\end{align}
Based on the definition of the Adjoint mapping we obtain
\begin{align}\nonumber
(\vect{\mathcal{X}}_\text{com}^{(1)})^{-1}=\exp([\vect{-\zeta}_2]_\wedge)\exp([-\vect{Ad}_{\Bvect{\mathcal{X}}_2^{-1}}\vect{\zeta}_1]_\wedge)\Bvect{\mathcal{X}}_2^{-1}\Bvect{\mathcal{X}}_1^{-1}, 
\end{align}
which results in
\begin{align}\label{eq:comp222}
\vect{\mathcal{X}}_\text{com}^{(1)}=\Bvect{\mathcal{X}}_1\Bvect{\mathcal{X}}_2\exp([\vect{Ad}_{\Bvect{\mathcal{X}}_2^{-1}}\vect{\zeta}_1]_\wedge)\exp([\vect{\zeta}_2]_\wedge). 
\end{align}
Thus, the mean  and uncertainty vector of $\vect{\mathcal{X}}_\text{com}^{(1)}=\Bvect{\mathcal{X}}_\text{com}^{(1)}\exp([\vect{\zeta}_{\text{com}}^{(1)}]_\wedge)$  considering the BCH first-order terms  are 
\begin{align}
\Bvect{\mathcal{X}}_\text{com}^{(1)}&=\Bvect{\mathcal{X}}_1\Bvect{\mathcal{X}}_2,\\ 
\vect{\zeta}_{\text{com}}^{(1)}&\approx\vect{\zeta}_2+\vect{Ad}_{\Bvect{\mathcal{X}}_2^{-1}}\vect{\zeta}_1.  
\end{align}
Assuming the independence of $\vect{\zeta}_1$ and $\vect{\zeta}_2$, the covariance of $\vect{\mathcal{X}}_\text{com}^{(1)}$ is calculated as
\begin{align}
    \vect{P}_{\text{com}}^{(1)}=\mathbb{E}[\vect{\zeta}_{\text{com}}^{(1)}(\vect{\zeta}_{\text{com}}^{(1)})^\top]\approx\vect{P}_{22}+\vect{Ad}_{\Bvect{\mathcal{X}}_2^{-1}}\vect{P}_{11}\vect{Ad}_{\Bvect{\mathcal{X}}_2^{-1}}^\top.
\end{align}
In the inductive step, we assume the following statement holds for some arbitrary $n=k$, i.e., 
\begin{align}\nonumber
\vect{\mathcal{X}}_\text{com}^{(k)}&=\Bvect{\mathcal{X}}_1\Bvect{\mathcal{X}}_2\cdots\Bvect{\mathcal{X}}_{k-1}\Bvect{\mathcal{X}}_k\exp([\vect{Ad}_{\Bvect{\mathcal{X}}_k^{-1}\cdots\Bvect{\mathcal{X}}_2^{-1}}\vect{\zeta}_1]_\wedge)\\\nonumber&\exp([\vect{Ad}_{\Bvect{\mathcal{X}}_k^{-1}\cdots\Bvect{\mathcal{X}}_3^{-1}}\vect{\zeta}_2]_\wedge)\cdots\exp([\vect{Ad}_{\Bvect{\mathcal{X}}_k^{-1}}\vect{\zeta}_{k-1}]_\wedge)\\\nonumber&\exp([\vect{\zeta}_k]_\wedge). 
\end{align}
Using the first-order terms of the BCH formula and assuming uncorrelated members, we obtain 
\begin{align}
\Bvect{\mathcal{X}}_\text{com}^{(k)}&=\Bvect{\mathcal{X}}_1\Bvect{\mathcal{X}}_2\cdots\Bvect{\mathcal{X}}_{k-1}\Bvect{\mathcal{X}}_k,\\\nonumber
 \vect{\zeta}_{\text{com}}^{(k)}&\approx\vect{\zeta}_{k}+\vect{Ad}_{\Bvect{\mathcal{X}}_k^{-1}}\vect{\zeta}_{k-1}+\cdots\\&+\vect{Ad}_{\Bvect{\mathcal{\mathcal{X}}}_k^{-1}\cdots\Bvect{\mathcal{X}}_3^{-1}}\vect{\zeta}_{2}+\vect{Ad}_{\Bvect{\mathcal{\mathcal{X}}}_k^{-1}\cdots\Bvect{\mathcal{X}}_2^{-1}}\vect{\zeta}_{1},\\\nonumber
 \vect{P}_{\text{com}}^{(k)}&\approx\vect{P}_{kk}+\vect{Ad}_{\Bvect{\mathcal{X}}_k^{-1}}\vect{P}_{k-1k-1}\vect{Ad}_{\Bvect{\mathcal{X}}_k^{-1}}^\top+\cdots\\\nonumber&+\vect{Ad}_{\Bvect{\mathcal{\mathcal{X}}}_k^{-1}\cdots\Bvect{\mathcal{X}}_3^{-1}}\vect{P}_{22}\vect{Ad}_{\Bvect{\mathcal{X}}_k^{-1}\cdots\Bvect{\mathcal{X}}_3^{-1}}^\top\\&+\vect{Ad}_{\Bvect{\mathcal{\mathcal{X}}}_k^{-1}\cdots\Bvect{\mathcal{X}}_2^{-1}}\vect{P}_{11}\vect{Ad}_{\Bvect{\mathcal{X}}_k^{-1}\cdots\Bvect{\mathcal{X}}_2^{-1}}^\top.
\end{align}
Next, the validity is shown for $n\!=\!k+1$. We have
\begin{align}
 \vect{\mathcal{X}}_\text{com}^{(k+1)}=\vect{\mathcal{X}}_1\cdots\vect{\mathcal{X}}_{k}\vect{\mathcal{X}}_{k+1}= \vect{\mathcal{X}}_\text{com}^{(k)} \vect{\mathcal{X}}_{k+1}.   
\end{align}
This can be expanded as
\begin{align}
 \vect{\mathcal{X}}_\text{com}^{(k+1)}=\Bvect{\mathcal{X}}_\text{com}^{(k)}\exp([\vect{\zeta}_{\text{com}}^{(k)}]_\wedge)\Bvect{\mathcal{X}}_{k+1}\exp([\vect{\zeta}_{k+1}]_\wedge),
\end{align}
and based on the result developed in~\eqref{eq:comp222} we obtain
\begin{align}\nonumber
\vect{\mathcal{X}}_\text{com}^{(k+1)}=\Bvect{\mathcal{X}}_{\text{com}}^{(k)}\Bvect{\mathcal{X}}_{k+1}\exp([\vect{Ad}_{\Bvect{\mathcal{X}}_{k+1}^{-1}}\vect{\zeta}_{\text{com}}^{(k)}]_\wedge)\exp([\vect{\zeta}_{k+1}]_\wedge). 
\end{align}
Therefore, the mean and uncertainty vector of $\vect{\mathcal{X}}_\text{com}^{(k+1)}$ are
\begin{align}
 \Bvect{\mathcal{X}}_\text{com}^{(k+1)}&=\Bvect{\mathcal{X}}_{\text{com}}^{(k)}\Bvect{\mathcal{X}}_{k+1}=\Bvect{\mathcal{X}}_1\cdots\Bvect{\mathcal{X}}_{k}\Bvect{\mathcal{X}}_{k+1}\\\nonumber\vect{\zeta}_{\text{com}}^{(k+1)}&\approx \vect{\zeta}_{k+1}+\vect{Ad}_{\Bvect{\mathcal{X}}_{k+1}^{-1}}\vect{\zeta}_{\text{com}}^{(k)}\\\nonumber&=\vect{\zeta}_{k+1}+\vect{Ad}_{\Bvect{\mathcal{X}}_{k+1}^{-1}}\vect{\zeta}_{k}+\vect{Ad}_{\Bvect{\mathcal{X}}_{k+1}^{-1}}\vect{Ad}_{\Bvect{\mathcal{X}}_k^{-1}}\vect{\zeta}_{k-1}\\&+\cdots+\vect{Ad}_{\Bvect{\mathcal{X}}_{k+1}^{-1}}\vect{Ad}_{\Bvect{\mathcal{\mathcal{X}}}_k^{-1}\cdots\Bvect{\mathcal{X}}_2^{-1}}\vect{\zeta}_{1}.
\end{align}
Knowing $\vect{Ad}_{\mathcal{X}\mathcal{Y}}=\vect{Ad}_{\mathcal{X}}\vect{Ad}_{\mathcal{Y}}$,  and considering the first-order terms in the BCH we obtain
\begin{align}\nonumber
 \vect{\zeta}_{\text{com}}^{(k+1)}&\approx\vect{\zeta}_{k+1}+\vect{Ad}_{\Bvect{\mathcal{X}}_{k+1}^{-1}}\vect{\zeta}_{k}+\vect{Ad}_{\Bvect{\mathcal{X}}_{k+1}^{-1}\Bvect{\mathcal{X}}_k^{-1}}\vect{\zeta}_{k-1}\\&+\cdots+\vect{Ad}_{\Bvect{\mathcal{X}}_{k+1}^{-1}\Bvect{\mathcal{\mathcal{X}}}_k^{-1}\cdots\Bvect{\mathcal{X}}_2^{-1}}\vect{\zeta}_{1}.   
\end{align}
Assuming the independence of $\vect{\zeta}_{k+1}$, and $\vect{\zeta}_{\text{com}}^{(k)}$, the covariance of $\vect{\mathcal{X}}_\text{com}^{(k+1)}$ is calculated as
\begin{align}\nonumber
    \vect{P}_{\text{com}}^{(k+1)}&\approx\vect{P}_{k+1k+1}+\vect{Ad}_{\Bvect{\mathcal{X}}_{k+1}^{-1}}\vect{P}_{kk}\vect{Ad}_{\Bvect{\mathcal{X}}_{kk}^{-1}}^\top\\\nonumber&\vect{Ad}_{\Bvect{\mathcal{\mathcal{X}}}_{k+1}^{-1}\Bvect{\mathcal{\mathcal{X}}}_{k}^{-1}}\vect{P}_{k-1k-1}\vect{Ad}_{\Bvect{\mathcal{\mathcal{X}}}_{k+1}^{-1}\Bvect{\mathcal{\mathcal{X}}}_{k}^{-1}}^\top+\cdots\\&+\vect{Ad}_{\Bvect{\mathcal{\mathcal{X}}}_{k+1}^{-1}\Bvect{\mathcal{\mathcal{X}}}_{k}^{-1}\cdots\Bvect{\mathcal{X}}_2^{-1}}\vect{P}_{11}\vect{Ad}_{\Bvect{\mathcal{\mathcal{X}}}_{k+1}^{-1}\Bvect{\mathcal{\mathcal{X}}}_{k}^{-1}\cdots\Bvect{\mathcal{X}}_2^{-1}}^\top.
\end{align}
This completes the proof for uncorrelated configurations. When the stochastic configurations are correlated, it implies that the uncertainties associated with them are  correlated, i.e., $\mathbb{E}[\vect{\zeta}_i\vect{\zeta}_j^\top]\neq\vect{0}_m$ for $i,j\in\{1,\cdots,n\}$. In this case, we write the uncertainty vector  as 
\begin{align}\nonumber
    \vect{\zeta}_{\text{com}}\!=\!\begin{bmatrix}
        \vect{I}_m&\vect{0}_m&\cdots&\vect{0}_m\\
         \vect{0}_m&\vect{Ad}_{\Bvect{\mathcal{X}}_n^{-1}}&\cdots&\vect{0}_m\\
          \vdots&\vdots&\ddots&\vdots\\
           \vect{0}_m&\vect{0}_m&\cdots&\vect{Ad}_{\Bvect{\mathcal{\mathcal{X}}}_n^{-1}\cdots\Bvect{\mathcal{X}}_2^{-1}}
    \end{bmatrix}\!\begin{bmatrix}
        \vect{\zeta}_n\\\vect{\zeta}_{n-1}\\\vdots\\\vect{\zeta}_1
    \end{bmatrix}\!=\!\vect{\mathcal{M}}\vect{\zeta}.
    \end{align}
The covariance is then calculated as
\begin{align}\nonumber
 \vect{P}_{\text{com}}&=\mathbb{E}[\vect{\zeta}_{\text{com}}\vect{\zeta}_{\text{com}}^\top]\approx\vect{\mathcal{M}}\mathbb{E}[\vect{\zeta}\vect{\zeta}^\top]\vect{\mathcal{M}}^\top.
 \end{align}
We let $\vect{P}_{ij}=\mathbb{E}[\vect{\zeta}_{i}\vect{\zeta}_{j}^\top]$ for $i,j\in\{1,\cdots,n\}$, then 
\begin{align}\nonumber
 \vect{P}_{\text{com}}\approx\vect{\mathcal{M}}\begin{bmatrix}
  \vect{P}_{nn}&\vect{P}_{nn-1}&\cdots&\vect{P}_{n1}\\ 
  \vect{P}_{n-1n}&\vect{P}_{n-1n-1}&\cdots&\vect{P}_{n-11}\\
  \vdots&\vdots&\ddots&\vdots\\
  \vect{P}_{1n}&\vect{P}_{1n-1}&\cdots&\vect{P}_{11}
 \end{bmatrix}\vect{\mathcal{M}}^\top.
 \end{align}

\section{Proof of Difference Operation}\label{app: diff}
Based on the difference definition we have 
$\vect{\mathcal{X}}_{\text{dif}}=\vect{\mathcal{X}}_1^{-1}\vect{\mathcal{X}}_2$. We let $\vect{\mathcal{Y}}=\vect{\mathcal{X}}_1^{-1}$ and use the inverse operation to get
\begin{align}
 \vect{\mathcal{Y}}\!=\!\vect{\mathcal{X}}_1^{-1}\!=\!\Bvect{\mathcal{Y}}\exp([\vect{\zeta}_{\mathcal{Y}}]_\wedge)\!=\!\Bvect{\mathcal{X}}_1^{-1}\exp(-[\vect{Ad}_{\Bvect{\mathcal{X}}_1}\vect{\zeta}_1]_\wedge).   
\end{align}
We therefore have
$
\vect{\mathcal{X}}_{\text{dif}}\!=\!\Bvect{\mathcal{Y}}\exp([\vect{\zeta}_\mathcal{Y}]_\wedge)\Bvect{\mathcal{X}}_2\exp([\vect{\zeta}_2]_\wedge).$
Using the composition rule results in  
\begin{align}
 \Bvect{\mathcal{X}}_{\text{dif}}&=\Bvect{\mathcal{Y}}\Bvect{\mathcal{X}}_2=\Bvect{\mathcal{X}}_1^{-1}\Bvect{\mathcal{X}}_2,\\\nonumber   \vect{\zeta}_{\text{dif}}&\approx\vect{\zeta}_2+\vect{Ad}_{\Bvect{\mathcal{X}}_2^{-1}}\vect{\zeta}_{\mathcal{Y}}=\vect{\zeta}_2-\vect{Ad}_{\Bvect{\mathcal{X}}_2^{-1}}\vect{Ad}_{\Bvect{\mathcal{X}}_1}\vect{\zeta}_1\\&=\vect{\zeta}_2-\vect{Ad}_{\Bvect{\mathcal{X}}_2^{-1}\Bvect{\mathcal{X}}_1}\vect{\zeta}_1,\\\nonumber \vect{P}_{\text{dif}}&=\mathbb{E}[\vect{\zeta}_{\text{dif}}\vect{\zeta}_{\text{dif}}^\top]\approx\vect{P}_{22}-\vect{P}_{12}^\top\vect{Ad}_{\Bvect{\mathcal{X}}_2^{-1}\Bvect{\mathcal{X}}_1}^\top-\vect{Ad}_{\Bvect{\mathcal{X}}_2^{-1}\Bvect{\mathcal{X}}_1}\vect{P}_{12}\\&+\vect{Ad}_{\Bvect{\mathcal{X}}_2^{-1}\Bvect{\mathcal{X}}_1}\vect{P}_{11}\vect{Ad}_{\Bvect{\mathcal{X}}_2^{-1}\Bvect{\mathcal{X}}_1}^\top.
\end{align}

\section{Constrained Fusion Derivation}\label{app: const}
For  $\Tvect{g}(\vect{\zeta}_c)=\vect{\Gamma}\vect{\zeta}_c$ the Lagrangian reduces to:
\begin{align}\nonumber
    \mathcal{L}\!=\!(\vect{\zeta}_u\!+\!\mathcal{J}^{-1}_r\!(\vect{\zeta}_u\!)\vect{\zeta}_c)^{\!\top}\vect{W}_c(\vect{\zeta}_u\!+\!\mathcal{J}^{-1}_r\!(\vect{\zeta}_u)\vect{\zeta}_c)\!+\!2\vect{\lambda}^{\!\top}(\vect{\Gamma}\vect{\zeta}_c\!-\!\Tvect{d}),
\end{align}
and therefore its derivatives with respect to $\vect{\zeta}_c$ and $\vect{\lambda}$ are
\begin{align}\label{eq: Lagrange set of equation2}
 &\mathcal{J}^{-\top}_r({\vect{\zeta}_u})\vect{W}_c(\vect{\zeta}_u\!+\!\mathcal{J}^{-1}_r(\vect{\zeta}_u)\vect{\zeta}_c)\!+\vect{\Gamma}^\top\vect{\lambda}\!=\!\vect{0}_{m\times 1},\\\label{eq: Lagrange set of equation 22}
  &\vect{\Gamma}\vect{\zeta}_c-\Tvect{d}=\vect{0}_{r\times 1}.
\end{align}
From~\eqref{eq: Lagrange set of equation2} we obtain
\begin{align}\label{eq: htg}
 \vect{\zeta}_c&=-\vect{\mathfrak{J}}
 \big(\mathcal{J}_r^{-\top}(\vect{\zeta}_u)\vect{W}_c\vect{\zeta}_u\!+\!\vect{\Gamma}^\top\vect{\lambda}\big),
 \end{align}
where $\vect{\mathfrak{J}}=\big(\mathcal{J}_r^{-\top}({\vect{\zeta}_u})\vect{W}_c\mathcal{J}_r^{-1}({\vect{\zeta}_u})\big)^{-1}$. Substituting~\eqref{eq: htg} in~\eqref{eq: Lagrange set of equation 22} and solving for $\vect{\lambda}$ gives $\vect{\lambda}=-\big(\vect{\Gamma}\vect{\mathfrak{J}}\vect{\Gamma}^\top\big)^{-1} \big(\vect{\Gamma}\vect{\mathfrak{J}}\mathcal{J}_r^{-\top}({\vect{\zeta}_u})\vect{W}_c\vect{\zeta}_u+\Tvect{d}\big). $
Finally, by substituting the calculated $\vect{\lambda}$ in~\eqref{eq: htg} and rearranging the terms, it is verified 
  \begin{align}\label{eq: const tc}\nonumber
    \vect{\zeta}_c&=\vect{\mathfrak{J}}\Big[\Big(\vect{\Gamma}^\top(\vect{\Gamma}\vect{\mathfrak{J}}\vect{\Gamma}^\top)^{-1}\vect{\Gamma}\vect{\mathfrak{J}}\mathcal{J}_r^{-\top}(\vect{\zeta}_u)\vect{W}_c\\&-\mathcal{J}_r^{-\top}(\vect{\zeta}_u)\vect{W}_c\Big)\vect{\zeta}_u+\vect{\Gamma}^\top(\vect{\Gamma}\vect{\mathfrak{J}}\vect{\Gamma}^\top)^{-1}\Tvect{d}\Big].
\end{align}

\end{document}